%% file: manuscript.tex
\pdfoutput=1
\documentclass{article}
\usepackage{iclr2027_conference,times}
\usepackage{amsmath,amssymb,booktabs,graphicx,longtable,xcolor,float}
\usepackage{placeins}
\usepackage{enumitem}
\usepackage{needspace}
\usepackage{titletoc}
\usepackage{hyperref}
\definecolor{linkblue}{rgb}{0.0,0.2,0.55}
\hypersetup{colorlinks=true,linkcolor=linkblue,citecolor=linkblue,urlcolor=linkblue,
  pdftitle={Safe to Omit, Worth Omitting? Keeping the Answer and Cutting Latency When an LLM Agent Skips Another's Message},
  pdfauthor={Xinyang Li, Chih-Hsuan Yang, Xiaolong Ma, Anjir Ahmed Chowdhury, Vikram Vasudevan, Rajeev Thakur, Feng Yan}}
\usepackage{url}
\iclrfinalcopy
\newcounter{algobox}
\newcommand{\rev}[1]{#1}
\newcommand{\revcolor}{}

\title{Safe to Omit, Worth Omitting?\\
Keeping the Answer and Cutting Latency\\
When an LLM Agent Skips Another's Message}

\author{\bfseries Xinyang Li$^{1,\dagger}$ \quad Chih-Hsuan Yang$^{2,\ast}$ \quad Xiaolong Ma$^{2}$ \\
\bfseries Anjir Ahmed Chowdhury$^{1}$ \quad Vikram Vasudevan$^{3}$ \\
\bfseries Rajeev Thakur$^{2}$ \quad Feng Yan$^{1}$ \\
\normalfont $^{1}$University of Houston \quad $^{2}$Argonne National Laboratory \\
\normalfont $^{3}$Oregon State University \\
\normalfont $^{\ast}$Corresponding author: \texttt{bellayang@anl.gov}}

\begin{document}
\maketitle
{\renewcommand{\thefootnote}{\fnsymbol{footnote}}\footnotetext[2]{Work done while visiting Argonne National Laboratory.}}
\vspace{-2.4em}
\begin{center}
\href{https://piedpiperrr.github.io/safe-to-omit/}{Project Website}\quad$|$\quad
\href{https://github.com/Piedpiperrr/safe-to-omit}{Code}\quad$|$\quad
\href{https://huggingface.co/datasets/AgentsSci/ICLR2027-SafeToOmit}{Dataset}
\end{center}

\begin{abstract}
\input{abstract.tex}
\end{abstract}

\input{sec_intro.tex}
\input{sec_problem.tex}
\input{sec_setup.tex}
\input{sec_results.tex}
\input{sec_discussion.tex}
\input{sec_related.tex}
\input{sec_limitations.tex}
\input{sec_conclusion.tex}

\input{sec_statements.tex}

\setlength{\bibsep}{3pt plus 1pt minus 0.5pt}
\bibliography{refs_p2}
\bibliographystyle{iclr2027_conference}

\clearpage
\appendix
\titlecontents{section}[1.8em]{\addvspace{2pt}}{\contentslabel{1.8em}}{\hspace*{-1.8em}}{\titlerule*[0.6pc]{.}\contentspage}
\titlecontents{subsection}[4.2em]{}{\contentslabel{2.4em}}{\hspace*{-2.4em}}{\titlerule*[0.6pc]{.}\contentspage}
\section*{Contents of the appendix}
\startcontents[appendix]
{\setlength{\parskip}{0pt}\printcontents[appendix]{}{1}{\setcounter{tocdepth}{2}}}
\clearpage
\input{new_app_A.tex}
\FloatBarrier
\input{new_app_B.tex}
\FloatBarrier
\input{new_app_C.tex}
\FloatBarrier
\input{new_app_D.tex}

\FloatBarrier
\input{new_app_E.tex}
\FloatBarrier
\input{new_app_H.tex}

\FloatBarrier
\input{new_app_I.tex}
\FloatBarrier
\input{new_app_J.tex}
\FloatBarrier
\input{new_app_F.tex}

\FloatBarrier
\input{app_prospective.tex}
\FloatBarrier
\input{new_app_G.tex}
\FloatBarrier
\input{new_app_K.tex}

\FloatBarrier
\input{new_app_L.tex}

\end{document}

%% file: abstract.tex
In collaborative multi-agent systems, a receiver agent often waits for a message from a helper agent, which can be either natural-language text or its latent KV cache. While this communication can improve performance, it also adds latency. We investigate when a receiver agent can skip the helper agent's message safely (the final answer rarely changes) and beneficially (the time saved exceeds the cost of deciding). We propose a dynamic skipping mechanism that reads the receiver agent's confidence without the message and applies a finite-sample statistical test. This test certifies a confidence threshold at which, with high probability, at most 5\% of skipped answers change; if no threshold passes, the system keeps communicating. We evaluate our mechanism on OpenBookQA, ARC-Challenge, and MMLU-Pro with three Qwen model pairs, on cross-family receiver and helper agents, and on SQuAD and GSM8K. Four main insights emerge. First, oracle headroom, a common metric for routing among paths, is weak evidence that messages help: control inputs without answer-specific content, matched to change answers as often as real messages, reproduce a median 76\% of the oracle's gain over the receiver agent. Second, whether a message can be skipped depends on how often and how predictably it alters the receiver agent's answer, not on its overall accuracy; every certified KV-cache path is less accurate than the receiver agent alone. Third, where text messages help, skipping can keep most of their gain: in the four such settings with unseen questions, the deployed policies skip 59 to 80\% of them and keep an estimated 76 to 95\% of the gain over the receiver agent. In two prospective tests with every rule fixed in advance, the procedure skips the verifier agent of a three-agent system on 39.9\% of sealed questions while keeping 12.55 of its 12.80-point gain, and a 32B helper agent on 68.4\% of held-out questions while keeping its whole gain. Fourth, safety does not determine efficiency: at batch size one, skipping text messages saves hundreds of milliseconds per query, whereas skipping latent KV cache saves far less. Certification extends to SQuAD, which a label-free rule chose from eight candidate settings, but not to GSM8K. In eleven tests on unseen questions, every certified policy we could test kept its change rate's 95\% upper bound below 5\% under our answer parsers.

%% file: sec_intro.tex
\section{Introduction}\label{sec:intro}

Collaborative large language model systems increasingly rely on agents consulting one another to solve complex tasks \citep{du2024debate,li2023camel}. A common paradigm has a \emph{helper agent} generate a message and pass it to a \emph{receiver agent}. The message is either natural-language text (\emph{Text}) or latent internal states, such as the key--value (KV) cache in which the helper stores what it has read (Cache-to-Cache, C2C; \citealp{fu2025c2c}). While this communication can enhance performance, it introduces a bottleneck: the receiver agent must wait for the helper agent's output, adding latency and computational overhead \citep{zhang2025agentprune}. For practical deployment, it is crucial to determine when such communication is necessary and when the receiver agent can safely skip the message to reduce inference time.

Identifying which queries genuinely require communication is challenging. Work on routing and ensembling motivates keeping several paths with \emph{oracle headroom} \citep{jiang2023blender,hu2024routerbench}: the accuracy of an oracle that picks a correct path for each question, minus the accuracy of the best single path. High headroom means that different paths answer different questions correctly. For agent communication, it would suggest that helper messages carry content the receiver agent lacks, but it is weak evidence of this. Control inputs without answer-specific content, matched in how often they change answers, yield comparable headroom (\S\ref{sec:complementarity}), so headroom can arise from answer changes alone. Systems therefore need a mechanism that prunes redundant messages based on actual behavior.

\begin{figure}[t]\centering
\includegraphics[width=\linewidth]{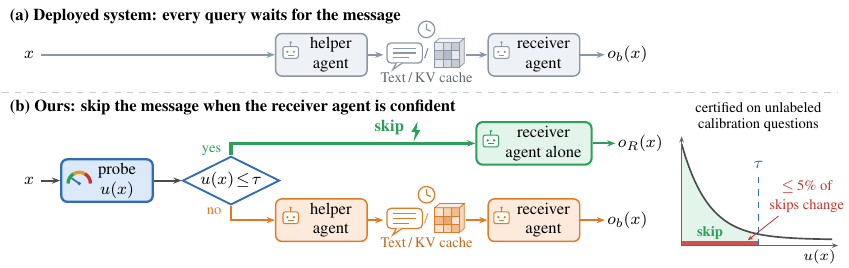}
\caption{\textbf{Dynamic message skipping in collaborative LLM agents.} (a) The deployed system always waits for the helper agent's message, natural-language text (Text) or a projection of its KV cache (C2C), before the receiver agent answers. (b) Our method reads the receiver agent's uncertainty $u(x)$ from one forward pass (the probe): if $u(x)\le\tau$, the receiver agent answers alone (green); otherwise the deployed system runs (orange). $\tau$ is certified on unlabeled calibration questions so that, with high probability, at most 5\% of skipped answers differ from the deployed system's (red strip); if no threshold passes, the system always communicates. Inset: distribution of $u(x)$.}
\label{fig:overview}
\end{figure}

We formalize the decision to skip a message based on two criteria: safety and value (Figure~\ref{fig:overview}). We treat the system that always sends the message as a fixed, deployed reference, and consider skipping safe if the receiver agent alone rarely gives a different answer from this reference. Bounding such answer changes is vital, as users notice changed answers even when overall accuracy remains constant \citep{yan2021pct,dutta2024accuracy}. This agreement-based formulation offers a practical advantage. Unlike accuracy, agreement can be checked without ground-truth labels, allowing operators to calibrate the skipping mechanism directly on abundant unlabeled deployment traffic. To achieve this, we read the receiver agent's confidence from one forward pass and apply a statistical test on a finite calibration sample, building on risk-controlled early exiting \citep{schuster2022calm,jazbec2024fastyetsafe} and Learn then Test \citep{angelopoulos2022ltt}. This test certifies a threshold at which, with high probability, at most 5\% of skipped answers change. If no threshold passes, the system conservatively keeps communicating. Furthermore, skipping is worth it only if the latency saved on skipped queries exceeds the computational cost of the confidence probe paid across all queries. \rev{Unlike an early-exit gate, which reads intermediate states of the computation it may cut short, our probe is a separate forward pass paid on every query, and the message it decides about comes from another agent and can make the answer less accurate.}

Our main contributions \rev{answer four questions in turn (Q1 to Q4; \S\ref{sec:results})}:
\begin{itemize}[leftmargin=1.2em,itemsep=1pt,topsep=2pt,parsep=0pt]
\item \textbf{\rev{Q1. Does high oracle headroom show that a message is needed? No.}} Control inputs without answer-specific content reproduce about three quarters of the oracle's gain over the receiver agent. Whether this null baseline separates text messages from latent representations depends on how the match is designed (\S\ref{sec:complementarity}).
\item \textbf{\rev{Q2. What makes a message safe to skip?}} \rev{Skippability is a behavioral property: it} depends on how often and how predictably the message alters the receiver agent's answer, not on its overall accuracy. \rev{The probe ranks answer changes better (higher AUROC) in every certified setting than in any fallback, and a label-free rule on fit-split counts reproduces all 25 outcomes it covers and predicted two later fallbacks (Appendix~\ref{app:predict}).} The certified C2C paths are all less accurate than the receiver agent alone under our answer parser, so certification here identifies redundant paths. It applies to multiple-choice tasks and one extractive setting (SQuAD; \S\ref{sec:rq1}). \rev{On unseen data, every deployed policy we could test keeps its change rate's 95\% upper bound below 5\% (\S\ref{sec:rq4}).}
\item \rev{\textbf{Q3. When a message helps, does skipping lose its gain? Mostly not.} In four settings where Text helps, deployed policies keep an estimated 76 to 95\% of its gain on unseen questions, with the kept gain's interval above zero in three (\S\ref{sec:rq2}). The certificate bounds the accuracy loss by $\kappa\alpha$ (skip rate times tolerance), which can exceed the gain; post hoc, 7 to 12 labels bounded the loss by one point for three of six such paths; two prospective tests passed (three agents; a 32B pair).}
\item \textbf{\rev{Q4. Does a safe skip save time? Only if the time saved on skips exceeds the probe's cost.}} \rev{A safe skip need not be a profitable one:} at batch size one, skipping text messages saves a mean of hundreds of milliseconds per query, whereas skipping latent ones saves far less (\S\ref{sec:rq3}).
\end{itemize}
\rev{Appendix~\ref{app:protocol} defines every term in one table and maps each appendix to these questions.}

%% file: sec_problem.tex
\section{A framework for dynamic message skipping}\label{sec:problem}

We frame the decision to skip communication as a risk-controlled routing problem. Given a question $x$, a deployed reference path $b$ (e.g., passing a Text or KV cache message from the helper agent to the receiver agent) produces an answer $o_b(x)$ with latency $c_b(x)$. The receiver agent answering alone, path $R$, produces an answer $o_R(x)$ with latency $c_R(x)$. An answer is the label that one fixed parser reads from the receiver agent's output, so the system outputs this parsed label, not raw text; outputs the parser cannot read share one \textsc{invalid} label and so agree.

A routing policy $\pi_\tau$ computes an uncertainty score $s(x)$ at cost $c_s(x)$, skips the message and routes to $R$ when $s(x)\le\tau$, and runs $b$ otherwise. The overall latency of the policy is $C_\pi(x)=c_s(x)+c_{\pi_\tau(x)}(x)$, and its \emph{coverage} $\kappa(\tau)=\Pr[s(x)\le\tau]$ is the fraction of queries that skip the message.

\paragraph{Bounding the consistency risk.} Skipping should rarely change the deployed system's answer. The \emph{change rate} $\rho(\tau)$ is the conditional probability of disagreement on skipped queries, relative to the deployed answer rather than the correct one:
\begin{equation}\label{eq:rho}
\rho(\tau)=\Pr\!\left[o_R(x)\neq o_b(x)\mid s(x)\le\tau\right].
\end{equation}
Skipping is safe at a tolerance $\alpha$ if $\rho(\tau)\le\alpha$. Because the policy and the reference differ only on skipped queries whose answers change, the accuracy change is bounded by the probability of skipping and changing the answer:
\begin{equation}\label{eq:acc}
\left|\mathrm{acc}(\pi_\tau)-\mathrm{acc}(b)\right|\le\kappa(\tau)\,\rho(\tau).
\end{equation}
When $\rho(\tau)\le\alpha$, accuracy changes by at most $\kappa(\tau)\alpha$, which can exceed a helpful path's gain (\S\ref{sec:rq2}).

\begin{figure}[b]\revcolor\setlength{\fboxsep}{5pt}\noindent\fbox{\parbox{\dimexpr\linewidth-2\fboxsep-2\fboxrule\relax}{\small\refstepcounter{algobox}\label{alg:certify}%
\textbf{Algorithm \thealgobox: certifying and deploying a skipping threshold.}\\[2pt]
\emph{Inputs:} receiver-alone path $R$, reference path $b$, score $s$, tolerance $\alpha$, per-test level $\delta$, $m$ candidate quantiles $q$, a fit split, and an unlabeled calibration split.\\[3pt]
\textbf{1.}~\emph{Fit.} For $q<1$, $\tau_q$ is the $q$-quantile of $s(x)$ on the fit split; the candidate $q=1$ always skips.\\[3pt]
\textbf{2.}~\emph{Calibrate without labels.} Of the $n_q$ calibration questions that candidate $q$ skips ($s(x)\le\tau_q$), $k_q$ have $o_R(x)\neq o_b(x)$. The candidate passes if $\Pr[\mathrm{Bin}(n_q,\alpha)\le k_q]\le\delta$ (Eq.~\eqref{eq:test}).\\[3pt]
\textbf{3.}~\emph{Optional accuracy check (\S\ref{sec:rq2}).} Label only the skipped disagreements and count the lost corrections $l_q$ ($b$ correct, $R$ not). With $N$ calibration questions and a loss tolerance $\varepsilon$, also require $\Pr[\mathrm{Bin}(N,\varepsilon)\le l_q]\le\delta$.\\[3pt]
\textbf{4.}~\emph{Deploy.} Use the largest candidate that passes: answer with $R$ when $s(x)\le\tau_q$ and with $b$ otherwise. If none passes, always run $b$ (fallback).}}\end{figure}
\paragraph{Statistical certification via finite-sample testing.} To certify $\rho(\tau)\le\alpha$ without ground-truth labels, we use Learn then Test \citep{angelopoulos2022ltt} \rev{(Algorithm~\ref{alg:certify})}. The score $s(x)$ is the receiver agent's uncertainty: one minus its maximum predicted answer probability from one forward pass. \rev{The $m$ candidate thresholds $\tau_q$ are $q$-quantiles of the scores on a separate fit split, set before any calibration outcome is read. The score need not be continuous, because tied scores only merge candidate thresholds. On a disjoint, unlabeled calibration split, $n_q$ queries fall at or below $\tau_q$, of which $k_q$ change the answer; an exact binomial test of $H_0\colon\rho(\tau_q)\ge\alpha$ accepts $\tau_q$ if}
\begin{equation}\label{eq:test}
\Pr\!\left[\mathrm{Bin}(n_q,\alpha)\le k_q\right]\le\delta .
\end{equation}
We deploy the largest accepted threshold and otherwise always run $b$; such a fallback means only that the test could not certify skipping with the available calibration data. Testing each of $m$ predetermined candidates at level $\delta$ bounds by $m\delta$ the probability of deploying a threshold whose true change rate is at least $\alpha$ (a Bonferroni correction; \citealp{dunn1961multiple}), assuming i.i.d.\ calibration questions and a parser fixed independently of them. The certificate covers a threshold together with the parser, so an operator should certify the answer form its application consumes.

\paragraph{Evaluating the empirical value.} Skipping is valuable only if the latency saved on skipped queries outweighs the probe's cost, paid on every query. The expected saving over the reference path is:
\begin{equation}\label{eq:saving}
\mathbb E\!\left[c_b(x)-C_\pi(x)\right] = \kappa(\tau)\,\mathbb E\!\left[c_b(x)-c_R(x)\mid s(x)\le\tau\right]-\mathbb E\!\left[c_s(x)\right].
\end{equation}
Skipping is worthwhile when the 95\% bootstrap interval of the paired mean saving is above zero.

%% file: sec_setup.tex
\section{Experimental setup}\label{sec:setup}

\paragraph{\rev{Models and communication paths.}} \rev{ We compare two ways for a helper agent to pass information to a receiver agent, in three helper and receiver pairs from the Qwen family \citep{qwen2025qwen25,yang2025qwen3}: \emph{small} (Qwen2.5-0.5B-Instruct with Qwen3-0.6B), \emph{medium} (Qwen2.5-1.5B-Instruct with Qwen3-1.7B), and \emph{large} (Qwen2.5-7B-Instruct with Qwen3-8B). On the Text path, the helper agent writes the background knowledge needed for the question without answering it, and the receiver agent reads this message before answering (prompts: Appendix~\ref{app:prompts}). On the latent path, Cache-to-Cache (C2C; \citealp{fu2025c2c}), a learned fuser projects the helper agent's key--value cache into the receiver agent. We use the released fusers and reproduce their official outputs (Appendix~\ref{app:revision}). All models decode greedily with thinking disabled, as in the official C2C evaluator.}

\paragraph{\rev{Certification setup and answer parsing.}} \rev{ The probe measures the receiver agent's uncertainty before it sees any message. We run one receiver agent prefill over the question and options followed by \texttt{The correct answer is}, renormalize the probabilities of the option labels, and set $s(x)=u(x)=1-\max_l p_l$, a score we call ProbeMax \citep{hendrycks2017baseline,geng2024survey}. SQuAD has no listed options, so the probe generates the receiver agent's answer, and the score is one minus that answer's probability. We set $\alpha=0.05$ and $\delta=0.001$ and use 20 candidate thresholds: 19 fit-split quantiles and always skipping. With 20 tests at level $\delta$, the chance of certifying an unsafe threshold is at most 0.02 per setting. Answers are compared after one fixed rule-based parser extracts a single final choice from the receiver agent's output (on SQuAD, a normalized answer string; Appendix~\ref{app:prompts}); unreadable or ambiguous outputs become \textsc{invalid}. As a second convention, we relabel outputs with the official C2C evaluator's extraction, which reads only labels A to D (Appendix~\ref{app:parser}).}

\input{tab_main_complementarity.tex}

\paragraph{\rev{Benchmarks and data splits.}} \rev{ Multiple choice lets us detect an answer change exactly, by comparing two parsed labels, without knowing the correct answer. Our primary benchmarks are therefore OpenBookQA (OBQA; \citealp{mihaylov2018obqa}) and ARC-Challenge (ARC; \citealp{clark2018arc}), two knowledge-dependent science benchmarks from C2C's published evaluation. All three pairs also run on the harder MMLU-Pro \citep{wang2024mmlupro}, which spans 14 disciplines. To test answers outside a fixed label set, we add SQuAD \citep{rajpurkar2016squad}, where the answer is a span of a passage, and GSM8K \citep{cobbe2021gsm8k}, where it is a number derived by reasoning. Each benchmark is divided into three disjoint splits: thresholds are set on the \emph{fit} split, tested on the \emph{calibration} split, and evaluated on the \emph{development} split. OBQA has 2,100, 1,366, and 742 questions in these splits, all from its official training set. ARC has 670, 448, and 299, and MMLU-Pro, built from its test rows, has 3,000, 6,000, and 2,641 (Appendix~\ref{app:protocol}). Evaluation data are of three kinds: \emph{development} data were observed during the study, \emph{held-out} data were never used by the pair tested, and \emph{sealed} data were untouched until a frozen policy ran (Appendix~\ref{app:provenance}).}

\paragraph{\rev{Evaluation settings and measurement.}} \rev{ A \emph{setting} is a model pair, a benchmark, and a reference path. The main study has 18 Qwen-pair settings, and 13 extension settings test other models and answer forms, for 31 certification outcomes in all (Figure~\ref{fig:boundary_map}). Nine extension settings use cross-family agents (an OLMo-2-1124-7B-Instruct receiver agent \citep{olmo2025olmo2}, a Llama-3.2-1B-Instruct helper agent \citep{grattafiori2024llama3}, and a Llama-3.1-8B-Instruct receiver agent with the large helper agent, the \emph{Llama-8B pair}), three use better-informed or stronger helper agents, and one uses SQuAD. Every extension protocol was frozen before its outcomes were generated, committing to report every setting. Latency is the end-to-end wall-clock time per query at batch size one. Each deployed policy is replayed on two A100 GPUs, interleaved with its reference path, on a fixed panel of 128 development questions per benchmark (the first 128 in a fixed pseudorandom order). The replays are repeated on a second hardware configuration (dedicated nodes with the same GPU type; Appendix~\ref{app:revision}). Latency and accuracy use paired bootstrap 95\% intervals; change-rate intervals are exact two-sided Clopper--Pearson ones \citep{clopper1934}. The development AUROC summarizes how well the score predicts answer changes: it is the probability that a question where the receiver agent alone disagrees with the reference gets a higher uncertainty than one where they agree (ties count one half).}

%% file: tab_main_complementarity.tex
\begin{table}[t]
\caption{Input changes without answer-specific content create comparable oracle headroom and reproduce about three quarters of the oracle's gain over the receiver agent alone ($R$). Correct answers: development questions each path answers correctly, out of $N$. Headroom: oracle over $R$, Text, and C2C minus the best single path (points). Null/real: the content-free null's headroom or oracle gain over $R$ divided by the real one; near or above 1, the null reproduces the effect. Stratified: the null also matches how many changes fall on questions $R$ answers correctly and incorrectly. $^{*}$95\% interval excludes 1 (per-path columns); $^{\dagger}$not fully matched (too few questions); $^{\S}$lower bound (0.99 rescaled). All intervals: Appendix~\ref{app:complementarity}.}
\label{tab:complementarity_main}
\centering
\footnotesize
\setlength{\tabcolsep}{4pt}
\resizebox{\linewidth}{!}{\begin{tabular}{llrrrcccccccc}
\toprule
& & \multicolumn{4}{c}{Real paths} & \multicolumn{3}{c}{Null / real, all paths} & \multicolumn{4}{c}{Null / real gain, per path}\\
\cmidrule(lr){3-6}\cmidrule(lr){7-9}\cmidrule(lr){10-13}
& & \multicolumn{3}{c}{Correct answers} & Oracle & \multicolumn{2}{c}{Headroom} & Gain & \multicolumn{2}{c}{Unstratified} & \multicolumn{2}{c}{Stratified}\\
\cmidrule(lr){3-5}\cmidrule(lr){7-8}\cmidrule(lr){10-11}\cmidrule(lr){12-13}
Pair & Benchmark ($N$) & \multicolumn{1}{c}{$R$} & \multicolumn{1}{c}{Text} & \multicolumn{1}{c}{C2C} & headroom (pp) & Unstratified & Stratified & over $R$ & Text & C2C & Text & C2C\\
\midrule
small & OBQA (742) & 285 & 346 & 366 & 17.79 & 1.05 & 0.79 & 0.66 & 0.72$^{*}$ & 0.69$^{*}$ & 0.80$^{*}$ & 0.80$^{*}$\\
small & ARC (299) & 110 & 126 & 155 & 16.39 & 1.15 & 0.73 & 0.74 & 0.85 & 0.75$^{*}$ & 0.84 & 0.86\\
\addlinespace[2pt]
medium & OBQA (742) & 490 & 523 & 486 & 9.43 & 1.30 & 1.00$^{\dagger}$ & 0.90 & 0.80$^{*}$ & 1.13 & 0.97 & 1.20\\
medium & ARC (299) & 217 & 221 & 202 & 9.03 & 0.95$^{\dagger}$ & 0.94$^{\dagger}$ & 0.85$^{\dagger}$ & 0.74 & 1.71$^{*}$ & 0.79 & 1.48\\
\addlinespace[2pt]
large & OBQA (742) & 617 & 645 & 598 & 3.37 & 1.60 & 1.08 & 0.76 & 0.55$^{*}$ & 1.31 & 0.81$^{*}$ & 1.08\\
large & ARC (299) & 268 & 274 & 266 & 2.01 & 1.27 & 0.97$^{\dagger}$ & 0.64 & 0.50$^{*}$ & 0.99 & 0.90$^{\S}$ & 1.35\\
large & MMLU-Pro (2641) & 1282 & 1319 & 1233 & 9.05 & 1.09 & 1.02 & 0.97 & 0.86$^{*}$ & 1.25$^{*}$ & 0.89 & 1.22$^{*}$\\
\midrule
\multicolumn{6}{l}{Median over the seven populations} & 1.15 & 0.97 & 0.76 & 0.74 & 1.13 & 0.84 & 1.20\\
\bottomrule
\end{tabular}}
\end{table}

%% file: sec_results.tex
\section{Results}\label{sec:results}

\rev{\noindent\fbox{\parbox{\dimexpr\linewidth-2\fboxsep-2\fboxrule\relax}{\footnotesize
\textbf{Reading guide.} A message can be skipped safely when it rarely and predictably changes the receiver agent's answer; oracle headroom does not show this, and accuracy does not decide it; whether the skip pays is a separate, measured question. The results answer four questions in turn.\\[1pt]
\textbf{Q1.} Does high oracle headroom show that a message is needed? No (\S\ref{sec:complementarity}).\\
\textbf{Q2.} What makes a message safe to skip? How often and how predictably it changes the answer (\S\ref{sec:rq1}, \S\ref{sec:rq4}).\\
\textbf{Q3.} When a message helps, does skipping lose its gain? Mostly not (\S\ref{sec:rq2}).\\
\textbf{Q4.} Does a safe skip save time? Only if the time saved on skips exceeds the probe's cost (\S\ref{sec:rq3}).\\[1pt]
\textbf{Example} (Llama-8B pair, ARC). The policy skips Text on 68\% of unseen questions and changes 1.9\% of the skipped answers (Q2), keeping 3.24 of Text's 3.41-point gain (Q3; Table~\ref{tab:gainkept}); the development replay saves 419~ms per query (Q4; Table~\ref{tab:e2e}).}}}

\subsection{Oracle headroom is weak evidence of communication utility}\label{sec:complementarity}

\rev{\textbf{Answer to Q1: no.} Headroom arises even when the added inputs carry no answer-specific content.} \textbf{What oracle headroom shows.} In each of our seven populations (a model pair on one benchmark), an oracle that picks a correct path among the receiver agent alone, Text, and C2C beats the best single path by 2.01 to 17.79 points (Table~\ref{tab:complementarity_main}). \rev{This shows only that different paths answer different questions correctly, not that the helper agent's messages carry content the receiver agent needs.}

\rev{\textbf{A null baseline without content.} To test whether headroom needs message content, we rerun the receiver agent on two inputs that say nothing about the answer: the same question with its options rotated by one position, and the question with a message written for another question. LLM answers are sensitive to such changes \citep{zheng2024mcq,pezeshkpour2024order}. An oracle over the receiver agent and these two inputs already matches or exceeds the real headroom in six of seven populations. The more answers an input changes, the more an oracle can gain, so we also match the null to the real paths: we resample the perturbed answers until the null changes as many answers as the real paths do, jointly or per path (Appendix~\ref{app:complementarity}). Counts are matched in total (unstratified) or separately where the receiver agent alone is right and wrong (stratified).}

\rev{\textbf{Content-free changes create comparable headroom.} With the matched null in place of the real messages, the oracle gains about as much over its own best single path as the real oracle does (median ratio 1.15 [1.00, 1.32], or 0.97 [0.80, 1.04] stratified; Table~\ref{tab:complementarity_main}). Measured from a common baseline, the receiver agent alone, the null reproduces a median 76\% [68, 89] of the real oracle's gain, or 85\% [73, 94] stratified. These ratios are descriptive: they show that headroom arises without answer-specific content, not how much of the real gain comes from content.}

\rev{\textbf{Whether Text stands apart from C2C depends on how the null is matched.} For each real path, we divide the matched null's gain by the path's gain. A ratio significantly below 1 means that the path adds correct answers beyond what answer changes alone produce. C2C does not reach this in any medium- or large-pair population under either match; only the small pair, whose receiver agent is the one in the C2C authors' main evaluation, does. Text's ratio is below 1 wherever it is determined, significantly so in five of seven unstratified but only two of six stratified (Appendix~\ref{app:complementarity}).}

\subsection{Messages that rarely and predictably change the answer can be skipped}\label{sec:rq1}

\begin{figure}[t]\centering
\includegraphics[width=\linewidth]{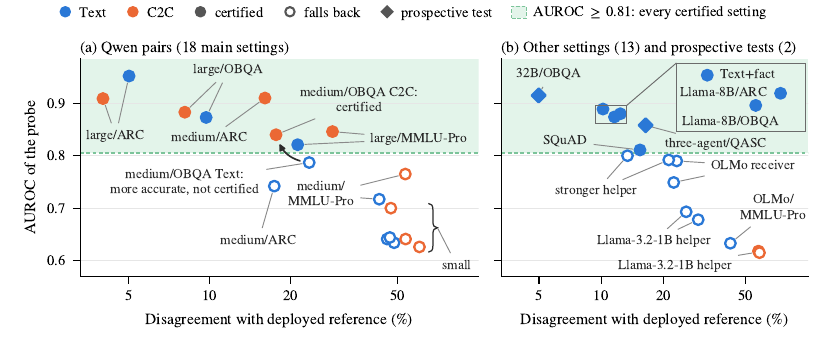}
\caption{Messages that rarely and predictably change the receiver agent's answer can be skipped. Each circle is one of the 31 settings (model pair, reference path, and benchmark); diamonds are the two prospective tests (\S\ref{sec:rq2}), the 32B pair on OBQA development questions (AUROC 0.915) and the three-agent system on its 2,000 sealed QASC test questions (0.858). x-axis: how often the receiver agent alone disagrees with the deployed reference on development questions; y-axis: how well the probe ranks those disagreements (AUROC, \S\ref{sec:setup}; ties count one half). Filled: a threshold is certified and deployed; open: the system falls back to always communicating. The shaded region marks the observed separation: every deployed setting has AUROC at least 0.81, every fallback at most 0.80. Arrow: on OBQA, the medium pair's more accurate Text path is not certified, while C2C is. Nominal certificates: large/OBQA/Text, large/OBQA/C2C, and Text+fact (\S\ref{sec:rq1}). Out-of-sample results, including the one policy that fails under the official C2C extraction (large-pair ARC C2C), are in Table~\ref{tab:e2e}. Inset: three close settings. Text+fact: the helper also sees the OBQA science fact.}
\label{fig:boundary_map}
\end{figure}

\rev{\textbf{Answer to Q2:} how often and how predictably the message changes the answer; accuracy does not decide. \textbf{Which settings are certified.} Of the 18 main settings, 8 are certified and 10 fall back (Figure~\ref{fig:boundary_map}). The large pair can skip both messages on every benchmark, the medium pair only C2C on OBQA and ARC, and the small pair neither. Two quantities explain the pattern: how often a message changes the answer, and how well the probe predicts which answers change. For the small pair, the message changes about half of the receiver agent's answers on OBQA and ARC, and the probe ranks these changes poorly. For the large pair, it changes 4 to 10\% there, and the probe ranks them well (AUROC 0.87 to 0.95). On MMLU-Pro, the large pair's messages change more answers (21.2 and 28.6\%), and the probe still ranks them well (AUROC 0.821 and 0.846). Sample size matters too: a threshold needs at least 135 skipped calibration questions to pass, and the small pair's lowest quantiles hold 18 to 65. The boundary is stable across 200 re-drawn splits but depends on $\alpha$ and the parser (Appendices~\ref{app:ledger} and~\ref{app:parser}).}

\rev{\textbf{What a certificate covers.} A certificate bounds the change rate over all skipped questions together, not within each subgroup. On MMLU-Pro, 12 of 72 skipped engineering answers change under C2C (Appendix~\ref{app:mmlu}). Certifying each discipline separately gives a bound within each one, but lowers the share of development questions skipped from 40.5\% to 11.1\% (Table~\ref{tab:mmlu_disc}). A certificate also presumes that the calibration questions played no part in designing the procedure. Three development certificates are \emph{nominal} in this respect: for large/OBQA/Text, large/OBQA/C2C, and Text+fact, the calibration questions had also been used to choose the probe, so the stated error bound does not apply, and we rely on their held-out tests (\S\ref{sec:rq4}).}

\rev{\textbf{Accuracy does not decide.} The medium pair is certified for C2C on OBQA and ARC but not for Text, the more accurate path. On OBQA, Text changes more of the receiver agent's answers than C2C does (23.5\% against 17.7\%). On ARC the two change about equally many, but the probe ranks C2C's changes far better (AUROC 0.910 against 0.742). The probe reads only the receiver agent, so it cannot tell when a Text message will overturn a confident answer. Adding a term for whether the helper agent agrees favors Text over C2C in every population but certifies no new setting (post hoc; Appendix~\ref{app:predict}). Every C2C path that can be skipped is less accurate than the receiver agent alone (for example, 598 against 617 of 742 on large-pair OBQA). This is a property of the released fusers, whose official outputs we reproduce (Appendix~\ref{app:revision}). For C2C, the certificate therefore identifies redundancy, not usefulness; an operator with labels would simply retire such a path. For medium/ARC/C2C, unparsed C2C answers account for 2 of 4 development changes and 7 of 15 out-of-sample changes.}

\rev{\textbf{Other receiver and helper agents fit the same boundary.} The same pattern appears when either agent changes. We first tried an OLMo-2-7B receiver agent and a Llama-3.2-1B helper agent, and both fell back; for OLMo, part of the reason is answer format, since 4 to 17\% of its calibration answers do not parse. Format is not the whole reason: no deployable parse rule certifies any OLMo setting, and neither does answering from the probe's most likely option, which needs no parser. A Llama-3.1-8B receiver agent that reads the large helper agent's messages is certified on OBQA and ARC and passes on unseen questions (Table~\ref{tab:e2e}; Appendix~\ref{app:xfam}). A stronger helper is not easier to skip: the large helper raises the medium receiver agent's accuracy but changes 21 and 13\% of its answers on OBQA and ARC, and both settings fall back. In the Text+fact setting (post hoc), the large helper also sees the science fact behind each OBQA question. Accuracy rises by 7.1 points, from 617 to 670 of 742. Skipping is still certified, nominally, because the receiver agent alone already gives the path's answer on 90\% of questions and the probe ranks the remaining changes well (AUROC 0.889). The policy answers 660 correctly, 1.35 points [0.54, 2.29] below the reference and within $\kappa\alpha=3.6$ (Appendix~\ref{app:informed}).}

\rev{\textbf{One extractive setting is certified.} To go beyond multiple choice, a label-free pilot scored eight short-answer settings, and a rule fixed before any pilot output selected SQuAD with the Llama-8B pair. The pilot predicted certification with probability 0.975, an estimate inflated by choosing the best of eight. It is certified (Appendix~\ref{app:squad}). On 1,000 new development questions, the receiver agent alone disagrees with Text on 15.4\% (AUROC 0.811), and 4.0\% of skipped answers change. Text adds only 1.0 point of accuracy here [$-0.9$, $+2.9$], so the message is redundant; its saving in Table~\ref{tab:e2e} is estimated, not replayed.}

\subsection{\rev{Where messages help, skipping keeps most of the gain}}\label{sec:rq2}

\begin{table}[t]\centering\footnotesize\revcolor
\caption{\rev{Where Text helps, certified skipping keeps most of its gain on unseen questions. Unseen: questions never used to fix the policy. Changed: changed/skipped answers. Accuracy (\%) of the receiver agent alone ($R$), the always-Text reference, and the certified policy; gains in points with paired-bootstrap 95\% CIs. Kept: (Policy $-R$)/(Text $-R$), point estimate. $\hat\kappa\alpha$: unseen coverage times $\alpha=0.05$, in points, a magnitude comparison for Eq.~\eqref{eq:acc}, not an additional guarantee. Llama-8B/OBQA's lower bound $-0.00$ is $-0.003$. $^\ddagger$Nominal certificate. $^\star$Prospective tests of the label-assisted certificate at $\varepsilon=0.01$, with the procedure fixed before any calibration output (Appendices~\ref{app:e1} and~\ref{app:e2b}); for the three-agent system, $R$ answers without the verifier agent and Text with it.}}
\label{tab:gainkept}
\setlength{\tabcolsep}{2.5pt}
\resizebox{\linewidth}{!}{\begin{tabular}{llccccccccc}\toprule
 & & Skipped & & \multicolumn{3}{c}{Accuracy (\%)} & Text $-R$ & Policy $-R$ & Kept & \\\cmidrule(lr){5-7}
Policy & Unseen questions & (\%) & Changed & $R$ & Text & Policy & [95\% CI] & [95\% CI] & (\%) & $\hat\kappa\alpha$\\\midrule
Llama-8B/OBQA/Text & held out (744) & 59.1 & 3/440 & 80.38 & 83.06 & 82.66 & 2.69 [0.40, 4.97] & 2.28 [$-0.00$, 4.44] & 85 & 2.96\\
Llama-8B/ARC/Text & test split (1,172) & 67.7 & 15/794 & 81.31 & 84.73 & 84.56 & 3.41 [1.71, 5.12] & 3.24 [1.62, 4.86] & 95 & 3.39\\
large/OBQA/Text$^\ddagger$ & held out (744) & 80.0 & 12/595 & 83.06 & 86.96 & 86.02 & 3.90 [1.88, 5.78] & 2.96 [1.21, 4.57] & 76 & 4.00\\
large/OBQA/Text+fact$^\ddagger$ & held out (744) & 75.0 & 16/558 & 83.06 & 91.13 & 89.65 & 8.06 [5.91, 10.22] & 6.59 [4.57, 8.47] & 82 & 3.75\\\midrule
three-agent/QASC$^\star$ & sealed (2,000) & 39.9 & 5/798 & 63.80 & 76.60 & 76.35 & 12.80 [11.30, 14.40] & 12.55 [11.10, 14.15] & 98 & --\\
32B pair/OBQA/Text$^\star$ & held out (744) & 68.4 & 2/509 & 90.32 & 92.88 & 92.88 & 2.55 [0.94, 4.30] & 2.55 [0.94, 4.17] & 100 & --\\\bottomrule
\end{tabular}}
\end{table}

\rev{\textbf{Answer to Q3: mostly not.} Four deployed policies skip a Text message that helps and have unseen questions on which the receiver agent alone also ran (Table~\ref{tab:gainkept}, first four rows; two certificates are nominal). They skip Text on 59 to 80\% of these questions and keep an estimated 76 to 95\% of its gain over the receiver agent. The gain that remains is significantly above zero for three policies; for Llama-8B/OBQA, its interval includes zero. Removing the helper agent altogether would give up the whole gain, 2.7 to 8.1 points (single-model baselines: Appendix~\ref{app:singlemodel}).}

\rev{\textbf{The certificate limits answer changes, not accuracy loss.} Accuracy can change only on skipped questions whose answer changes, so the certificate implies that accuracy moves by at most $\kappa\alpha$, the skip rate times the tolerance (Eq.~\eqref{eq:acc}). For a helpful path, this limit can exceed the gain itself: for large-pair Text on OBQA it is 3.9 points against a gain of 3.8. In practice the change is smaller. Across the eight deployed Qwen-pair policies, development accuracy moves by $-1.34$ to $+0.94$ points (Appendix~\ref{app:ledger}). The retentions in Table~\ref{tab:gainkept} are thus measured, not guaranteed: on unseen questions, $\hat\kappa\alpha$ is about as large as the gain or larger for three of the four policies, and well below it only for Text+fact, whose certificate is nominal. The loss can also be real: on held-out questions, the large/OBQA/Text policy is 0.94 points [0.13, 1.88] less accurate than always sending Text. A helpful path therefore needs a check on accuracy itself.}

\rev{\textbf{Labels on a few skipped disagreements bound the accuracy loss.} Skipping loses accuracy only through a \emph{lost correction}: a skipped question that the reference answers correctly and the receiver agent alone does not. Lost corrections occur only among skipped disagreements, so only those need gold labels (Algorithm~\ref{alg:certify}, step 3). Post hoc, labeling 7 to 12 calibration disagreements per path certified a loss of at most one point for three of the six helpful Text paths (two of them nominal). These policies skip 45 to 62\% of development questions and keep 98 to 106\% of the development gain. On the 744 held-out OBQA questions, each loses at most two corrections (95\% upper bounds 0.75 to 0.97\%; Appendix~\ref{app:accuracy}). Labels do not rescue the medium pair's OBQA Text path, which fails the change-rate test at every threshold. Even run alone, with labels on 36 calibration disagreements, the lost-correction test accepts no threshold at $\varepsilon=0.01$ or $0.02$.}

\rev{\textbf{Two prospective tests pass.} We then tested this label-assisted certificate prospectively, fixing every rule before any calibration output (Table~\ref{tab:gainkept}, last two rows). On a new three-agent system (QASC; \citealp{khot2020qasc}), the policy skips the verifier agent on 39.9\% of 2,000 sealed questions and loses 5 corrections. With a 32B helper and receiver agent on OBQA, it skips the helper agent on 68.4\% of 744 held-out questions and loses one. In both tests, the 95\% upper bound on the share of test questions with a lost correction is below the prespecified 1\% (0.58\% and 0.75\%).}

\subsection{Skipping pays when the message is slow}\label{sec:rq3}

\input{tab_main_economics.tex}

\rev{\textbf{Answer to Q4:} only if the time saved on skips exceeds the probe's cost, paid on every query. \textbf{Savings follow the cost of producing the message.} Every certified policy on OBQA and ARC saves time (Table~\ref{tab:e2e}). A Text message must be generated by the helper agent, and skipping it saves 319 to 718~ms per query. A C2C message needs only a prefill and a projection, and skipping it saves 37 to 111~ms. Batching keeps the saving: in static batches of 1, 8, and 32, with skipped requests returning early, the four deployed Text policies lower mean latency by 35 to 67\% (Appendix~\ref{app:e2a}).}

\rev{\textbf{Every query pays for the probe.} The probe runs on every query, so a policy that skips few queries or saves little per skip can lose time (Eq.~\eqref{eq:saving}). MMLU-Pro shows the limit. Both large-pair policies skip the same 48 of 128 panel questions, a skip rate 3 points below the full development set's. Skipping Text still saves time, but for C2C the time saved barely covers the 46.1~ms probe, and the panel does not establish a saving. Most queries pay for both the probe and the message, so the median saving is negative for both policies. Over all 2,641 development questions, replayed on the second configuration, the C2C saving is small but positive: 23.3~ms [14.5, 32.2]. Three repeats on the second configuration left every policy's saving class (established or not) unchanged, and no policy's 99th-percentile latency exceeds its reference's (Appendix~\ref{app:e2e}).}

\rev{\textbf{Answering from the probe.} Our policy runs the receiver agent again after the probe, so a skipped query returns its usual answer. Returning the probe's most likely option instead avoids the rerun, but that answer rule needs its own certificate. Recertification keeps every threshold but one (medium/OBQA/C2C needs a lower one). In one replay on the second configuration, it raised the MMLU-Pro C2C panel saving from 4.6 to 85.0~ms (Appendix~\ref{app:revision}).}

\subsection{The tested policies hold on unseen questions}\label{sec:rq4}

\rev{\textbf{Sealed and held-out tests (Q2, out of sample).} A certificate is a statement about new questions, so we test every deployed policy we can on questions it has not seen. Each frozen policy ran once, with routes and outputs saved before the gold labels were decoded. In all eleven tests, the 95\% upper bound on the change rate stays below 5\% under our answer parsers (Table~\ref{tab:e2e}; Appendix~\ref{app:sealed}). On the 1,172 sealed ARC-Challenge test questions, the two large-pair policies change under 2\% of skipped answers and run faster than their references. One result depends on the parser: under the official C2C extraction, the large-pair ARC C2C policy changes 5.2\% [3.9, 6.7]. All 44 outputs that the two parsers read differently are an option letter followed by that option's text, which the official extraction rejects for strings such as \texttt{tan}. The Text result is identical under both.}

\rev{The other nine tests are listed in Table~\ref{tab:e2e}. For large/OBQA/Text, large/OBQA/C2C, and Text+fact, whose certificates are nominal, the held-out OBQA test is the primary evidence. Seven of the eleven tests also reject a change rate of 5\% or more at level 0.001; the others, including Text+fact, reject only at 0.025. Pass rules were prespecified only for the four latest tests: the two Llama-8B policies, medium/ARC/C2C, and SQuAD. The two MMLU-Pro policies have no unused questions and are untested.}

%% file: tab_main_economics.tex
\begin{table}[t]\centering\footnotesize
\caption{Every deployed policy tested on unseen questions keeps the change rate's 95\% upper bound below 5\% under our parsers; skipping Text saves hundreds of milliseconds per query, skipping C2C much less. Skipped: \% of development questions answered by the receiver agent alone. Changed: changed skipped answers on development and on sealed or held-out (unseen) questions. Level: the test rejects $\rho\ge0.05$ at 0.001, or only the upper bound is below 5\% (0.025). $\Delta$acc: policy minus reference accuracy on unseen questions (SQuAD: point estimate). Gain ($G$): development reference minus receiver-alone accuracy; Gain Kept: \% of it retained (--: $G\le0$). Saving: paired mean per query on 128-question panels, cold requests included. $^\ddagger$Nominal certificate. $^{\circ}$Pass rule prespecified. $^{\S}$Official C2C extraction: 54/1,044. $^{*}$Certified only at a lower quantile $q$ without the parser-review questions. $^{b}$Second hardware configuration. $^{e}$Estimated, not replayed. Full results: Tables~\ref{tab:e2e_full} and~\ref{tab:dev_full}.}
\label{tab:e2e}
\setlength{\tabcolsep}{3pt}
\resizebox{\linewidth}{!}{\begin{tabular}{lllcrlcrccl}\toprule
 & & & & \multicolumn{2}{c}{Changed} & & \multicolumn{1}{c}{$\Delta$acc (pp)} & & & \multicolumn{1}{c}{Saving (ms)}\\\cmidrule(lr){5-6}
Pair & Task & Path & Skipped (\%) & \multicolumn{1}{c}{Dev.} & \multicolumn{1}{c}{Unseen} & Level & \multicolumn{1}{c}{[95\% CI]} & Gain (pp) & Gain Kept (\%) & \multicolumn{1}{c}{[95\% CI]}\\\midrule
medium & OBQA & C2C$^{*}$ & 52.6 & 11/390 & held out 8/388 & 0.025 & $+0.27$ [$-0.54$, $+1.08$] & $-0.5$ & -- & 79.7 [48.1, 129.1]\\
medium & ARC & C2C & 63.2 & 4/189 & held out 15/733$^{\circ}$ & 0.001 & $+0.34$ [$-0.26$, $+0.94$] & $-5.0$ & -- & 100.7 [74.5, 133.1]\\\midrule
large & OBQA & Text$^\ddagger$ & 77.1 & 19/572 & held out 12/595 & 0.001 & $-0.94$ [$-1.88$, $-0.13$] & 3.8 & 75 & 517.1 [447.8, 583.6]\\
large & OBQA & C2C$^\ddagger$ & 77.1 & 13/572 & held out 13/595 & 0.001 & $+1.21$ [$+0.40$, $+2.15$] & $-2.6$ & -- & 36.6 [18.0, 56.8]\\
large & ARC & Text & 95.0 & 9/284 & sealed 20/1,100 & 0.001 & $-0.43$ [$-1.19$, $+0.26$] & 2.0 & 33 & 717.6 [670.6, 763.2]\\
large & ARC & C2C & 90.3 & 3/270 & sealed 11/1,047$^{\S}$ & 0.001 & $+0.34$ [$-0.17$, $+0.85$] & $-0.7$ & -- & 110.9 [82.0, 142.8]\\
large & MMLU-Pro & Text & 40.5 & 40/1,069 & none & -- & -- & 1.4 & 105 & 346.3 [249.8, 443.0]\\
large & MMLU-Pro & C2C & 40.5 & 37/1,069 & none & -- & -- & $-1.9$ & -- & 8.2 [$-20.5$, $+44.0$]\\
large & OBQA & Text+fact$^\ddagger$ & 72.2 & 13/536 & held out 16/558 & 0.025 & $-1.48$ [$-2.55$, $-0.54$] & 7.1 & 81 & 343.4 [297.7, 386.0]$^{b}$\\\midrule
Llama-8B & OBQA & Text$^{*}$ & 59.0 & 7/438 & held out 3/440$^{\circ}$ & 0.001 & $-0.40$ [$-0.94$, 0.00] & 4.4 & 106 & 318.6 [261.1, 372.1]$^{b}$\\
Llama-8B & ARC & Text$^{*}$ & 68.9 & 4/206 & held out 15/794$^{\circ}$ & 0.001 & $-0.17$ [$-0.77$, $+0.43$] & 7.0 & 90 & 418.9 [361.7, 473.7]$^{b}$\\
Llama-8B & SQuAD & Text & 39.6 & 16/396 & sealed 9/364$^{\circ}$ & 0.025 & $+0.4$ & 1.0 & 140 & 198.5 [167.2, 229.9]$^{e}$\\\bottomrule
\end{tabular}}
\end{table}

%% file: sec_discussion.tex

%% file: sec_related.tex
\section{Related work}\label{sec:related}

\textbf{Communication, routing, and efficiency.} Agents communicate in natural language \citep{du2024debate,li2023camel} or through internal states \citep{fu2025c2c,heo2026crossmodelkv,zou2025latentmas}; learned gates \citep{singh2019ic3net} and AgentPrune \citep{zhang2025agentprune} drop messages to improve task performance. Systems frequently trade quality against cost through routers \citep{ding2024hybrid,ong2024routellm}, cascades \citep{chen2023frugalgpt,yue2024mot,gupta2024cascades}, and adaptive retrieval \citep{mallen2023trust,jiang2023flare}. These choices are often motivated by oracle headroom \citep{jiang2023blender,hu2024routerbench}. We show that it is weak evidence of communication utility, because inputs without answer-specific content reproduce comparable headroom (\S\ref{sec:complementarity}).

\textbf{Statistical risk control.} Finite-sample risk is bounded in selective classification \citep{geifman2017selective}, Learn then Test \citep{angelopoulos2022ltt}, and conformal risk control \citep{angelopoulos2024crc,pona2026ctd}. Our target is closest to the consistency risk of CALM \citep{schuster2022calm} and risk-controlled early exiting \citep{jazbec2024fastyetsafe}, whose \rev{risk-control framework} we share. \citet{uddin2026conformal} certify LLM routing with a Clopper--Pearson test and find routability model- and task-dependent, and \citet{uddin2026proactive} give a population condition for it that holds in every multiple-choice setting certified here and in eight that fall back (Appendix~\ref{app:predict}). Both target a routed answer's correctness; we certify agreement with a deployed path that may be the less accurate one.

\rev{\textbf{What changes when the skipped computation is a message.} We claim no new test, and neither guarantee requires the reference to be the more accurate one. CALM bounds the expected inconsistency with the full model over all inputs; we bound the change rate among skipped questions, with an exact binomial test per candidate threshold. The gates read different states. Within each decoding step, an early-exit gate reads intermediate states of the computation it may cut short; our gate reads a separate message-free prefill of the receiver agent, which in our implementation adds a forward pass to every query, so we measure whether skipping pays (\S\ref{sec:rq3}; Appendix~\ref{app:related}). The message comes from another agent: it may carry no answer-specific content (\S\ref{sec:complementarity}) or make the system less accurate (\S\ref{sec:rq1}), so we measure agreement with the reference and retention of its gain separately (\S\ref{sec:rq2}).}

%% file: sec_limitations.tex
\section{Discussion and limitations}\label{sec:limitations}\label{sec:discussion}

\rev{\textbf{Practical value and predictability.} The certificate needs only unlabeled traffic. Without labels, always skipping would be a blind change, since answering alone fails the test in every setting. Whether a setting will certify can be anticipated in two ways. A binormal model of the scores reproduces all 30 earlier certification outcomes from the disagreement rate, the AUROC, and the calibration size. A label-free plug-in rule applies the test to fit-split counts rescaled to the calibration size and reproduces all 25 outcomes it covers (not an independent check). Among ten later settings, an AUROC cut near 0.80, fixed in advance, got one wrong, the stronger helper agent on ARC, and the plug-in rule predicted it (Appendix~\ref{app:predict}).}

\rev{\textbf{Scope and failure modes.} The main experiments study two agents that read the same multiple-choice question and exchange one message. Three extensions go beyond this: one extractive setting (SQuAD, chosen from eight; an existence result), Text+fact, where the helper agent has information the receiver agent lacks, and a three-agent pipeline (Appendix~\ref{app:e1}). No setting uses tools or several rounds. Receiver agents have at most 8B parameters in the main settings; a 32B pair repeats the label-assisted certificate on OBQA (Appendix~\ref{app:e2b}). We have not run closed models: a closed helper agent fits once the policy is recalibrated, a closed receiver agent must expose option-label probabilities, and C2C needs the helper agent's KV cache. The method does not extend to GSM8K, where a prespecified check fell back. Text changes 19.3\% of development answers there, but the score ranks these changes poorly (AUROC 0.58). The receiver agent restates a number it has already derived, so its final answer carries no residual uncertainty, and such tasks need a score read before the answer is committed (Appendix~\ref{app:gsm}). We decode greedily without reasoning tokens; sampling at temperature 0.7 keeps large-pair Text skipping certified on OBQA (nominally) and ARC (Appendix~\ref{app:sampling}). Latency is measured at batch size one, the only mode the C2C release supports; in static batches of up to 32, skipping Text still lowers mean latency (\S\ref{sec:rq3}).}

\textbf{Exposure.} The score, $\alpha$, $\delta$, grid, and parser were fixed after OBQA and ARC development accuracy was seen, and ProbeMax was chosen on large/OBQA/Text calibration labels (Appendix~\ref{app:provenance}). The parser was written after reviewing 341 later calibration questions with gold labels, but an earlier one written without them gives identical labels in every certified Qwen setting. An earlier project stage scored small-pair outputs on the held-out OBQA questions, so they are not sealed (Appendix~\ref{app:sealed}).

\rev{\textbf{Reliance on zero-uncertainty predictions.} The Qwen certificates rest on questions where the probe reports zero uncertainty. For Qwen receiver agents, 60 to 92\% of skipped development questions have $u=0$, and on questions with $u>0$ alone the test certifies only 1 of 9 settings. Neither recomputing the probe in FP32 nor breaking ties by the log of the non-top probability mass changes a certification decision in any of 23 settings. Zeros are not required, however: the Llama-8B receiver agent has none, and its policies are certified and hold on unseen questions (Appendix~\ref{app:restson}). Part of this confidence follows option position: when the options are rotated, the receiver agent's answer changes on 1.8 to 6.3\% of development questions with $u=0$. The certificate therefore covers the deployed prompt format. Under rotation, the five frozen large-pair policies still change 1.7 to 2.6\% of skipped sealed and held-out answers, with upper bounds below 5\% (Appendix~\ref{app:sealed}).} \rev{Pretraining exposure does not by itself invalidate the certificate, which assumes that deployment queries come from the calibration distribution; it matters when that assumption fails, for instance when the receiver agent has seen the benchmark questions but not the deployment traffic. Our OBQA splits come from its official training split, so an operator should recertify on its own traffic. The Llama-8B result does not rule out contamination.}

%% file: sec_conclusion.tex
\section{Conclusion}\label{sec:conclusion}

A receiver agent often does not need the message a helper agent sends. A lightweight probe and a finite-sample test find these queries and bound how often skipping alters the answer. In the two-agent systems we study, on multiple-choice questions and one extractive task, headroom, redundancy, and net latency value are distinct properties of a communication path. Oracle headroom is weak evidence of useful content: control inputs without answer-specific content reproduce about three quarters of the oracle's gain over the receiver agent. Whether a message can be skipped follows how often and how predictably it changes the answer, not its accuracy. \rev{In four settings where Text helps, skipping kept an estimated 76 to 95\% of its gain on unseen questions.} Post hoc, labels on a few skipped disagreements bounded the accuracy loss of three helpful Text paths (two nominally) by one point\rev{; two prospective tests (three agents; a 32B pair) passed}. At batch size one, skipping Text saves a mean of hundreds of milliseconds per query, whereas skipping C2C saves far less unless the receiver agent answers from the probe. The eleven frozen out-of-sample tests of \S\ref{sec:rq4} kept the change rate's 95\% upper bound below 5\% under our answer parsers (a pass rule was prespecified for four).

%% file: sec_statements.tex
\subsection*{Acknowledgments}
This research used resources of the Argonne Leadership Computing Facility, a U.S. Department of Energy (DOE) Office of Science user facility at Argonne National Laboratory (ANL) operated under Contract No.~DE-AC02-06CH11357. The work was also supported under the same contract by the DOE Office of Science's Advanced Scientific Computing Research Program and by Laboratory Directed Research and Development (LDRD) funding from ANL, provided by the Director, DOE Office of Science. The work of the University of Houston coauthors was supported in part by the National Science Foundation under grants CAREER-2305491 and CNS-2504762.

\subsection*{AI use statement}
In this work, we used generative AI tools for three purposes: to edit the manuscript for readability and polish its wording; to search for and identify relevant literature and to format references; and to assist with data processing and analysis, including drafting and refining the code for these tasks and for the figures. All AI-generated text and code were reviewed by the authors, and LLM-generated code was verified and tested for correctness by the authors. We have not used generative AI tools to develop the theoretical framework, formulate mathematical claims or proofs, propose hypotheses, design the methodology or experiments, or interpret results; generating synthetic data, translation, and qualitative data analysis are not applicable to this work. We take responsibility for the final content of this work, including text, claims or artifacts produced with the aid of generative AI.

\subsection*{Reproducibility statement}
Sections~\ref{sec:problem} and~\ref{sec:setup} and Appendix~\ref{app:protocol} specify the score, label tokens, splits, threshold grid, and acceptance rule, and Appendix~\ref{app:prompts} gives every prompt verbatim and the answer parser's rules. Section~\ref{sec:setup} and Appendices~\ref{app:protocol}, \ref{app:medium}, and~\ref{app:e2e} specify models, decoding, references, the timing boundary, freeze order, and statistical procedures; Appendix~\ref{app:ledger} summarizes the development results of all settings and their sensitivity, Appendix~\ref{app:sealed} the out-of-sample tests, and Appendix~\ref{app:revision} the port check and repeated replays. The code repository (\url{https://github.com/Piedpiperrr/safe-to-omit}) contains the code, frozen configurations, exact partition and question IDs, model revisions, the parser's exact expressions, scripts that regenerate every figure, and a list of the details that the appendix leaves out for length. The released dataset (\url{https://huggingface.co/datasets/AgentsSci/ICLR2027-SafeToOmit}) contains the per-question confidence scores and routes, raw action outputs, per-request timing records, and every calibration test, including rejected grid points and fallback settings. Model weights and full-vocabulary logits are not redistributed; all models and datasets are publicly available.

%% file: new_app_A.tex
\renewcommand{\topfraction}{0.95}\renewcommand{\bottomfraction}{0.8}
\renewcommand{\textfraction}{0.05}\renewcommand{\floatpagefraction}{0.9}
\setcounter{topnumber}{4}\setcounter{bottomnumber}{3}\setcounter{totalnumber}{6}
\setlength{\textfloatsep}{8pt plus 2pt minus 4pt}\setlength{\floatsep}{8pt plus 2pt minus 2pt}\setlength{\intextsep}{8pt plus 2pt minus 2pt}
\providecommand{\fitbox}[1]{\setbox0=\hbox{#1}\ifdim\wd0>\linewidth\resizebox{\linewidth}{!}{\box0}\else\box0\fi}

\section{Method and protocol}\label{app:protocol}\label{app:details}

Appendices~\ref{app:protocol}--\ref{app:prompts} give the records of every setting and check. The code repository holds the code and configurations, and the released dataset holds the per-question records. As in the main text, a policy \emph{skips} the message when the receiver answers alone, and ``helper'' and ``receiver'' abbreviate the two agents. Every setting uses one frozen procedure: a probe score from one receiver prefill, 20 thresholds from the fit split, and one exact binomial test per threshold on the calibration split (parser exposure: Appendix~\ref{app:parser}; what the certificates rest on: Appendix~\ref{app:restson}).

\rev{\paragraph{Roadmap.} Each appendix supports one or two of the four questions of Section~\ref{sec:results} (Q1: does high oracle headroom show that a message is needed; Q2: what makes a message safe to skip; Q3: when a message helps, does skipping lose its gain; Q4: does a safe skip save time) or documents the method.}

{\revcolor\begin{center}\footnotesize
\begin{tabular}{@{}llp{0.62\linewidth}@{}}\toprule
Appendix & Supports & What it shows\\\midrule
\ref{app:protocol} & method & Terms and notation, score, splits, threshold grid, test, and decoding.\\
\ref{app:provenance} & method & Which protocols were fixed before which outcomes; parser exposure.\\
\ref{app:complementarity} & Q1 & The content-free null and its matched designs.\\
\ref{app:ledger} & Q2 & Every setting's test, re-drawn splits, and per-discipline MMLU-Pro certification.\\
\ref{app:extensions} & Q2, Q3 & Cross-family agents, an informed helper (Text+fact), SQuAD, GSM8K, and sampling.\\
\ref{app:sealed} & Q2 & The eleven sealed and held-out tests of \S\ref{sec:rq4} and the option-rotation check.\\
\ref{app:restson} & Q2 & What the certificates assume and where exposure remains.\\
\ref{app:predict} & Q2 & A label-free rule that predicts which settings certify.\\
\ref{app:accuracy} & Q3 & Gain retention, the label-assisted loss certificate, and single-model baselines.\\
\ref{app:prospective} & Q3, Q4 & Prospective tests on a three-agent system (QASC; sealed test) and a 32B helper and receiver pair (OBQA; held out), with latency.\\
\ref{app:e2e} & Q4 & End-to-end latency replays, the second hardware configuration, and batched serving.\\
\ref{app:related} & method & Relation to risk-controlled early exiting and routing.\\
\ref{app:prompts} & method & Every prompt and the answer parser.\\\bottomrule
\end{tabular}\end{center}}

\rev{\paragraph{Terms and notation.} The table collects the terms the paper uses, with the section that introduces each.}

{\revcolor\footnotesize\setlength{\LTpre}{4pt}\setlength{\LTpost}{4pt}
\begin{longtable}{@{}p{0.25\linewidth}p{0.72\linewidth}@{}}\toprule
Term & Meaning\\\midrule
Reference path $b$ & The deployed system that always sends the message (Text or C2C). Skipping is judged against its answer $o_b(x)$ (\S\ref{sec:problem}).\\
Receiver alone $R$ & The receiver agent answering without the message, with answer $o_R(x)$. In the three-agent system (Appendix~\ref{app:e1}), the pipeline without the verifier agent.\\
Score $s(x)=u(x)$ & The receiver agent's uncertainty from one forward pass, the probe. The main multiple-choice settings use ProbeMax, one minus the largest option probability (\S\ref{sec:setup}); SQuAD and Appendices~\ref{app:e1} and~\ref{app:e2b} use variants defined there.\\
Threshold $\tau$, quantile $q$ & The message is skipped when $s(x)\le\tau$. Candidate thresholds $\tau_q$ are quantiles of the fit-split scores.\\
Coverage $\kappa$ & The fraction of queries that skip the message.\\
Change rate $\rho$ & The probability that a skipped answer differs from the reference's answer (Eq.~\eqref{eq:rho}).\\
$\alpha$, $\delta$ & Tolerance on the change rate (0.05) and level of each test (0.001). With 20 candidates, the error bound is 0.02 per setting.\\
$\kappa\alpha$ & The largest accuracy change the certificate allows (Eq.~\eqref{eq:acc}).\\
$\varepsilon$ & The accuracy loss certified by the label-assisted test (0.01 or 0.02).\\
Lost correction & A skipped question that the reference answers correctly and the receiver agent alone does not.\\
Certified, fallback & A threshold passes the test and is deployed; or none passes and the system always communicates.\\
Nominal certificate & The threshold passes the test, but its error bound is not established because the calibration questions were also used to choose the probe. The held-out test is the evidence (\S\ref{sec:rq4}).\\
Post hoc, prospective & An analysis specified after its data were seen (descriptive); or a test whose every rule was fixed before any calibration output.\\
Development, held out, sealed & Data observed during the study; never used by the pair tested; untouched until a frozen policy ran (\S\ref{sec:setup}).\\
Setting, population & A model pair, benchmark, and reference path; a model pair on one benchmark.\\
Oracle headroom & Accuracy of an oracle that picks a correct path per question, minus that of the best single path (\S\ref{sec:intro}).\\
Gain, gain kept & Reference minus receiver-alone accuracy; the share of it the policy retains.\\\bottomrule
\end{longtable}\addtocounter{table}{-1}}

\paragraph{Decision, answer change, and parsing.} Each setting is a binary choice between receiver-only $R$ and one prespecified reference $b$ (Text or C2C); no Text/C2C selector is learned, and C2C is each pair's official fuser, not one of the latent protocols of earlier project stages. An answer change is $d_b=\mathbb{I}[o_R\neq o_b]$ on frozen parsed outputs. The rule-based parser reads only the raw output and displayed labels (never gold labels, IDs, or option texts), ignoring quoted, hedged, conditional, and math-like contexts (rules: Appendix~\ref{app:prompts}). Unparseable outputs share one \textsc{invalid} label (two \textsc{invalid} outputs agree, and each is scored as incorrect). Runtime failures are recorded separately, and gold labels are joined only after routes are frozen.

\paragraph{Models and prompts.} All models run in bfloat16 with SDPA attention, greedily, with thinking disabled, at batch size one. C2C uses each pair's official pretrained fuser, and, as in the official pipeline, the helper runs on the receiver's token IDs so that transferred cache positions align. The Text helper writes at most 256 tokens and the receiver at most 64 (prompts: Appendix~\ref{app:prompts}). No two paths share a helper message, KV cache, or prefill.

\paragraph{Score.} ProbeMax uses one receiver prefill of the receiver-alone prompt followed by \texttt{The correct answer is}. We renormalize (in FP32) the predicted probabilities of the question's $K$ displayed labels (A--J on MMLU-Pro), and define the confidence score as $s(x)=u=1-\max_l p_l$. A label's probability sums over the non-special single tokens whose decoding, stripped of whitespace, equals the displayed label. The probe's argmax is not $R$'s answer $o_R$ and serves only as a diagnostic that affects no decision, although it matches $o_R$ on 740 of 742 (large/OBQA), 299 of 299 (large/ARC), 720 of 742 (medium/OBQA), 294 of 299 (medium/ARC), 711 of 742 (small/OBQA), and 264 of 299 (small/ARC) development questions.

\paragraph{Grid, acceptance, and ties.} Candidate $j=1,\ldots,19$ ($q=j/20$) is the $\lceil jN_{\rm fit}/20\rceil$-th smallest fit-split score (ties included; $q=1$ is fixed $R$), frozen before reading calibration data. A threshold is accepted if $\Pr[\mathrm{Bin}(n_q,0.05)\le k_q]\le 0.001$ on its $n_q$ skipped calibration groups, of which $k_q$ changed. We deploy the largest accepted candidate $q$; otherwise, the system falls back to fixed $b$. Clopper--Pearson $0.999$ bounds are diagnostics. Exact zero uncertainties are common (large receiver: 354 of 742 OBQA and 217 of 299 ARC development questions; small: none). Equality routes to $R$, and ties only make neighboring candidates identical, because the test uses only $n_q$ and $k_q$; every AUROC credits ties one half.

\paragraph{Families and assumptions.} Testing 20 predetermined candidates per setting bounds by $0.02$ (Bonferroni) the chance of deploying a threshold whose true change rate is $\ge\alpha$, assuming i.i.d.\ calibration groups and that the score, thresholds, and parser were fixed independently of their outcomes. Each extension is its own family (e.g., 80 tests for the medium pair and 40 for large-pair MMLU-Pro, ideal bounds 0.08 and 0.04). There is no joint guarantee across settings or for subgroups, and the records do not establish a guarantee covering the full selection process. Of the 160 small- and large-pair tests, the 20 of large/OBQA/Text were reused from an earlier 100-test family and carry only that family's nominal $0.10$ bound. The other 140 tests are new. The large/OBQA/C2C and Text+fact certificates are also nominal (Appendix~\ref{app:history}).

\paragraph{Splits.} Fit, calibration, and development hold 2,100, 1,366, and 742 OBQA questions (a partition, inherited from an earlier project stage, of the first 4,208 of the 4,952 deduplicated training questions in a fixed pseudorandom order, drawn without using labels; the other 744 are held out; all development evidence), 670, 448, and 299 ARC groups (1,119 training rows in 1,118 groups, and the validation split; the 1,172-question test split stayed sealed), and 3,000, 6,000, and 2,641 MMLU-Pro groups (all 12,032 test rows in 14 categories with 3 to 10 options, grouped by normalized text; all exposed development evidence). Because ARC's calibration and development sets originate from different official splits, we compared them under a rule fixed before computing: calibration disagreement is lower in 5 of 10 settings, the Mantel--Haenszel odds ratio (development vs.\ calibration) is 0.98 [0.87, 1.10], and a permutation test gives $p=0.82$.

\paragraph{Timing boundary.} Unless marked as the second configuration (Appendix~\ref{app:revision}), latency is synchronized wall-clock time per question from input preparation to parsed answer on two A100 40GB GPUs of a shared node, with the helper on one GPU and the receiver and fuser on the other (startup excluded; no warmup, retry, or probe reuse; cold requests kept). Each policy runs interleaved with its fixed reference on a fixed 128-question development panel, taking the first questions in a fixed pseudorandom order of question IDs (stratified by category on MMLU-Pro). Fallback settings run no probe.

\paragraph{Statistics.} Answer changes among skipped questions are reported as counts. Paired differences in accuracy and latency use 2,000 seed-0 paired bootstrap resamples over question groups. Change-rate intervals are exact two-sided 95\% Clopper--Pearson intervals. All intervals are descriptive: they exclude model-training uncertainty and support neither equivalence nor noninferiority claims.

%% file: new_app_B.tex
\section{Provenance and exposure}\label{app:provenance}

Different parts of this paper rest on evidence of varying strength. Table~\ref{tab:provenance} specifies when each protocol was fixed relative to its reported outcomes. \emph{Sealed} data remained untouched until the frozen policy was evaluated. \emph{Held out} data were never used by the tested model pair, although another pair or an earlier project stage had used them. \emph{Development} data were observed during the study. A protocol \emph{frozen before outcomes} was written, with a commitment to report every outcome, before any output it covers existed; this does not imply public preregistration. \emph{Post hoc} denotes analyses designed after observing the results.

\begin{table}[!htbp]\centering\footnotesize
\caption{Most evidence rests on protocols fixed before the outcomes they cover, and exposed entries say so. Every certification outcome uses development data. \emph{Before outputs}: written, with a commitment to report every outcome, before any output it covers existed. \emph{Before computing}: rules fixed on stored outputs before any number was computed. \emph{Post hoc}: designed after observing the results. \emph{Exposed}: gold labels or scored outputs had been read before the rule or test was fixed. \emph{Gate amended}: a gate found earlier use, so the questions count as held out. \emph{Pass rule}: a 95\% upper bound below 5\%. $^\ddagger$Nominal certificate: the calibration labels are exposed or coupled to the score choice (Appendix~\ref{app:history}). Appendix: where the full record is.}
\label{tab:provenance}
\renewcommand{\arraystretch}{1.2}\setlength{\tabcolsep}{4.5pt}
{\itshape (a) Certification of settings and out-of-sample tests}\par\smallskip
\begin{tabular}{@{}p{3.0cm}p{5.35cm}p{4.05cm}r@{}}\toprule
Evidence & Settings or policies & Protocol fixed & \llap{Appendix}\tabularnewline\midrule
\multicolumn{4}{@{}l}{\emph{Certification of settings}}\tabularnewline\addlinespace[2pt]
\raggedright Original settings & \raggedright small and large pairs, OBQA and ARC (6) & \raggedright before calibration statistics & \ref{app:protocol}, \ref{app:ledger}\tabularnewline
\raggedright Reused certification$^\ddagger$ & \raggedright large/OBQA/Text & \raggedright earlier 100-test family; exposed & \ref{app:history}\tabularnewline
\raggedright Coupled labels$^\ddagger$ & \raggedright large/OBQA/C2C & \raggedright before calibration statistics & \ref{app:history}\tabularnewline
\raggedright Medium pair & \raggedright 4 settings & \raggedright before outputs & \ref{app:medium}, \ref{app:ledger}\tabularnewline
\raggedright MMLU-Pro & \raggedright large (2), then small and medium pairs (4) & \raggedright before outputs & \ref{app:mmlu}\tabularnewline
\raggedright Informed helper$^\ddagger$ & \raggedright large/OBQA/Text+fact & \raggedright before outputs & \ref{app:informed}\tabularnewline
\raggedright Cross-family pairs & \raggedright OLMo and Llama pairs (7) & \raggedright before outputs & \ref{app:xfam}\tabularnewline
\raggedright Llama-8B receiver & \raggedright 2 settings, added after the OLMo and Llama pairs fell back & \raggedright before outputs; post hoc & \ref{app:xfam}\tabularnewline
\raggedright Stronger helper & \raggedright medium receiver, OBQA and ARC Text & \raggedright before outputs; post hoc & \ref{app:informed}\tabularnewline
\raggedright Sampled answers & \raggedright large pair, OBQA and ARC Text & \raggedright before outputs; post hoc & \ref{app:sampling}\tabularnewline
\raggedright Numeric-answer check & \raggedright large/GSM8K/Text & \raggedright before outputs & \ref{app:gsm}\tabularnewline
\raggedright Derivation-prefix score & \raggedright large/GSM8K/Text; not completed & \raggedright before prefix scores; post hoc & \ref{app:gsm}\tabularnewline
\raggedright SQuAD & \raggedright Llama-8B pair (1), chosen label-free from 8 pilot settings after the GSM8K fallback & \raggedright before outputs; selection rule before any pilot output & \ref{app:squad}\tabularnewline
\midrule
\multicolumn{4}{@{}l}{\emph{Out-of-sample tests}}\tabularnewline\addlinespace[2pt]
\raggedright Sealed ARC test & \raggedright large/ARC, 2 policies & \raggedright before test access & \ref{app:sealed}\tabularnewline
\raggedright Held-out OBQA test & \raggedright 4 policies, 5 tests & \raggedright before outputs; gate amended & \ref{app:sealed}\tabularnewline
\raggedright Later held-out tests & \raggedright Llama-8B (2), medium/ARC/C2C & \raggedright before outputs, with a pass rule & \ref{app:sealed}\tabularnewline
\raggedright Sealed SQuAD test & \raggedright 1 policy & \raggedright before outputs, with a pass rule & \ref{app:squad}\tabularnewline
\raggedright Option rotation & \raggedright 5 large-pair policies; post hoc & \raggedright before any rotated output & \ref{app:sealed}\tabularnewline
\bottomrule
\end{tabular}

\bigskip
{\itshape (b) Supporting analyses, by when their rules were fixed}\par\smallskip
\begin{tabular}{@{}p{3.0cm}p{9.05cm}r@{}}\toprule
Rules fixed & Analysis & Appendix\tabularnewline\midrule
Before any output & end-to-end replays and second-hardware repeats & \ref{app:e2e}\tabularnewline
 & probe precision and ties & \ref{app:restson}\tabularnewline
 & helper-aware score, as a hypothesis & \ref{app:predict}\tabularnewline\addlinespace[3pt]
Before computing & null control, null headroom, and stratified null & \ref{app:complementarity}\tabularnewline
 & argmax recertification & \ref{app:xfam}\tabularnewline
 & gain retention and tolerance grid & \ref{app:accuracy}\tabularnewline
 & busy GPU time and panel representativeness & \ref{app:e2e}\tabularnewline
 & sealed official extraction & \ref{app:sealed}\tabularnewline
 & zero-uncertainty split, saving by source, and unparseable changes & \ref{app:restson}\tabularnewline
 & small-sample check and fit-split prediction & \ref{app:predict}\tabularnewline\addlinespace[3pt]
Before computing; & label-light loss certificate, on calibration labels read earlier & \ref{app:accuracy}\tabularnewline
post hoc & unsaturated AUROC and receiver rotation & \ref{app:restson}\tabularnewline\addlinespace[3pt]
Post hoc & test and score variants, and re-split stability & \ref{app:ledger}\tabularnewline
 & port check on the 500 public OBQA test questions & \ref{app:revision}\tabularnewline\addlinespace[3pt]
Exposed & answer parser, written after reviewing 341 later calibration questions & \ref{app:parser}\tabularnewline
\bottomrule
\end{tabular}
\end{table}

Four entries carry known exposure. First, the reused large/OBQA/Text certification shares its calibration labels with 440 other tests (460 total, 360 outside its declared 100-test family). Consequently, its family bound is nominal rather than established. Because large/OBQA/C2C and Text+fact reuse these calibration questions and receiver outputs, and Text and C2C change the same development questions much more frequently than independent changes would (17 against 5.8; Appendix~\ref{app:complementarity}), all three bounds are nominal and rest on their held-out tests instead. Second, the parser's rules were formulated after reviewing raw outputs and gold labels for 341 questions that later entered calibration splits. However, an earlier parser developed without gold labels yields identical calibration labels for every certified Qwen setting. Removing the reviewed questions leaves all deployed thresholds certified, except medium/OBQA/C2C (which passes at $q=0.50$) and the two later Llama-8B policies (which pass at $0.50$ and $0.55$). Third, the held-out OBQA test is the only place where a protocol was amended: after a gate detected prior use of the questions by an earlier project stage, the run was stopped. The amendment, written before any output existed, reports them as held out rather than sealed (Appendix~\ref{app:sealed}). Finally, the label-light loss certificate reuses calibration labels read earlier, and a lost-correction count for Text+fact existed prior to its protocol, which stated that none was known (Appendix~\ref{app:accuracy}).

\paragraph{The reused large/OBQA/Text certification.}\label{app:history} The 20 tests for large/OBQA/Text originated from an earlier stage declaring a 100-test family: 60 new tests using ProbeMax, a probe-entropy score, and a word-feature disagreement classifier, and 40 reused tests, whose outcomes were already known, of a learned hidden-state predictor and a correctness-trained head. Across three project stages, the same 1,366 calibration questions and receiver--Text labels served 460 tests, and the 360 outside the declared family preceded it (60 for three other score families and 300 for correctness-trained heads under label budgets), so the records cannot exclude that the family's design depended on earlier calibration outcomes. The computation itself uses no correctness labels: thresholds rely on fit-split scores frozen before that stage read the calibration disagreement labels, and all 20 tests recompute from the saved fit scores, calibration scores, and answer-only disagreement labels. For the three nominal bounds, the evidence we rely on is the held-out test on 744 questions (Table~\ref{tab:holdout}), which rejects $\rho\ge 0.05$ at $\delta=0.001$ for large/OBQA/Text and C2C, and only at level 0.025 for Text+fact ($p=9.2\times10^{-3}$). No statistic restricted to the calibration split of any of the other 13 settings was computed before freezing ProbeMax, $\alpha$, $\delta$, and the grid, although aggregate statistics over the full training populations that contain these splits had been. The ARC calibration split was drawn, and medium-pair and MMLU-Pro outputs produced, only after freezing these choices; the reused probe file, which included old development scores, was loaded before the C2C calibration, but only fit scores and calibration disagreement enter it.

\paragraph{Parser exposure.}\label{app:parser} The parser was frozen before computing calibration disagreement, but after reviewing raw outputs with gold labels for 341 questions later assigned to calibration splits (208 OBQA and 133 ARC in 355 reviews across small and large pairs; no MMLU-Pro). Questions entered review because two parser versions disagreed, or the earlier version flagged ambiguity; thus, they are not a random sample and exhibit higher disagreement (0.135 vs.\ 0.099 for large/OBQA/Text, 0.075 vs.\ 0.035 for large/ARC/C2C). We prespecified three checks. (i) The primary check removes the 341 questions from every calibration split and reruns all 20 tests with unchanged fit thresholds. The check conditions on the questions that were not reviewed. Every deployed Qwen threshold remains certified except medium/OBQA/C2C, where $q=0.55$ gives 18 changes among 640 skipped ($p=0.0044$), while $q=0.50$ passes (skipping 355 of 742 development questions, changing 7, and answering 490 correctly vs.\ 486 for fixed C2C). No fallback gains a threshold, and the two Llama-8B policies (added later and never reviewed) pass at $0.50$ and $0.55$ instead of $0.60$ and $0.70$ (post hoc). The original tests remain the certificates for these three policies. (ii) The earlier parser, written from 24 synthetic outputs and 1 real output without gold labels and frozen before the review, lacks rules for embedded answers, choice verbs, reversed phrasing, scope, refusals, and some hedge words. Yet, it parses every calibration, development, held-out, sealed, and replay output of the 14 Qwen OBQA, ARC, and large-pair MMLU-Pro settings, the 7 OLMo/Llama settings, and Text+fact exactly as the final parser does, so for the nine certified Qwen-receiver policies every test statistic is what a parser fixed independently of the calibration data would give. The two differ only on some small- and medium-pair MMLU-Pro outputs (which fall back) and five small-pair C2C fit/panel outputs (where the earlier parser returns \textsc{invalid}). (iii) The official C2C evaluator extracts only labels A to D, so it cannot score MMLU-Pro. It reads hedges and repeated phrases as their first letter (\textsc{invalid} under our parser), marks an output unparseable if its option text contains a string it treats as a math term (\texttt{tan}, \texttt{sin}, \texttt{/}), and reads ``A.~0$^\circ$C'' as the last capital letter. Under its labels, medium C2C and large Text thresholds pass, large/OBQA/C2C passes only at $q=0.70$ (its label differences all lie inside the 341 reviewed questions), and large/ARC/C2C falls back. Of the 19 settings in (ii) with labels A to D, 6 certify under both conventions and none under the official extraction alone. One five-option ARC development question lies outside the official range and is left out of that relabeling. The certified risk is defined by the frozen parser, so another notion of answer change can move the boundary.

\paragraph{Extensions and their freeze order.}\label{app:medium} The medium pair, MMLU-Pro (large pair first; small and medium later as an 80-test family), cross-family pairs, Text+fact, stronger helper, GSM8K, SQuAD, and sampled answers were each added after earlier results (post-hoc motivated), and each used a protocol frozen before any of its outputs, committing to report every setting (fixing models, fusers, populations, paths, parser, score, grid, acceptance, and analysis). Before collecting new outputs, later runs reproduced saved outputs bit for bit, and smoke tests read no gold labels; the code repository gives the freeze order and each reproduction check.

%% file: new_app_C.tex
\section{Headroom and the null control}\label{app:complementarity}

\begin{figure}[!htbp]\centering
\includegraphics[width=\linewidth]{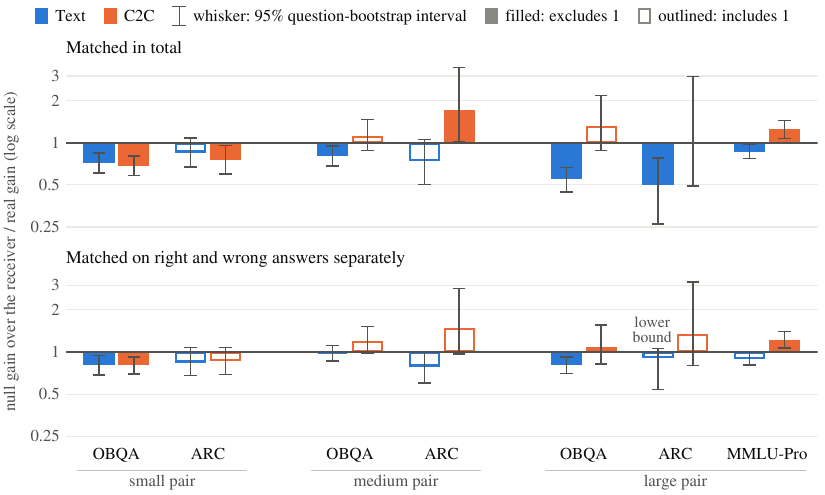}
\caption{Matched one path at a time, the content-free null never falls significantly below the medium- and large-pair C2C paths, whereas whether it falls significantly below Text depends on the match: in five of seven populations unstratified and in two of six stratified. Bar: the null's gain over the receiver divided by the real path's gain (mean of 1,000 draws), drawn from 1 on a log scale, so a bar below 1 means that the null gained less than the real path; whisker: 95\% question-bootstrap interval; filled: the interval excludes 1; outlined: it includes 1. Top: unstratified; bottom: stratified. Large-pair ARC Text under the stratified match is a lower bound. Every value is in Tables~\ref{tab:null_matched} and~\ref{tab:e22}.}
\label{fig:null_paired}
\end{figure}

Input changes without answer-specific content create about as much oracle headroom as the real paths, so headroom alone does not attribute complementarity to communication content. This post-hoc audit of saved development outputs (9,369 action observations on OBQA and ARC, 7,923 on large-pair MMLU-Pro) ran no router. Oracle quantities use gold labels and are non-deployable upper bounds, with \textsc{invalid} scored as incorrect. Figure~\ref{fig:null_paired} summarizes each path against the null.

\paragraph{Null control.} Each receiver agent was rerun alone with its options rotated by one position (with the answer mapped back) and with the Text prompt carrying the saved helper message of another question from the same population (the question $\lfloor N/2\rfloor$ positions later in the frozen order). An unchanged rerun reproduced every receiver-only output, confirming these changes are not run-to-run noise. Option rotation changes 6.69 to 75.20\% of the receiver's answers, and another question's message changes 4.01 to 29.77\% (Table~\ref{tab:null_headroom}). Before matching, an oracle over the receiver and these two inputs achieves at least as much headroom as the oracle over $R$, Text, and C2C in six of seven populations. Text answers more questions correctly than another question's message in six of seven populations, by 1 to 30 (small-pair ARC is the exception, 126 against 130). Because an oracle gains more when more answers change, a rate-matched null action replaces $o_R(x)$ with a random differing perturbed answer on randomly drawn questions. It changes exactly as many answers as Text only ($a$), C2C only ($b$), and both jointly ($c$) (joint design; the per-reference and independent designs match one reference at a time). Ratios average 1,000 draws with 2,000-resample question bootstraps, all fixed before any of these numbers was computed. For medium ARC, the joint ratio is a lower bound due to under-matching (73 eligible questions for 77). Another question's message alone cannot serve as a rate-matched null, because it changes fewer answers than Text across all populations. When Text and C2C change a question to different answers, the two joint nulls must still agree on it if only one alternative answer exists (on average 1.0 to 133.3 questions per draw). This lowers the null gain, so the joint ratio, if anything, understates what the null reproduces.

\paragraph{Gain and headroom.} As gain over $R$, the joint null reproduces a median 76.1\% [67.6, 88.5] of the real gain (64.1 to 97.0\%), and the independent design a median 73.8\%. Matched one reference at a time, the null gains less than Text in all seven populations, but more than C2C in four. This is descriptive and does not split the real gain into a part due to content and a remainder, because real and null changes occur on different questions. Joint-null headroom over the best fixed action equals or exceeds the real one in six of seven populations (median ratio 1.149 [1.000, 1.316]; independent 1.111, Text only 1.235, C2C only 1.306). This is partly because Text outperforms $R$ by 4 to 61 questions, whereas each null action is on average below $R$ in six of seven, so $R$ remains the best fixed action of the null triple in 0.5 to 95\% of draws. Matching separately on questions $R$ answers correctly and incorrectly gives a headroom ratio of 0.967 [0.800, 1.040] (0.905 over the four populations where both strata match fully) and a gain-over-$R$ ratio of 0.849 [0.733, 0.937].

\begin{table}[!htbp]\centering\scriptsize
\caption{Rate-matched nulls without answer-specific content reproduce a median 0.761 of the real gain over $R$ and a median 1.149 of the real headroom over the best fixed action (development sets). Ratios: matched null over real (mean of 1,000 draws) [question bootstrap 95\% CI]. Best fixed action: the best single path (for the null, of $R$ and the two null actions). Joint: two nulls changing as many answers as Text only, C2C only, and both (all paths in Table~\ref{tab:complementarity_main}); Independent: one null per reference, each matched to that reference alone. (a) Real, Null: oracle gain over $R$ in questions for $\{R,\mathrm{Text},\mathrm{C2C}\}$ and for the unmatched null over $R$ and the two perturbed inputs (option rotation: each option's text moved one position down, labels kept; other message: the Text prompt carrying another question's helper message); Text, C2C: per-reference design, each reference against its own matched null (per path in Table~\ref{tab:complementarity_main}). (b) Rotated, Other message: share of $R$'s answers that rotating the options or another question's message changes; Real, Null: headroom in questions (null: joint design, mean over draws); Stratified: joint design run separately on questions $R$ answers correctly and incorrectly. $^{\dagger}$Not fully matched: in (a) the joint ratio, a lower bound; in (b) the stratified ratios (for medium ARC, all three).}
\label{tab:null_matched}\label{tab:null_headroom}\label{tab:null}
\renewcommand{\arraystretch}{1.2}\setlength{\tabcolsep}{2pt}
\vspace{5pt}\textit{(a) Gain over $R$}\par\smallskip
\begin{tabular*}{\linewidth}{@{\extracolsep{\fill}}llcrcccc@{}}\toprule
& & \multicolumn{2}{c}{\begin{tabular}[b]{@{}c@{}}Oracle gain\\(questions)\end{tabular}} & \multicolumn{4}{c@{}}{Null/real ratio [95\% CI]}\\\cmidrule(lr){3-4}\cmidrule(l){5-8}
Pair & Task & Real & \multicolumn{1}{c}{Null} & Independent & Text & C2C & Joint\\\midrule
small & OBQA & 213 & 203 & 0.686 [0.595, 0.782] & 0.719 [0.606, 0.845] & 0.686 [0.584, 0.804] & 0.664 [0.573, 0.759]\\
small & ARC & 94 & 90 & 0.738 [0.608, 0.889] & 0.851 [0.671, 1.080] & 0.747 [0.598, 0.953] & 0.737 [0.605, 0.893]\\
medium & OBQA & 103 & 106 & 0.805 [0.692, 0.937] & 0.800 [0.680, 0.945] & 1.125 [0.880, 1.464] & 0.903 [0.769, 1.048]\\
medium & ARC$^{\dagger}$ & 31 & 28 & 0.772 [0.535, 1.060] & 0.742 [0.505, 1.057] & 1.705 [1.019, 3.449] & 0.845 [0.567, 1.144]\\
large & OBQA & 53 & 62 & 0.701 [0.578, 0.856] & 0.547 [0.443, 0.665] & 1.313 [0.882, 2.175] & 0.761 [0.613, 0.935]\\
large & ARC & 12 & 9 & 0.562 [0.334, 0.803] & 0.498 [0.263, 0.778] & 0.986 [0.490, 2.970] & 0.641 [0.369, 0.900]\\
large & MMLU-Pro & 276 & 317 & 0.884 [0.800, 0.981] & 0.861 [0.769, 0.969] & 1.246 [1.071, 1.447] & 0.970 [0.877, 1.079]\\\midrule
\multicolumn{4}{@{}l}{Median over populations} & 0.738 [0.659, 0.825] & 0.742 [0.644, 0.840] & 1.125 [0.895, 1.349] & 0.761 [0.676, 0.885]\\\bottomrule
\end{tabular*}
\par\vspace{17pt}
\textit{(b) Headroom over the best fixed action}\par\smallskip
\begin{tabular*}{\linewidth}{@{\extracolsep{\fill}}llcccrcrr@{}}\toprule
& & \multicolumn{2}{c}{\begin{tabular}[b]{@{}c@{}}$R$'s answers\\changed (\%)\end{tabular}} & \multicolumn{2}{c}{\begin{tabular}[b]{@{}c@{}}Headroom\\(questions)\end{tabular}} & \multicolumn{3}{c@{}}{Null/real ratio [95\% CI]}\\\cmidrule(lr){3-4}\cmidrule(lr){5-6}\cmidrule(l){7-9}
Pair & Task & Rotated & Other message & Real & \multicolumn{1}{c}{Null} & \begin{tabular}[b]{@{}c@{}}Headroom,\\unstratified\end{tabular} & \multicolumn{1}{c}{\begin{tabular}[b]{@{}c@{}}Headroom,\\stratified\end{tabular}} & \multicolumn{1}{c@{}}{\begin{tabular}[b]{@{}c@{}}Gain over $R$,\\stratified\end{tabular}}\\\midrule
small & OBQA & 75.20 & 28.71 & 132 & 138.1 & 1.046 [0.883, 1.211] & 0.787 [0.709, 0.873] & 0.729 [0.640, 0.826]\\
small & ARC & 69.90 & 29.77 & 49 & 56.3 & 1.149 [0.922, 1.490] & 0.725 [0.592, 0.873] & 0.754 [0.626, 0.904]\\
medium & OBQA$^{\dagger}$ & 24.66 & 20.49 & 70 & 90.9 & 1.298 [0.994, 1.587] & 0.996 [0.884, 1.124] & 0.938 [0.791, 1.083]\\
medium & ARC$^{\dagger}$ & 17.73 & 12.04 & 27 & 25.6 & 0.950 [0.630, 1.405] & 0.939 [0.626, 1.126] & 0.849 [0.583, 1.058]\\
large & OBQA & 18.46 & 9.16 & 25 & 40.1 & 1.603 [1.071, 2.498] & 1.084 [0.946, 1.280] & 0.877 [0.755, 1.005]\\
large & ARC$^{\dagger}$ & 6.69 & 4.01 & 6 & 7.6 & 1.267 [0.513, 3.764] & 0.967 [0.537, 1.500] & 0.750 [0.429, 1.084]\\
large & MMLU-Pro & 34.61 & 20.26 & 239 & 260.8 & 1.091 [0.956, 1.217] & 1.023 [0.923, 1.108] & 0.956 [0.866, 1.058]\\\midrule
\multicolumn{6}{@{}l}{Median over populations} & 1.149 [1.000, 1.316] & 0.967 [0.800, 1.040] & 0.849 [0.733, 0.937]\\\bottomrule
\end{tabular*}
\end{table}

\begin{table}[!htbp]\centering\scriptsize
\caption{The per-reference null matched on right and wrong answers separately (stratified) next to the original per-reference null (unstratified), development sets. Per-reference null: a null matched to one reference alone (per path in Table~\ref{tab:complementarity_main}). Changed by reference: answers the reference changes, on questions $R$ answers correctly ($n_+$) or incorrectly ($n_-$); eligible for null: questions with a perturbed alternative. Oracle gain over $R$: real and null (null: mean of 1,000 draws). $^{\S}$Undetermined: fewer eligible wrong answers than changes; the ratio is a lower bound (0.990 rescaled to $n_-$ changes).}
\label{tab:e22}
\renewcommand{\arraystretch}{1.15}\setlength{\tabcolsep}{2.5pt}
\begin{tabular*}{\linewidth}{@{}lll@{\hspace{5pt}\extracolsep{\fill}}cccccccc@{}}\toprule
& & & \multicolumn{2}{c}{\begin{tabular}[b]{@{}c@{}}Changed by\\reference\end{tabular}} & \multicolumn{2}{c}{\begin{tabular}[b]{@{}c@{}}Eligible\\for null\end{tabular}} & \multicolumn{2}{c}{\begin{tabular}[b]{@{}c@{}}Oracle gain\\over $R$\end{tabular}} & \multicolumn{2}{c}{Null / real gain [95\% CI]}\\\cmidrule(lr){4-5}\cmidrule(lr){6-7}\cmidrule(lr){8-9}\cmidrule(l){10-11}
Pair & Task & Reference & \begin{tabular}[b]{@{}c@{}}$R$ right\\($n_+$)\end{tabular} & \begin{tabular}[b]{@{}c@{}}$R$ wrong\\($n_-$)\end{tabular} & $R$ right & $R$ wrong & Real & Null & Stratified & Unstratified\\\midrule
small & OBQA & Text & 75 & 264 & 179 & 411 & 136 & 109.19 & 0.803 [0.687, 0.940] & 0.719 [0.606, 0.845]\\
small & ARC & Text & 39 & 106 & 60 & 171 & 55 & 46.18 & 0.840 [0.680, 1.074] & 0.851 [0.671, 1.080]\\
medium & OBQA & Text & 49 & 125 & 103 & 150 & 82 & 79.46 & 0.969 [0.859, 1.115] & 0.800 [0.680, 0.945]\\
medium & ARC & Text & 21 & 31 & 32 & 41 & 25 & 19.67 & 0.787 [0.601, 1.004] & 0.742 [0.505, 1.057]\\
large & OBQA & Text & 18 & 54 & 82 & 84 & 46 & 37.24 & 0.809 [0.702, 0.924] & 0.547 [0.443, 0.665]\\
large & ARC & Text & 4 & 11 & 17 & 10 & 10 & 9.00 & 0.900$^{\S}$ [0.538, 1.056] & 0.498 [0.263, 0.778]\\
large & MMLU-Pro & Text & 129 & 432 & 288 & 827 & 166 & 148.20 & 0.893 [0.806, 1.006] & 0.861 [0.769, 0.969]\\
\addlinespace[2pt]
\multicolumn{9}{@{}l}{Median over the 7 Text populations} & 0.840 [0.769, 0.936] & \\
\midrule
small & OBQA & C2C & 66 & 284 & 179 & 411 & 147 & 117.72 & 0.801 [0.694, 0.923] & 0.686 [0.584, 0.804]\\
small & ARC & C2C & 24 & 136 & 60 & 171 & 69 & 59.08 & 0.856 [0.691, 1.078] & 0.747 [0.598, 0.953]\\
medium & OBQA & C2C & 48 & 83 & 103 & 150 & 44 & 52.83 & 1.201 [0.984, 1.517] & 1.125 [0.880, 1.464]\\
medium & ARC & C2C & 25 & 23 & 32 & 41 & 10 & 14.75 & 1.475 [0.964, 2.844] & 1.705 [1.019, 3.449]\\
large & OBQA & C2C & 35 & 25 & 82 & 84 & 16 & 17.24 & 1.077 [0.823, 1.558] & 1.313 [0.882, 2.175]\\
large & ARC & C2C & 6 & 6 & 17 & 10 & 4 & 5.41 & 1.351 [0.800, 3.151] & 0.986 [0.490, 2.970]\\
large & MMLU-Pro & C2C & 204 & 552 & 288 & 827 & 155 & 189.48 & 1.222 [1.067, 1.406] & 1.246 [1.071, 1.447]\\
\addlinespace[2pt]
\multicolumn{9}{@{}l}{Median over the 7 C2C populations} & 1.201 [0.947, 1.331] & \\
\multicolumn{9}{@{}l}{Median over the 5 medium- and large-pair C2C populations} & 1.222 [1.059, 1.448] & \\\bottomrule
\end{tabular*}
\end{table}

\begin{table}[!htbp]\centering\scriptsize
\caption{Direction of the changes, development sets. For each path: its changes on questions $R$ answers incorrectly (to right, to another wrong answer, and the share that become right) and on questions $R$ answers correctly (to wrong). $g_-$: expected share of a null change on a wrong answer that becomes right. (a) The two references; (b) the two perturbed inputs (option rotation: each option's text moved one position down, labels kept; other question's message: the Text prompt carrying another question's helper message).}
\label{tab:e22dir}
\renewcommand{\arraystretch}{1.15}\setlength{\tabcolsep}{2.5pt}
\vspace{5pt}\textit{(a) References}\par\smallskip
\begin{tabular*}{\linewidth}{@{}ll@{\hspace{5pt}\extracolsep{\fill}}ccccccccc@{}}\toprule
& & & \multicolumn{4}{c}{Text} & \multicolumn{4}{c}{C2C}\\\cmidrule(lr){4-7}\cmidrule(l){8-11}
& & Null share & \multicolumn{3}{c}{$R$ wrong} & $R$ right & \multicolumn{3}{c}{$R$ wrong} & $R$ right\\\cmidrule(lr){4-6}\cmidrule(lr){7-7}\cmidrule(lr){8-10}\cmidrule(l){11-11}
Pair & Task & right ($g_-$) & $\to$right & $\to$wrong & share right & $\to$wrong & $\to$right & $\to$wrong & share right & $\to$wrong\\\midrule
small & OBQA & 0.414 & 136 & 128 & 0.515 & 75 & 147 & 137 & 0.518 & 66\\
small & ARC & 0.436 & 55 & 51 & 0.519 & 39 & 69 & 67 & 0.507 & 24\\
medium & OBQA & 0.637 & 82 & 43 & 0.656 & 49 & 44 & 39 & 0.530 & 48\\
medium & ARC & 0.634 & 25 & 6 & 0.806 & 21 & 10 & 13 & 0.435 & 25\\
large & OBQA & 0.690 & 46 & 8 & 0.852 & 18 & 16 & 9 & 0.640 & 35\\
large & ARC & 0.900 & 10 & 1 & 0.909 & 4 & 4 & 2 & 0.667 & 6\\
large & MMLU-Pro & 0.343 & 166 & 266 & 0.384 & 129 & 155 & 397 & 0.281 & 204\\\bottomrule
\end{tabular*}
\par\vspace{17pt}
\textit{(b) Perturbed inputs}\par\smallskip
\begin{tabular*}{\linewidth}{@{}ll@{\hspace{5pt}\extracolsep{\fill}}ccccccccc@{}}\toprule
& & & \multicolumn{4}{c}{Rotated options} & \multicolumn{4}{c}{Other question's message}\\\cmidrule(lr){4-7}\cmidrule(l){8-11}
& & Null share & \multicolumn{3}{c}{$R$ wrong} & $R$ right & \multicolumn{3}{c}{$R$ wrong} & $R$ right\\\cmidrule(lr){4-6}\cmidrule(lr){7-7}\cmidrule(lr){8-10}\cmidrule(l){11-11}
Pair & Task & right ($g_-$) & $\to$right & $\to$wrong & share right & $\to$wrong & $\to$right & $\to$wrong & share right & $\to$wrong\\\midrule
small & OBQA & 0.414 & 164 & 235 & 0.411 & 159 & 78 & 88 & 0.470 & 47\\
small & ARC & 0.436 & 65 & 92 & 0.414 & 52 & 38 & 33 & 0.535 & 18\\
medium & OBQA & 0.637 & 66 & 43 & 0.606 & 74 & 66 & 37 & 0.641 & 49\\
medium & ARC & 0.634 & 19 & 14 & 0.576 & 20 & 16 & 7 & 0.696 & 13\\
large & OBQA & 0.690 & 48 & 23 & 0.676 & 66 & 31 & 13 & 0.705 & 24\\
large & ARC & 0.900 & 6 & 0 & 1.000 & 14 & 7 & 1 & 0.875 & 4\\
large & MMLU-Pro & 0.343 & 232 & 454 & 0.338 & 228 & 151 & 257 & 0.370 & 127\\\bottomrule
\end{tabular*}
\end{table}

\paragraph{Matching one path at a time, on right and wrong answers separately.} After the main results, under rules fixed before computing (post hoc), we evaluated the per-reference design by matching the reference's change rate separately on questions $R$ answers correctly ($S^+$) and incorrectly ($S^-$), then pooling the two draws. Because only changes on $S^-$ affect the gain over $R$, this stratified ratio equals the share of changed wrong answers that the null turns right (denoted $g_-$) divided by the reference's share (Table~\ref{tab:e22dir}). If $S^-$ contains fewer eligible questions than the reference changes, the cell is undetermined and its ratio serves as a lower bound. Intervals follow the prior protocol (2,000 resamples, 100 draws each, seed 0). As detailed in Table~\ref{tab:e22}, Text's ratio is below 1 in all six determined populations, with the interval below 1 in two (small-pair and large-pair OBQA). The large-pair ARC setting is undetermined (0.900, or 0.990 after rescaling to the reference's count). The five medium- and large-pair C2C ratios range from 1.077 to 1.475; none has an interval below 1, and the interval for large-pair MMLU-Pro exceeds 1. A prespecified Monte Carlo check (mean null gain deviating by more than three standard errors from $\min(n_-,\mathrm{elig}_-)\,g_-$) flagged medium/ARC/C2C at 3.28 standard errors. A diagnostic across 400 seeds found no systematic bias (mean deviation $-0.026$ standard errors), and seed 0 was the most extreme of the 400. After observing all results, we lifted the stop and report the prespecified numbers unchanged. This analysis reveals the direction of every change (Table~\ref{tab:e22dir}): relative to $g_-$, the share of changed wrong answers that become right differs by $+0.9$ to $+17.2$ points for Text, $-23.3$ to $+10.4$ for C2C, $-5.8$ to $+10.0$ for rotated options, and $-2.5$ to $+9.9$ for another question's message across the seven populations.

\begin{figure}[!htbp]\centering
\includegraphics[width=\linewidth]{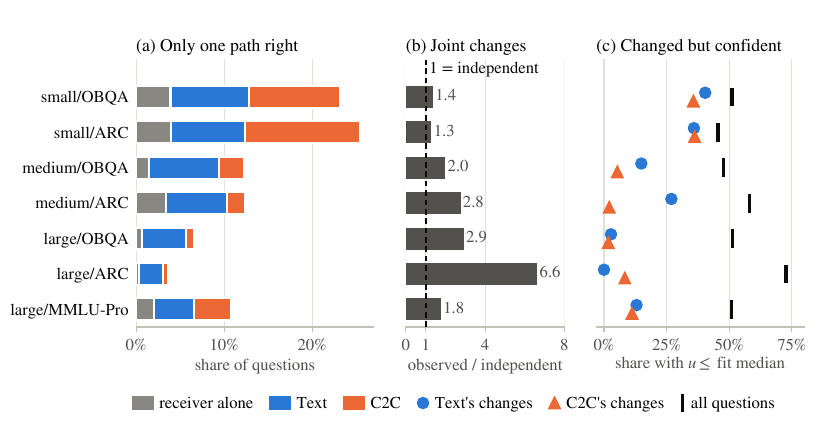}
\caption{Each path is the only one right on some questions, the two references change largely the same questions, and both mostly change answers on which the receiver is unsure (development sets). (a) Questions that only the receiver alone, only Text, or only C2C answers correctly, as a share of the population. (b) Questions changed by both references, divided by the number that independent changes at the observed rates would give. (c) Share of the answers each reference changes whose receiver uncertainty $u$ is at or below the fit-split median; tick: share of all questions.}
\label{fig:null_sources}
\end{figure}

\paragraph{Overlap, confidence, and which path is right.} Each path is uniquely correct on certain questions (Figure~\ref{fig:null_sources}(a)). Text and C2C change the same questions 1.3 to 6.6 times as often as independent changes at their rates would (Figure~\ref{fig:null_sources}(b)). Both references mostly change answers when the receiver is unsure: 0 to 40.4\% of changed answers have an uncertainty $u$ at or below the fit-split median, against 45.5 to 72.6\% of all questions. For the medium pair, 14.9 and 26.9\% of Text's changes are at or below it, compared to 5.3 and 2.1\% of C2C's changes (Figure~\ref{fig:null_sources}(c)). Where exactly one of Text and C2C is right on the medium pair (155 OBQA questions, 96 with Text right; 53 ARC questions, 36 with Text right), ProbeMax separates the two with AUROC 0.569 [0.472, 0.667] (OBQA) and 0.608 [0.438, 0.753] (ARC), with bootstrap resamples stratified by class. Since both intervals include 0.5, this descriptive check establishes no signal for which path is right. Small- and medium-pair MMLU-Pro headroom is in Appendix~\ref{app:mmlu}.

%% file: new_app_D.tex
\section{All settings and robustness of the boundary}\label{app:ledger}

\begin{figure}[!t]\centering
\includegraphics[width=\linewidth]{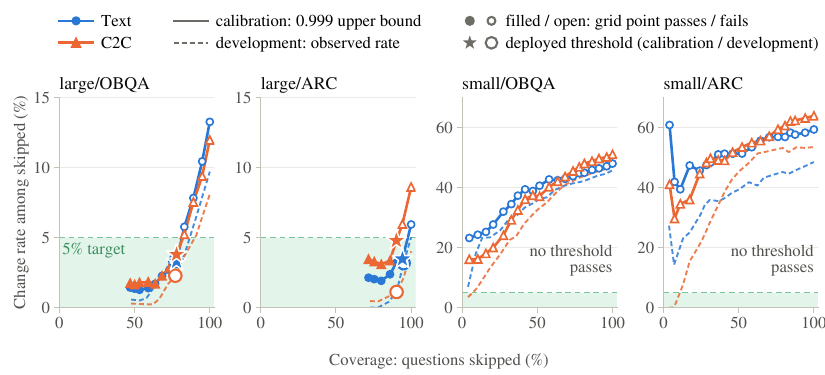}
\caption{The large pair passes the test over a wide range of thresholds under both references, whereas no threshold passes for the small pair. Each curve runs over the 20 predetermined quantiles of the same ProbeMax score, plotted against the share of questions skipped on its own population. Solid: calibration Clopper--Pearson $0.999$ upper bound on the change rate among skipped questions; a grid point passes the exact test ($p\le0.001$) exactly when this bound is below the 5\% target (green band; filled markers). Dashed: observed change rate among skipped development questions, which does not certify a threshold. Star and ring: deployed threshold on calibration and development. Low quantiles that share the zero threshold coincide. Vertical ranges differ by pair; medium-pair summaries are in Table~\ref{tab:dev_full}. Both large/OBQA certificates are nominal (Appendix~\ref{app:history}).}
\label{fig:boundaries}
\end{figure}

\begin{figure}[!t]\centering
\includegraphics[width=\linewidth]{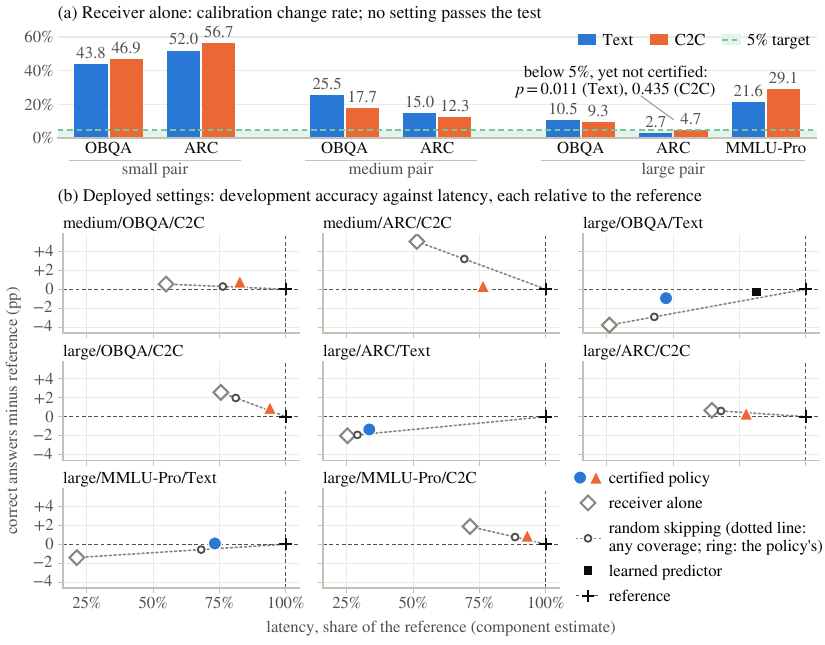}
\caption{Receiver-alone execution fails the test everywhere and random skipping carries no certificate, although both can answer as well as the certified policy or run faster. (a)~Fixed receiver-only execution: calibration change rate in the 14 OBQA, ARC, and large-pair MMLU-Pro settings; none passes the test, including the two large-pair ARC settings below the 5\% target (green band; $p$: exact-test p-value). (b)~The eight deployed settings, one panel each, in the accuracy--latency plane relative to the fixed reference ($+$, dashed lines). Horizontal: mean latency as a share of the reference's, recomposed from per-question development latencies (component estimates, not end-to-end measurements); the policy includes its probe, and random skipping has none. Vertical: development correct answers minus the reference, in percentage points. Filled circle or triangle: the certified policy (Text or C2C); open diamond: receiver alone; dotted line: random skipping at any coverage, which lies on the line from receiver alone to the reference, with a ring at the policy's coverage (expected count); black square: the historical learned predictor (large/OBQA/Text), an earlier stage's receiver hidden-state logistic regression, at its latency in that stage's accounting.}
\label{fig:baselines}
\end{figure}

The certification boundary is stable: each of the 27 re-split settings (Tables~\ref{tab:dev_full} and~\ref{tab:mmlu_small_medium}) retains its outcome in at least 96\% of 200 re-drawn splits. At $\alpha=0.05$, none of the tested post-hoc alternative tests, scores, or learned predictors (Table~\ref{tab:sensitivity}) certifies a setting that ProbeMax rejects. Figure~\ref{fig:boundaries} shows every grid point of the large and small pairs, and Figure~\ref{fig:baselines} compares the policies with simple baselines under the same target. All grid tests are provided in the released dataset; gain retention and the label-light certificate are in Appendix~\ref{app:accuracy}, and boundary predictions in Appendix~\ref{app:predict}.

\begin{table}[!htbp]\centering\scriptsize
\caption{Ten of these 23 settings deploy and thirteen fall back (development). OLMo: large helper, OLMo-2-1124-7B-Instruct receiver; Llama: Llama-3.2-1B-Instruct helper, small receiver; Llama-8B: large helper, Llama-3.1-8B-Instruct receiver. Disagreements: $R$ and the reference answer differently; smallest 0.999 bound: smallest calibration Clopper--Pearson $0.999$ bound over the grid (diagnostic); fallback: no threshold passes; --: no question skipped; correct: development answers correct under the policy and under the reference. $^\ddagger$Nominal certificate (Appendix~\ref{app:history}). $^{*}$Without the 341 parser-review questions, only a lower $q$ is certified: 0.50 for medium/OBQA/C2C (47.8\% coverage, 7 of 355 changed), 0.50 for Llama-8B/OBQA, and 0.55 for Llama-8B/ARC. Parser-review questions: 341 calibration questions whose outputs were read with gold labels before the parser was frozen.}
\label{tab:dev_full}
\renewcommand{\arraystretch}{1.1}\setlength{\tabcolsep}{2.5pt}
\begin{tabular*}{\linewidth}{@{}lll@{\hspace{5pt}\extracolsep{\fill}}rccrccc@{}}\toprule
& & & \multicolumn{1}{c}{Disagreements} & & Smallest & Deployed & Coverage & Changed & Correct: policy\\
Pair & Task & Reference & \multicolumn{1}{c}{/ questions} & AUROC & 0.999 bound & \multicolumn{1}{c}{$q$} & (\%) & / skipped & / reference\\\midrule
small & OBQA & Text & 339/742 & 0.641 & 0.232 & fallback & 0.0 & -- & 346/346\\
small & OBQA & C2C & 350/742 & 0.700 & 0.161 & fallback & 0.0 & -- & 366/366\\
small & ARC & Text & 145/299 & 0.634 & 0.394 & fallback & 0.0 & -- & 126/126\\
small & ARC & C2C & 160/299 & 0.641 & 0.296 & fallback & 0.0 & -- & 155/155\\\midrule
medium & OBQA & Text & 174/742 & 0.787 & 0.096 & fallback & 0.0 & -- & 523/523\\
medium & OBQA & C2C$^{*}$ & 131/742 & 0.840 & 0.032 & 0.55 & 52.6 & 11/390 & 492/486\\
medium & ARC & Text & 52/299 & 0.742 & 0.102 & fallback & 0.0 & -- & 221/221\\
medium & ARC & C2C & 48/299 & 0.910 & 0.031 & 0.60 & 63.2 & 4/189 & 203/202\\\midrule
large & OBQA & Text$^\ddagger$ & 72/742 & 0.873 & 0.013 & 0.80 & 77.1 & 19/572 & 638/645\\
large & OBQA & C2C$^\ddagger$ & 60/742 & 0.883 & 0.016 & 0.80 & 77.1 & 13/572 & 605/598\\
large & ARC & Text & 15/299 & 0.952 & 0.019 & 0.95 & 95.0 & 9/284 & 270/274\\
large & ARC & C2C & 12/299 & 0.909 & 0.031 & 0.90 & 90.3 & 3/270 & 267/266\\
large & MMLU-Pro & Text & 561/2641 & 0.821 & 0.021 & 0.40 & 40.5 & 40/1069 & 1321/1319\\
large & MMLU-Pro & C2C & 756/2641 & 0.846 & 0.015 & 0.40 & 40.5 & 37/1069 & 1257/1233\\\midrule
OLMo & OBQA & Text & 172/742 & 0.790 & 0.118 & fallback & 0.0 & -- & 516/516\\
OLMo & ARC & Text & 67/299 & 0.749 & 0.169 & fallback & 0.0 & -- & 222/222\\
OLMo & MMLU-Pro & Text & 1112/2641 & 0.633 & 0.252 & fallback & 0.0 & -- & 664/664\\\midrule
Llama & OBQA & Text & 191/742 & 0.693 & 0.095 & fallback & 0.0 & -- & 333/333\\
Llama & OBQA & C2C & 427/742 & 0.618 & 0.325 & fallback & 0.0 & -- & 358/358\\
Llama & ARC & Text & 88/299 & 0.678 & 0.296 & fallback & 0.0 & -- & 145/145\\
Llama & ARC & C2C & 174/299 & 0.615 & 0.247 & fallback & 0.0 & -- & 146/146\\\midrule
Llama-8B & OBQA & Text$^{*}$ & 86/742 & 0.874 & 0.022 & 0.60 & 59.0 & 7/438 & 638/636\\
Llama-8B & ARC & Text$^{*}$ & 37/299 & 0.880 & 0.033 & 0.70 & 68.9 & 4/206 & 258/260\\
\bottomrule
\end{tabular*}
\end{table}

\paragraph{Calibration statistics.} Table~\ref{tab:calstats} lists the calibration test at the deployed quantile for the 14 OBQA, ARC, and large-pair MMLU-Pro settings (Table~\ref{tab:dev_full}), or at the grid point with the smallest $p$-value for fallbacks. With $\alpha=0.05$ and $\delta=0.001$, a skipped subset with no observed changes passes only if it contains at least 135 calibration questions, since $0.95^{135}<0.001<0.95^{134}$. At $q=0.05$, the small pair skips only 65 OBQA and 18 ARC calibration questions. Figure~\ref{fig:boundaries} shows the full grid: the large pair's 0.999 upper bound stays below 5\% until 78 to 94\% of calibration questions are skipped, whereas the small pair's never falls below 16\%.

\begin{table}[!htbp]\centering\scriptsize
\caption{Deployed settings pass the exact test with 290 to 2,358 skipped calibration questions, whereas the small pair's best candidates skip at most 65. Rows: small, medium, and large pairs only. Tested $q$: the deployed quantile, or for a fallback the grid point with the smallest $p$; $n$, $k$: skipped and changed calibration questions at the tested $q$; $p$: exact-test p-value; --: not applicable. Reviewed out: the test without the 341 parser-review questions. Parser-review questions: 341 calibration questions whose outputs were read with gold labels before the parser was frozen (none on MMLU-Pro). Re-split: share of 200 re-drawn splits that deploy. Needed $N_{\rm cal}$: calibration size at which the observed rate would pass (never: the observed rate is at least 0.05). $^\ddagger$Nominal certificate (Text: reused from an earlier 100-test family; Appendix~\ref{app:history}).}
\label{tab:calstats}
\renewcommand{\arraystretch}{1.1}\setlength{\tabcolsep}{2.5pt}
\begin{tabular*}{\linewidth}{@{}lll@{\hspace{5pt}\extracolsep{\fill}}ccccccccc@{}}\toprule
& & & & & \multicolumn{3}{c}{Full calibration split} & \multicolumn{2}{c}{Reviewed out} & & \\\cmidrule(lr){6-8}\cmidrule(lr){9-10}
Pair & Task & Reference & Deployed $q$ & Tested $q$ & $n$ & $k$ & $p$ & $k/n$ & $p$ & Re-split & Needed $N_{\rm cal}$\\\midrule
small & OBQA & Text & fallback & 0.05 & 65 & 5 & 0.894 & -- & -- & 0.00 & never\\
small & OBQA & C2C & fallback & 0.05 & 65 & 2 & 0.363 & -- & -- & 0.00 & 21,793\\
small & ARC & Text & fallback & 0.10 & 33 & 5 & 0.995 & -- & -- & 0.00 & never\\
small & ARC & C2C & fallback & 0.10 & 33 & 2 & 0.773 & -- & -- & 0.00 & never\\
\addlinespace[2pt]
medium & OBQA & Text & fallback & 0.05 & 601 & 36 & 0.884 & -- & -- & 0.00 & never\\
medium & OBQA & C2C & 0.55 & 0.55 & 739 & 19 & $6.9\times10^{-4}$ & 18/640 & $4.4\times10^{-3}$ & 1.00 & --\\
medium & ARC & Text & fallback & 0.05 & 256 & 12 & 0.483 & -- & -- & 0.00 & 79,333\\
medium & ARC & C2C & 0.60 & 0.60 & 290 & 1 & $5.6\times10^{-6}$ & 1/210 & $2.5\times10^{-4}$ & 0.96 & --\\
\addlinespace[2pt]
large & OBQA & Text$^\ddagger$ & 0.80 & 0.80 & 1064 & 18 & $1.2\times10^{-8}$ & 17/908 & $7.8\times10^{-7}$ & 1.00 & --\\
large & OBQA & C2C$^\ddagger$ & 0.80 & 0.80 & 1064 & 22 & $6.7\times10^{-7}$ & 19/908 & $5.4\times10^{-6}$ & 1.00 & --\\
large & ARC & Text & 0.95 & 0.95 & 422 & 4 & $4.8\times10^{-6}$ & 2/297 & $3.4\times10^{-5}$ & 1.00 & --\\
large & ARC & C2C & 0.90 & 0.90 & 404 & 7 & $5.4\times10^{-4}$ & 2/282 & $6.6\times10^{-5}$ & 1.00 & --\\
large & MMLU-Pro & Text & 0.40 & 0.40 & 2358 & 67 & $1.3\times10^{-7}$ & 67/2358 & $1.3\times10^{-7}$ & 1.00 & --\\
large & MMLU-Pro & C2C & 0.40 & 0.40 & 2358 & 75 & $1.0\times10^{-5}$ & 75/2358 & $1.0\times10^{-5}$ & 1.00 & --\\\bottomrule
\end{tabular*}
\end{table}

\paragraph{Stability and sensitivity.} We re-drew the fit and calibration splits of these 27 settings 200 times using the original grouping, sizes, and shuffle rule (for MMLU-Pro, the seed of its fixed pseudorandom order), recomputed the 20 thresholds, and reran the test; development sets remained unchanged. Medium/ARC/C2C deploys in 96\% of re-draws, every other deployment in all 200, and no fallback setting deploys in any (the stronger-helper fallbacks certify in none and 15.5\%, Appendix~\ref{app:informed}; Text+fact and GSM8K were not re-split). Among deploying re-draws, the median deployed $q$ matches the original everywhere except large/ARC/Text (0.90 instead of 0.95). Five fallback settings observe a change rate below 0.05 but would need 3,975 (OLMo/ARC/Text), 7,494 (Llama/OBQA/Text), 13,386 (Llama/ARC/C2C), 21,793 (small/OBQA/C2C), and 79,333 (medium/ARC/Text) calibration questions to pass. The medium pair's disagreement remains similar across its fit, calibration, and development splits (OBQA: 24.4, 25.5, and 23.5\% with Text, 16.8, 17.7, and 17.7\% with C2C; ARC: 20.0, 15.0, and 17.4\%, and 16.0, 12.3, and 16.1\%), and its probe produced no \textsc{invalid} score.

\paragraph{Post-hoc variants.} Table~\ref{tab:sensitivity} recomputes every setting from saved records under post-hoc variants modifying one element each. The variants are: setting $\alpha=0.02$ or $0.10$ with the same per-candidate cutoff; fixed-sequence testing (evaluating candidates in ascending order of $q$, each at the family's 0.02 Bonferroni bound instead of 0.001, deploying the last accepted $q$); replacing ProbeMax with probe entropy or one minus the top-two probability margin (read from the same prefill); and using the learned predictor below. This predictor is refitted per setting on its fit split using historical features and hyperparameters (fixed on large/OBQA/Text, not tuned per setting). Thresholds always derive from fit-split quantiles. At $\alpha=0.05$, no variant certifies a setting that ProbeMax rejects, and every small-pair setting falls back under all variants. Setting $\alpha=0.02$ removes the medium pair, large/ARC/C2C, and MMLU-Pro/Text, while $\alpha=0.10$ adds medium/OBQA/Text. Fixed receiver-only execution ($q=1$), which fails the prespecified test everywhere, is accepted for large-pair ARC at $\alpha=0.10$ (both references) and under fixed-sequence testing (Text; 12 changes among 448 calibration questions, $p=0.011$). The learned predictor's development AUROC for receiver--reference disagreement trails ProbeMax's in every setting (0.55--0.64 for the small pair, 0.69--0.75 for the medium pair, 0.64--0.88 for the large pair); its lower coverage does not imply learned skipping is useless. Fixed-sequence testing yields an equal or larger $q$ across all settings here, but it depends on testing order and stops at the first failure. Bonferroni remains primary because it was prespecified before choosing these variants.

\begin{table}[!htbp]\centering\scriptsize
\caption{At $\alpha=0.05$, no post-hoc variant certifies a setting that the prespecified procedure rejects. (a) Deployed $q$, the quantile that sets the skip threshold (a larger $q$ skips more questions, and 1 skips all); --: no threshold passes, so the setting keeps communicating (fallback). (b) At the deployed $q$, skipped development questions whose answer changes, out of all skipped ones. Primary: ProbeMax, $\alpha=0.05$, $\delta=0.001$. Row small/both/both: the four small-pair OBQA and ARC settings. Fixed sequence: candidates tested in ascending order of $q$, each at 0.02 instead of 0.001, until one fails. Entropy, Margin: probe entropy, or one minus the top-two probability margin, replaces ProbeMax. Learned: learned predictor (receiver hidden-state logistic regression refitted on each fit split). $^\ddagger$Nominal certificate (Text: reused from an earlier 100-test family; Appendix~\ref{app:history}); $^\S$tested on its calibration labels before.}
\label{tab:sensitivity}
\renewcommand{\arraystretch}{1.2}\setlength{\tabcolsep}{3pt}
\vspace{5pt}\textit{(a) Deployed $q$}\par\smallskip
\begin{tabular*}{\linewidth}{@{}lll@{\extracolsep{\fill}}ccccccc@{}}\toprule
& & & & \multicolumn{6}{c}{Post-hoc variants}\\\cmidrule(l){5-10}
Pair & Task & Reference & Primary & $\alpha=0.02$ & $\alpha=0.10$ & Fixed sequence & Entropy & Margin & Learned\\\midrule
small & both & both & -- & -- & -- & -- & -- & -- & --\\
\addlinespace[3pt]
medium & OBQA & Text & -- & -- & 0.45 & -- & -- & -- & --\\
medium & OBQA & C2C & 0.55 & -- & 0.65 & 0.55 & 0.55 & 0.55 & --\\
medium & ARC & Text & -- & -- & -- & -- & -- & -- & --\\
medium & ARC & C2C & 0.60 & -- & 0.75 & 0.60 & 0.60 & 0.60 & --\\
\addlinespace[3pt]
large & OBQA & Text$^\ddagger$ & 0.80 & 0.65 & 0.90 & 0.80 & 0.80\rlap{$^\S$} & 0.80 & 0.30\rlap{$^\S$}\\
large & OBQA & C2C$^\ddagger$ & 0.80 & 0.65 & 0.95 & 0.85 & 0.80 & 0.80 & 0.55\\
large & ARC & Text & 0.95 & 0.80 & 1 & 1 & 0.95 & 0.95 & 0.90\\
large & ARC & C2C & 0.90 & -- & 1 & 0.95 & 0.90 & 0.90 & 0.65\\
large & MMLU-Pro & Text & 0.40 & -- & 0.60 & 0.45 & 0.40 & 0.40 & 0.05\\
large & MMLU-Pro & C2C & 0.40 & 0.25 & 0.50 & 0.40 & 0.40 & 0.40 & --\\\bottomrule
\end{tabular*}
\par\vspace{17pt}
\textit{(b) Changed / skipped development questions at the deployed $q$}\par\smallskip
\begin{tabular*}{\linewidth}{@{}lll@{\extracolsep{\fill}}ccccccc@{}}\toprule
& & & & \multicolumn{6}{c}{Post-hoc variants}\\\cmidrule(l){5-10}
Pair & Task & Reference & Primary & $\alpha=0.02$ & $\alpha=0.10$ & Fixed sequence & Entropy & Margin & Learned\\\midrule
small & both & both & -- & -- & -- & -- & -- & -- & --\\
\addlinespace[3pt]
medium & OBQA & Text & -- & -- & 18/321 & -- & -- & -- & --\\
medium & OBQA & C2C & 11/390 & -- & 25/481 & 11/390 & 11/389 & 11/391 & --\\
medium & ARC & Text & -- & -- & -- & -- & -- & -- & --\\
medium & ARC & C2C & 4/189 & -- & 7/219 & 4/189 & 3/187 & 3/187 & --\\
\addlinespace[3pt]
large & OBQA & Text$^\ddagger$ & 19/572 & 6/460 & 44/676 & 19/572 & 18/572 & 19/572 & 6/230\\
large & OBQA & C2C$^\ddagger$ & 13/572 & 1/460 & 45/703 & 22/613 & 13/572 & 13/572 & 13/399\\
large & ARC & Text & 9/284 & 0/233 & 15/299 & 15/299 & 9/284 & 9/284 & 7/272\\
large & ARC & C2C & 3/270 & -- & 12/299 & 6/284 & 3/270 & 3/270 & 4/194\\
large & MMLU-Pro & Text & 40/1069 & -- & 109/1596 & 56/1205 & 40/1068 & 40/1069 & 5/128\\
large & MMLU-Pro & C2C & 37/1069 & 6/697 & 85/1341 & 37/1069 & 37/1068 & 38/1069 & --\\\bottomrule
\end{tabular*}
\end{table}

\paragraph{Same-target baselines (post hoc).} Figure~\ref{fig:baselines} compares each policy with two simple skipping baselines under the same target; no threshold, reference, or split was changed, and neither baseline was replayed end-to-end. Fixed receiver-only execution ($q=1$) fails the test in all 14 OBQA, ARC, and large-pair MMLU-Pro settings (calibration change rate 2.7 to 56.7\%). The two large-pair ARC rates fall below 5\% but yield $p=0.011$ (Text, 12 of 448) and $p=0.435$ (C2C, 21 of 448); there, fixed $R$ answers about as many development questions correctly as the certified policies, at a lower component-estimated latency. A rule skipping questions at random, independently of content, changes answers at the receiver--reference disagreement rate in expectation regardless of coverage (2.7 to 29.1\% on calibration for the eight deployed settings). At the policy's development coverage $\kappa$, its expected correct count $(1-\kappa)\,a_b+\kappa\,a_R$ (where $a_b$ and $a_R$ are the correct counts of the reference and $R$) exceeds the policy's in three C2C settings (medium ARC, large OBQA, large ARC) and trails it in the other five. Its component-estimated latency is 18 to 67~ms lower (it runs no probe, and its decision overhead was not measured). The probability that a random subset of a grid point's size, drawn without replacement from the fixed calibration split, passes the test remains below $10^{-4}$ at every grid point in every setting except large ARC with Text, peaking at 0.017 (at $q=0.80$). These are per-grid-point probabilities, not the probability that a search over the grid passes.

\paragraph{Historical learned predictor.} An earlier stage evaluated ProbeMax and a learned predictor on large/OBQA/Text using the same grid, $\alpha$, and cutoff, within its declared 100-test family (Bonferroni level 0.10; Appendix~\ref{app:history}), where both passed. The predictor is an L2-regularized logistic regression on 8,192-dimensional receiver hidden states (last-token outputs of an intermediate layer and the final normalization layer of Qwen3-8B), fitted on the 2,100 fit questions to predict receiver--Text disagreement. It requires reference executions and offline feature extraction but no correctness labels. It skipped 230 of 742 development questions with 6 changes and 643 correct answers (component latency 797.2~ms in that stage's accounting), against 572 skipped, 19 changes, 638 correct, and 460.9~ms for ProbeMax (462.9~ms in the unified records of Figure~\ref{fig:baselines}).

\begin{table}[!htbp]\centering\scriptsize
\caption{All four small- and medium-pair MMLU-Pro settings fall back, and none certifies in any of 200 re-drawn splits ($N=2641$ development questions). Disagreement: share of development questions on which $R$ and the reference answer differently. Best candidate: smallest exact-test p-value over the 20 candidates, with its $k/n$ (changed/skipped calibration questions). Correct and \textsc{invalid}: development counts of $R$ and the reference. Re-split: share of the 200 re-drawn splits that deploy.}
\label{tab:mmlu_small_medium}
\renewcommand{\arraystretch}{1.1}\setlength{\tabcolsep}{2.5pt}
\begin{tabular*}{\linewidth}{@{}ll@{\hspace{5pt}\extracolsep{\fill}}cccccccccc@{}}\toprule
& & & & \multicolumn{2}{c}{Best candidate} & & \multicolumn{2}{c}{Correct} & \multicolumn{2}{c}{\textsc{invalid}} & \\\cmidrule(lr){5-6}\cmidrule(lr){8-9}\cmidrule(lr){10-11}
Pair & Reference & Disagreement (\%) & AUROC & $p$ & $k/n$ & Deployed $q$ & $R$ & Reference & $R$ & Reference & Re-split\\\midrule
small & Text & 46.76 & 0.644 & 1 & 102/280 & fallback & 403 & 516 & 151 & 229 & 0.000\\
small & C2C & 60.24 & 0.626 & 1 & 106/280 & fallback & 403 & 462 & 151 & 420 & 0.000\\
medium & Text & 42.71 & 0.717 & 1 & 64/690 & fallback & 813 & 913 & 160 & 18 & 0.000\\
medium & C2C & 53.39 & 0.765 & 0.193 & 29/690 & fallback & 813 & 530 & 160 & 1124 & 0.000\\\bottomrule
\end{tabular*}
\end{table}

\begin{table}[!htbp]\centering\scriptsize
\caption{The pooled MMLU-Pro certificate gives no subgroup guarantee: some categories exceed 5\% changed answers (development group representatives; descriptive). Both references skip the same questions. Policy: large pair at its deployed $q$; $N$: development questions in the category.}
\label{tab:mmlu_subgroups}
\begin{tabular*}{\linewidth}{@{}l@{\extracolsep{\fill}}rcccc@{}}\toprule
& & & & \multicolumn{2}{c@{}}{Changed / skipped}\\\cmidrule(l){5-6}
Category & \multicolumn{1}{c}{$N$} & Skipped & Coverage (\%) & Text & \multicolumn{1}{c@{}}{C2C}\\\midrule
biology & 144 & 100 & 69.4 & 1/100 & 0/100\\
business & 177 & 47 & 26.6 & 1/47 & 2/47\\
chemistry & 257 & 60 & 23.3 & 4/60 & 6/60\\
computer science & 93 & 42 & 45.2 & 4/42 & 1/42\\
economics & 190 & 116 & 61.1 & 2/116 & 0/116\\
engineering & 220 & 72 & 32.7 & 4/72 & 12/72\\
health & 155 & 95 & 61.3 & 5/95 & 0/95\\
history & 86 & 47 & 54.7 & 1/47 & 0/47\\
law & 217 & 62 & 28.6 & 4/62 & 2/62\\
math & 305 & 97 & 31.8 & 4/97 & 9/97\\
other & 210 & 96 & 45.7 & 5/96 & 1/96\\
philosophy & 113 & 39 & 34.5 & 2/39 & 0/39\\
physics & 293 & 82 & 28.0 & 0/82 & 3/82\\
psychology & 181 & 114 & 63.0 & 3/114 & 1/114\\
\bottomrule\end{tabular*}
\end{table}

\begin{table}[!htbp]\centering\scriptsize\revcolor
\caption{\rev{Certifying each MMLU-Pro discipline separately trades most of the pooled coverage for per-discipline guarantees (development; post-hoc aggregation of the per-discipline tests). Each discipline uses its own fit quantiles and calibration questions; a discipline without a certified threshold always runs the reference. Coverage: total skipped questions over all 2,641 development questions. Changed: changed among skipped, with Clopper--Pearson 95\% CI for certified disciplines. Per discipline at $\delta=0.001$, the nominal family-wise error over the 14 disciplines is at most 0.28 per reference; with $\delta/14$ it is at most 0.02.}}
\label{tab:mmlu_disc}
\setlength{\tabcolsep}{3pt}
\begin{tabular*}{\linewidth}{@{}l@{\extracolsep{\fill}}lccc@{}}\toprule
\multicolumn{5}{@{}l}{\textit{(a) Overall}}\\
Reference & Strategy & Skipped & Coverage (\%) & Changed\\\midrule
Text & pooled certificate & 1,069 & 40.5 & 40/1,069\\
Text & per discipline, $\delta=0.001$ & 292 & 11.1 & 6/292\\
Text & per discipline, $\delta=0.001/14$ & 0 & 0.0 & --\\
C2C & pooled certificate & 1,069 & 40.5 & 37/1,069\\
C2C & per discipline, $\delta=0.001$ & 293 & 11.1 & 3/293\\
C2C & per discipline, $\delta=0.001/14$ & 97 & 3.7 & 0/97\\\midrule
\multicolumn{5}{@{}l}{\textit{(b) Certified disciplines}}\\
Reference & Discipline ($\delta$) & $q$ & Skipped / $N$ & Changed [95\% CI, \%]\\\midrule
Text & biology (0.001) & 0.60 & 93/144 & 1/93 [0.03, 5.85]\\
Text & economics (0.001) & 0.45 & 97/190 & 0/97 [0.00, 3.73]\\
Text & engineering (0.001) & 0.35 & 102/220 & 5/102 [1.61, 11.07]\\
C2C & health (0.001) & 0.45 & 74/155 & 0/74 [0.00, 4.86]\\
C2C & other (0.001) & 0.50 & 101/210 & 2/101 [0.24, 6.97]\\
C2C & psychology (0.001) & 0.65 & 118/181 & 1/118 [0.02, 4.63]\\
C2C & psychology (0.001/14) & 0.60 & 97/181 & 0/97 [0.00, 3.73]\\\bottomrule
\end{tabular*}
\end{table}

\paragraph{MMLU-Pro.}\label{app:mmlu} The large receiver shows exact zero uncertainty on 713 of 3,000 fit groups (1,442 of 6,000 calibration groups), merging quantiles $q\le0.20$ into one threshold. Both large-pair references pass at $q=0.40$ (67 and 75 changes among 2,358 skipped calibration groups) and fail at $q=0.45$ (105 and 118 of 2,654; $p=0.0063$ and $0.101$). On development, ProbeMax's average precision for disagreement is 0.522 (Text) and 0.629 (C2C). Of the policies' changed answers, 14 (Text) and 27 (C2C) became correct, 12 and 3 became incorrect, and 14 and 7 stayed incorrect. \textsc{invalid} counts are 6 ($R$), 2 (Text), and 20 (C2C), and 2 and 19 for the two policies. The pooled certificate gives no subgroup guarantee: some development categories exceed 5\% changed answers (e.g., engineering C2C, 12 of 72; Table~\ref{tab:mmlu_subgroups}). Run within each of the 14 categories under a rule fixed before computing, using category-specific fit quantiles and calibration questions, the test certifies three categories for Text (biology, economics, and engineering; the latter changing 5 of 102 skipped development answers under its own threshold) and three for C2C (health, other, and psychology). With $\delta$ further divided by 14, it certifies none for Text and one (psychology) for C2C. \rev{Skipping only within certified disciplines lowers development coverage from 40.5\% to 11.1\% for both references, and to 0.0\% (Text) and 3.7\% (C2C) with $\delta/14$ (Table~\ref{tab:mmlu_disc}).} All four small- and medium-pair settings fall back (smallest $p=0.193$ over 80 tests; disagreement 42.7 to 60.2\%, AUROC 0.626 to 0.765; Table~\ref{tab:mmlu_small_medium}). Their answers fail to parse more often, above all with C2C: \textsc{invalid} shares of development answers are 5.7 ($R$), 8.7 (Text), and 15.9\% (C2C) for the small pair, and 6.1, 0.7, and 42.6\% for the medium pair, compared to 0.2, 0.1, and 0.8\% for the large pair. Consequently, medium-pair C2C answers far fewer questions correctly than the receiver alone (530 against 813). Their three-action headroom is 11.59 (small) and 7.91 points (medium) over the best fixed action, against 9.05 for the large pair, and 15.87 and 11.70 points over the receiver alone (no null control was run for them).

%% file: new_app_E.tex
\section{Extensions}\label{app:extensions}

The same pattern holds across all extensions: messages that rarely and predictably change the receiver's answer are certified (the Llama-8B receiver, Text+fact nominally, SQuAD, and sampled answers, on OBQA nominally). A stronger helper for the medium receiver falls back because its calibration splits are too small, while the remaining settings fall back (the OLMo and Llama pairs, GSM8K). Each ran under a protocol fixed before its outputs (Appendix~\ref{app:medium}).

\subsection{Cross-family receivers and helpers}\label{app:xfam}

Each cross-family pair ran the main study's protocol, using prompts in each model's native chat template, under its own frozen protocol, with every setting reported. The OLMo pair combines the large helper (whose saved messages serve as Text messages, matching token counts on every row) with an OLMo-2-1124-7B-Instruct receiver \citep{olmo2025olmo2}, for which no C2C fuser exists (a 60-test family). The Llama pair combines a Llama-3.2-1B-Instruct helper \citep{grattafiori2024llama3} with the small receiver on OBQA and ARC (an 80-test family). Its C2C outputs come from the official evaluator and cross-family fuser run on our rows, with receiver prompts byte-identical to ours across all 5,626 rows and 16 saved small-pair C2C outputs reproduced bit for bit. All seven settings fall back (Table~\ref{tab:dev_full}). OLMo disagrees with Text on 23.2, 22.4, and 42.1\% of OBQA, ARC, and MMLU-Pro development questions (Qwen3-8B: 9.7, 5.0, and 21.2\%), at AUROC 0.790, 0.749, and 0.633 (0.873, 0.952, and 0.821), and answers 71.6, 68.2, and 22.4\% correctly alone. The Llama helper raises the small receiver's accuracy from 38.4 to 44.9 (Text) and 48.2\% (C2C) on OBQA, and from 36.8 to 48.5 and 48.8\% on ARC. Its Text messages change fewer answers than the Qwen helper's (25.7 and 29.4\% against 45.7 and 48.5\%), but they are ranked weakly (AUROC 0.693 and 0.678), and the fuser changes 57.5 and 58.2\%.

Part of the OLMo fallback stems from answer format, as unparseable outputs count as changes: 4.3, 9.2, and 17.1\% of OLMo's receiver-only and 5.9, 8.0, and 13.1\% of its Text calibration answers are \textsc{invalid} (on development 2.8, 12.0, and 17.0\% and 3.5, 6.4, and 13.7\%; Llama-pair receiver 3.4 and 10.7\%, references 2.7 to 6.9\%; Llama-8B pair at most 0.1\%). Furthermore, 72 of 320, 47 of 95, and 968 of 2,457 calibration disagreements involve exactly one \textsc{invalid} output. A non-deployable, post-hoc filter retaining only questions where both answers parse certifies all three OLMo settings ($q=0.45$, 0.40, and 0.05; development disagreement 18.8, 15.5, and 33.5\%, AUROC 0.855, 0.881, and 0.701) but no Llama setting. No deployable rule we tested certifies OLMo or any Llama setting. Post hoc, answering skipped questions with the probe's argmax agrees with OLMo's own answer on 95.2, 90.2, and 81.1\% of calibration questions but passes no threshold (smallest $p=0.901$, 0.377, and 1.000; AUROC 0.808, 0.765, and 0.653), because the reference's unparseable answers (15.8, 29.2, and 31.2\% of development disagreements) remain changes; routing unparseable receiver answers to the reference also fails (Appendix~\ref{app:restson}).

\paragraph{Llama-8B pair.} A Llama-3.1-8B-Instruct receiver reading the large helper's saved messages was added after these fell back, as the first of three candidates in an order fixed before any output (followed by Mistral-7B-Instruct-v0.3 and Granite-3.1-8B-Instruct, which produced no outputs, making OLMo-2 the only other non-Qwen receiver run). It used its default chat template and a 40-test family. It disagrees with Text on 11.6 and 12.4\% of OBQA and ARC development questions (AUROC 0.874 and 0.880) and is certified at $q=0.60$ (21 of 800 skipped calibration answers change, $p=5.7\times10^{-4}$) and $q=0.70$ (4 of 320, $p=3.2\times10^{-4}$). On development, 7 of 438 (1.60\% [0.64, 3.27]) and 4 of 206 (1.94\% [0.53, 4.90]) skipped answers change. Both deploy in all 200 re-drawn splits and keep their quantile on questions whose two answers parse and with argmax answers; no score is exactly zero. Text raises accuracy from 603 to 636 of 742 and from 239 to 260 of 299, with the policies answering 638 and 258 correctly.

\subsection{Informed and stronger helpers}\label{app:informed}

\paragraph{Text+fact.} Post hoc, the large helper's input gains one line before the question, \texttt{Useful fact: <fact1>}, providing the OpenBookQA science fact behind the question. Consequently, the receiver sees the fact only through the message. Receiver-only outputs, scores, and thresholds are reused in a separate 20-test family. The message changes 12.4\% of calibration answers (170 of 1,366; plain Text 10.5\%) and 10.2\% of development answers (AUROC 0.889). The test deploys $q=0.75$ (22 of 999, $p=4.8\times10^{-6}$; $q=0.80$ fails, 33 of 1,064, $p=0.0016$), also without the 208 reviewed questions (20 of 852). The certificate is nominal because ProbeMax was chosen on the plain Text labels of the same questions; held out, 16 of 558 skipped answers change ($p=9.2\times10^{-3}$). On development, the policy skips 536 of 742 questions and changes 13. The message corrects 62 development answers and breaks 9; the policy skips 11 of the corrections and 1 of the breaks, answering 660 of 742 correctly against 670 for Text+fact and 617 for $R$ ($-1.35$ points [$-2.29$, $-0.54$], within $\kappa\alpha\approx3.6$). This is the only deployed policy whose development accuracy interval lies below zero.

\paragraph{A stronger helper.} Under a protocol fixed before its outputs, the large helper's stored messages were given to the medium receiver on OBQA and ARC (40 tests). The helper alone answers 86.3 and 87.6\% of development questions correctly (receiver: 66.0 and 72.6\%). Its messages raise the receiver's accuracy by 7.3 and 4.3 points but change 21.2 and 13.4\% of its development answers (calibration: 297 of 1,366 and 55 of 448; AUROC 0.792 and 0.800). Both settings fall back (smallest $p=0.26$ and 0.026); they certify in none and 15.5\% of 200 re-drawn splits, and would need about 21,500 and 880 calibration questions to pass.

\subsection{Extractive questions: SQuAD}\label{app:squad}

\begin{table}[!htbp]\centering\scriptsize
\caption{The label-free pilot makes three Llama-8B settings eligible and selects SQuAD with the Llama-8B receiver, whose predicted probability of certification is the highest (400 questions per setting). Helper: the large helper; TriviaQA and NQ-Open: closed-book. Disagreement: receiver-only versus Text. AUROC of the first-token, whole-answer, and per-token-mean scores (one minus the top probability at the first alphanumeric token, one minus the whole answer's probability, and one minus its per-token geometric mean). Changes in lowest 25\%: changes among the 100 lowest-score questions (ranked by the score with the highest pilot AUROC, the whole-answer score in all eight settings); Eligible: at most 4\% of these change. P(certify): certification probability predicted by the frozen binormal model (Appendix~\ref{app:predict}) at 1,000, 2,000, and 3,000 calibration questions. Contain: share of changes in which one normalized answer contains the other. No answer was \textsc{invalid}.}
\label{tab:squad_pilot}
\setlength{\tabcolsep}{2pt}
\begin{tabular*}{\linewidth}{@{}l@{\hspace{6pt}}l@{\hspace{4pt}\extracolsep{\fill}}lcccrccccc@{}}\toprule
& & & \multicolumn{3}{c}{AUROC} & & & \multicolumn{3}{c}{\begin{tabular}[b]{@{}c@{}}P(certify) by\\calibration size\end{tabular}} & \\\cmidrule(lr){4-6}\cmidrule(lr){9-11}
Dataset & Receiver & \multicolumn{1}{c}{\begin{tabular}[b]{@{}c@{}}Disagreement\\(\%) [95\% CI]\end{tabular}} & \begin{tabular}[b]{@{}c@{}}First\\token\end{tabular} & \begin{tabular}[b]{@{}c@{}}Whole\\answer\end{tabular} & \begin{tabular}[b]{@{}c@{}}Per-token\\mean\end{tabular} & \multicolumn{1}{c}{\begin{tabular}[b]{@{}c@{}}Changes in\\lowest 25\%\end{tabular}} & Eligible & 1,000 & 2,000 & 3,000 & \multicolumn{1}{c@{}}{\begin{tabular}[b]{@{}c@{}}Contain\\(\%)\end{tabular}}\\\midrule
SQuAD & Qwen3-8B & 24.5 [20.4, 29.0] & 0.751 & 0.805 & 0.791 & 7 of 100 & no & 0.022 & 0.075 & 0.192 & 68\\
SQuAD & Llama-8B & 14.3 [11.0, 18.1] & 0.773 & 0.836 & 0.738 & 1 of 100 & yes & 0.598 & 0.975 & 1.000 & 60\\
NQ (passage) & Qwen3-8B & 28.3 [23.9, 32.9] & 0.719 & 0.745 & 0.743 & 12 of 100 & no & 0.000 & 0.000 & 0.002 & 51\\
NQ (passage) & Llama-8B & 22.3 [18.3, 26.6] & 0.742 & 0.808 & 0.754 & 3 of 100 & yes & 0.039 & 0.179 & 0.393 & 44\\
TriviaQA & Qwen3-8B & 40.5 [35.7, 45.5] & 0.746 & 0.792 & 0.784 & 9 of 100 & no & 0.000 & 0.000 & 0.002 & 16\\
TriviaQA & Llama-8B & 29.0 [24.6, 33.7] & 0.803 & 0.838 & 0.821 & 4 of 100 & yes & 0.035 & 0.162 & 0.413 & 15\\
NQ-Open & Qwen3-8B & 62.3 [57.3, 67.0] & 0.707 & 0.793 & 0.760 & 30 of 100 & no & 0.000 & 0.000 & 0.000 & 10\\
NQ-Open & Llama-8B & 48.3 [43.3, 53.3] & 0.792 & 0.856 & 0.806 & 9 of 100 & no & 0.000 & 0.006 & 0.024 & 17\\
\bottomrule\end{tabular*}
\end{table}

\begin{table}[!htbp]\centering\scriptsize
\caption{SQuAD certifies: the test accepts $q=0.15$ to $0.35$ and deploys $q=0.35$. Threshold $\tau$ on the whole-answer score; skipped ($n$) and changed ($k$) calibration questions; exact binomial $p$ for $H_0:\rho\ge 0.05$; CP: Clopper--Pearson $0.999$ upper bound on $\rho$ (diagnostic). Accepted in bold.}
\label{tab:squad_cal}
\setlength{\tabcolsep}{3.6pt}\renewcommand{\arraystretch}{1.2}
\begin{tabular}{llrrll@{\hspace{12pt}}llrrll}\toprule
\multicolumn{1}{c}{$q$} & \multicolumn{1}{c}{$\tau$} & \multicolumn{1}{c}{$n$} & \multicolumn{1}{c}{$k$} & \multicolumn{1}{c}{$p$} & \multicolumn{1}{c@{\hspace{12pt}}}{CP} & \multicolumn{1}{@{}c}{$q$} & \multicolumn{1}{c}{$\tau$} & \multicolumn{1}{c}{$n$} & \multicolumn{1}{c}{$k$} & \multicolumn{1}{c}{$p$} & \multicolumn{1}{c}{CP}\\\midrule
0.05 & 0.000188 & 111 & 2 & 0.080 & 0.097 & 0.55 & 0.350 & 1,166 & 58 & 0.519 & 0.073\\
0.10 & 0.000655 & 212 & 3 & 0.0057 & 0.060 & 0.60 & 0.398 & 1,261 & 72 & 0.887 & 0.080\\
\textbf{0.15} & 0.00161 & 307 & 3 & $1.2\times10^{-4}$ & 0.042 & 0.65 & 0.433 & 1,322 & 76 & 0.903 & 0.080\\
\textbf{0.20} & 0.00492 & 409 & 4 & $8.3\times10^{-6}$ & 0.036 & 0.70 & 0.469 & 1,390 & 85 & 0.973 & 0.084\\
\textbf{0.25} & 0.0161 & 519 & 7 & $7.9\times10^{-6}$ & 0.037 & 0.75 & 0.511 & 1,501 & 107 & 0.9999 & 0.094\\
\textbf{0.30} & 0.0421 & 592 & 11 & $6.1\times10^{-5}$ & 0.043 & 0.80 & 0.559 & 1,607 & 138 & 1.000 & 0.110\\
\textbf{0.35} & 0.0992 & 721 & 17 & $2.5\times10^{-4}$ & 0.047 & 0.85 & 0.615 & 1,696 & 158 & 1.000 & 0.117\\
0.40 & 0.167 & 847 & 29 & 0.017 & 0.058 & 0.90 & 0.686 & 1,806 & 196 & 1.000 & 0.133\\
0.45 & 0.230 & 952 & 40 & 0.145 & 0.066 & 0.95 & 0.792 & 1,908 & 236 & 1.000 & 0.149\\
0.50 & 0.298 & 1,063 & 48 & 0.260 & 0.068 & 1 & $\infty$ & 2,000 & 302 & 1.000 & 0.177\\\bottomrule
\end{tabular}
\end{table}

\paragraph{Design and selection.} A pilot protocol fixed before any pilot output defined eight candidates (SQuAD and NQ with passages, TriviaQA and NQ-Open closed-book \citep{rajpurkar2016squad,kwiatkowski2019nq,fisch2019mrqa,joshi2017triviaqa,lee2019latent}, each with a Qwen3-8B and a Llama-8B receiver and the large helper), three scores, and a label-free selection rule. A full-run protocol fixed before any full-run output added a 1,000-question test split and narrowed the reporting rule, so that an unqualified claim for extractive questions required a passing test as well as a certificate (recorded deviations from the pilot protocol); the certificate was fixed before any development or test output existed. For SQuAD, the training pool is deduplicated and limited to passages of at most 300 words (86,830 questions), and all splits are disjoint blocks of a fixed pseudorandom order. The receiver answers directly with at most 32 new tokens, and answers are compared using official SQuAD normalization. The three scores are one minus the top probability at the first generated token containing a letter or digit (analogous to ProbeMax), one minus the whole answer's probability, and one minus its per-token geometric mean; the whole-answer score had the best pilot AUROC in all eight settings. It needs the receiver-only answer, which is served when the message is skipped and wasted otherwise. A setting was eligible if at most 4\% of its 25\% lowest-score pilot questions changed; eligible settings were ranked by the certification probability $P$ that the frozen binormal model (Appendix~\ref{app:predict}) predicts from the pilot's disagreement rate and AUROC, and at most two with $P\ge0.5$ would run, each at the smallest calibration size with $P\ge0.8$. The four Qwen3-8B settings and NQ-Open with the Llama-8B receiver are ineligible (7 to 30 of 100 changed), and the rule selected SQuAD with the Llama-8B receiver at 2,000 calibration questions ($P=0.975$, inflated by taking the best of eight; conservative 0.081) and no second setting (next: 0.413 and 0.393; Table~\ref{tab:squad_pilot}).

\paragraph{Results.} Before production, the runtime reproduced 16 pilot questions exactly. On calibration, the receiver disagrees with Text on 302 of 2,000 questions (15.1\%), and the test deploys $q=0.35$ (17 of 721, $p=2.5\times10^{-4}$; Table~\ref{tab:squad_cal}). All 200 re-drawn splits certify (median $q=0.375$, median development coverage 41.5\%). On development (disagreement 15.4\% [13.2, 17.8], AUROC 0.811 [0.771, 0.848]), the policy skips 39.6\% of questions, and 16 of 396 skipped answers change (4.04\% [2.3, 6.5]). In 15 of these, one normalized answer contains the other; counting containment as agreement leaves 1. On the sealed test, 9 of 364 skipped answers change (2.47\% [1.14, 4.64], $p=0.012$); the pass rule is a two-sided 95\% upper bound below 5\%, so the test passes, though $p$ exceeds 0.001. On development and test, $R$ answers 659 and 664 of 1,000 questions correctly, Text 669 and 682, and the policy 673 and 686. Among the skipped changes, Text would have corrected 4 and 2 answers and broken 8 and 6. Text's development gain, 1.0 point [$-0.9$, $+2.9$], is below $\kappa\alpha=2.0$, and its interval includes zero, so the protocol's redundancy rule holds. From recorded per-question times (sequential execution, on a different machine than the \S\ref{sec:rq3} replays), skipping saves an estimated 198.5~ms [167.2, 229.9] per question, while always skipping saves 785.4~ms (helper 772.5~ms; receiver 153.5~ms alone and 166.4~ms with the message, since the receiver-only answer is always computed). This is an estimate, not a replay.

\subsection{A numeric-answer check: GSM8K}\label{app:gsm}

A prespecified check (large pair, Text) used 400 fit, 1,000 calibration, and 400 development training questions in a fixed content-based order (a recorded deviation from the main protocol's seeded shuffle) and raised the receiver's limit from 64 to 320 tokens (a deviation fixed in advance), because the receiver derives before answering. A change is a mismatch of the normalized final number (after the last ``answer is'', otherwise the last number; the extractor was tested on twelve hand-written strings and, as prespecified, never changed). The score is one minus the length-normalized probability of the receiver's answer tokens under teacher forcing, which reduces to ProbeMax for a single label token. No candidate passes (every $p=1.000$), although all 3,200 outputs parse and the receiver disagrees with Text on 17.4\% of calibration and 19.3\% of development questions (accuracy 87.75\% alone, 86.50\% with Text). The score fails: 378 of 400 fit questions score exactly zero after the derivation, so 18 of the 20 thresholds coincide. Its development AUROC is 0.5795 (0.6444 for the well-spread whole-continuation probability, 0.5680 for the first answer token). The last-number rule fired on 15.2\% of receiver-only outputs, tracking the 320-token cap (reached by 43 fit, 106 calibration, and 49 development outputs); the Text path is barely affected (6 of 1,000 and 2 of 400). A later protocol, written after this fallback, planned a score over the first 32 generated tokens but stopped at its reproduction check (differences up to 0.04 and 0.06 on two capped outputs, against a tolerance of $10^{-4}$), so no prefix score was tested (Table~\ref{tab:provenance}).

\subsection{Sampled receiver answers}\label{app:sampling}

Post hoc, under a protocol fixed before its outputs, the large pair's OBQA and ARC Text settings were rerun with receiver answers sampled on both paths (the Qwen3 model card's settings with thinking disabled: temperature 0.7, top-$p$ 0.8, top-$k$ 20, min-$p$ 0; one sample per question and path, seeded independently), with messages, scores, thresholds, and the test unchanged in a 40-test family (OBQA stays nominal). Both settings deploy at the greedy thresholds ($q=0.80$: 19 of 1,064, $p=3.4\times10^{-8}$; $q=0.95$: 4 of 422, $p=4.8\times10^{-6}$) and in all 200 re-drawn splits, and 3.32\% [2.01, 5.14] and 2.82\% [1.22, 5.47] of skipped development answers change. Sampled disagreement is 9.4 and 4.7\% (AUROC 0.867 and 0.956; greedy: 9.7 and 5.0\%), most skipped questions still have zero uncertainty (354 of 572 and 217 of 284), and accuracy is 617 and 646 of 742 (receiver alone, Text) and 268 and 275 of 299. Sampling changes few of this receiver's labels (6 of 742 receiver-only and 2 Text answers on OBQA, 0 and 1 of 299 on ARC), so this check says nothing about receivers whose samples vary more or that generate reasoning tokens.

%% file: new_app_H.tex
\section{Out-of-sample tests}\label{app:sealed}

\begin{figure}[!t]\centering
\includegraphics[width=\linewidth]{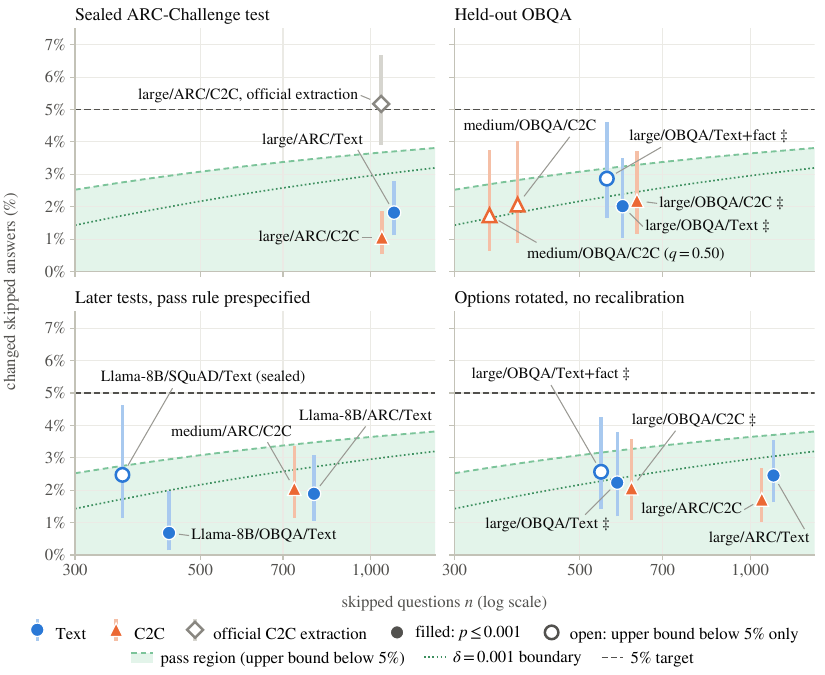}
\caption{Every frozen policy tested out of sample keeps the 95\% upper bound of its change rate below 5\% under our parsers; only the sealed C2C policy read with the official C2C extraction exceeds it. Each marker is one frozen run on questions its pair had not used (official extraction: the sealed C2C run re-read), placed at its number of skipped questions $n$ (log scale) and the share of skipped answers that change, with its two-sided 95\% Clopper--Pearson interval as a vertical line; the counts are in the tables below. Shaded region under the dashed curve: rates whose upper bound is below 5\% at that $n$ (pass rule: a 95\% upper bound below 5\%); dotted curve: the largest rate at which the test rejects $\rho\ge0.05$ at $\delta=0.001$; dashed horizontal line: the 5\% target. Filled: the test rejects at $\delta=0.001$; open: only the upper bound is below 5\% (level 0.025). Official C2C extraction (gray diamond): the official C2C evaluator's answer extraction applied to the stored outputs. $q=0.50$: sensitivity threshold for medium/OBQA/C2C (deployed $q=0.55$). Options rotated: each option's text moved one position down, labels kept. $^\ddagger$: nominal development certificate (Appendix~\ref{app:history}). Where a Text and a C2C run share $n$, the C2C marker is drawn slightly to the right. The rotated runs set no pass rule, and their p-values are descriptive. Every statistic is in Tables~\ref{tab:e2e}, \ref{tab:holdout}, \ref{tab:oos_later}, and~\ref{tab:rotation} and Appendices~\ref{app:sealed} and~\ref{app:squad}.}
\label{fig:oos}
\end{figure}

Every frozen policy tested out of sample keeps its change rate's 95\% upper bound below 5\% under our parsers (all eleven tests: Figure~\ref{fig:oos}, Tables~\ref{tab:e2e} and~\ref{tab:holdout}). Each ran once, with routes and outputs saved before gold labels were decoded. Only the large-pair ARC and SQuAD tests are sealed, and MMLU-Pro has no unused questions. These runs are not a second calibration. The sealed and held-out protocols define no pass rule, whereas the later ones do. The contamination caveat of Appendix~\ref{app:restson} applies.

\paragraph{Sealed ARC test.} The population is all 1,172 ARC-Challenge test rows of \texttt{allenai/ai2\_arc} (1,172 groups after frozen normalization), with none removed. The sealing order was: (1) a freeze recorded that the test population had not been accessed and sealed the protocol, configuration, source index, and analysis snapshot; (2) only the \texttt{id}, \texttt{question}, and \texttt{choices} were projected; (3) after all scores, routes, outputs, probe records, and timings were written, a second freeze locked them; (4) only then was the answer key decoded. The thresholds are the development ones ($q=0.95$ and $0.90$), and no test quantile was computed. The four paths (fixed Text, Text policy, fixed C2C, C2C policy) were interleaved in one allocation, completing all 4,688 requests and 2,344 probes. The two policies' independent probes had identical inputs and scores. All 197 reference-routed outputs match the fixed-reference path, and \textsc{invalid} counts were 0/0 (Text) and 1/1 (C2C). The Text and C2C policies change 20 of 1,100 and 11 of 1,047 skipped answers ([1.11\%, 2.79\%] and [0.53\%, 1.87\%]; 1.71 and 0.94\% of all questions). Relative to the reference, they gain 7 and lose 12 correct answers (Text), and gain 7 and lose 3 (C2C). The probe costs 42.5 and 40.5~ms (accuracy and latency: Table~\ref{tab:e2e_full}). On questions below the frozen thresholds, the sealed split does not significantly differ from the training split (Fisher $p=0.65$ and $0.68$). Its change rates are no higher than on development (Text 1.8 vs.\ 3.2\%, C2C 1.05 vs.\ 1.11\%).

\paragraph{Sealed ARC under the official extraction.} Under a rule fixed before computing, we re-read the stored sealed outputs using the official C2C extraction on the 1,169 questions lacking a fifth option (both policies skip the other three, none of which changes under our parser). Two unparseable answers count as agreement; counting them as changes leaves every count unchanged. The Text policy is unaffected (20 of 1,097 skipped answers change under both extractions, 1.82\% [1.12, 2.80]). However, 54 of 1,044 skipped C2C answers change (5.17\% [3.91, 6.70]; $p=0.635$), against 11 under our parser. Labels differ on 44 skipped questions, accounting for all 43 additional changes. A mechanical audit, fixed before computing, found that each of these 44 outputs starts with a single option letter followed by that option's own text, contains no second option label, and receives that letter from our parser. The official extraction marks every one unparseable because the option text contains a string treated as a math term (\texttt{tan} in 17, a minus sign in 10, \texttt{sin} in 7, a plus sign or slash in 4 each, \texttt{cos} in 3, and \texttt{mod} in 2; some contain two). The Text policies remain nearly unaffected elsewhere: on the held-out OBQA questions, the official extraction changes 11 of 595 skipped large-pair Text answers (1.85\% [0.93, 3.28], against 12 under our parser) and 16 of 558 Text+fact answers, matching our parser. As fixed in advance for a point estimate above 5\%, we state that the sealed C2C result holds under our parser only. The official C2C release includes an ARC test configuration, but we could not verify whether its fusers were exposed to these questions.

\paragraph{Held-out OBQA.} These are the 744 OBQA questions that the project's partition never assigned to fit, calibration, or development. The protocol's gate initially stopped the run because an earlier project stage had generated small-pair outputs on them under four communication channels and scored them against the answer key. Before any output existed, the protocol was amended to report them as held out rather than sealed, after two checks verified that no output, score, route, or latency of the medium or large pair existed for these questions, and that no artifact of this study reads the earlier stage's predictions or scores (two aggregate small-pair accuracies and two rows of the earlier scoring were read during the gate check, as the amendment, fixed before the run, discloses). Only \texttt{id}, \texttt{question}, and \texttt{choices} (plus \texttt{fact1} for Text+fact) were projected. Saved fit rows were reproduced bit for bit, and one shard was rerun after a bookkeeping failure that produced no output. With no pass rule set, all five upper bounds fall below 5\% (level 0.025). Only large Text and C2C (two nominal settings) also reject at $\delta=0.001$ ($p=1.5\times10^{-4}$ and $3.7\times10^{-4}$; with Bonferroni over the five tests, their family level is 0.005). Text+fact, also nominal, rejects only at 0.025. Always skipping would change 8.2 to 17.5\% of answers (Table~\ref{tab:holdout}). The two medium thresholds reach identical accuracy: of the 42 questions whose skip status differs, only two have an answer that C2C would change, and these cancel out. No latency was measured.

\begin{table}[!htbp]\centering\scriptsize
\caption{All five held-out OBQA tests keep the 95\% upper bound below 5\%, and large Text and C2C also reject at $\delta=0.001$ (744 questions). Change rate with its two-sided Clopper--Pearson 95\% CI; $p$: single prespecified test of $H_0:\rho\ge 0.05$; Level: stricter level at which $H_0$ is rejected (0.025: upper end below 5\%); Always skip: change rate of answering alone; (b) accuracy of the policy and the fixed reference, and their difference $\Delta$acc with its paired-bootstrap 95\% CI; pp: percentage points. The medium $q=0.50$ row is a sensitivity threshold (deployed $q=0.55$).}
\label{tab:holdout}
\renewcommand{\arraystretch}{1.1}\setlength{\tabcolsep}{3pt}
\vspace{5pt}\textit{(a) Changed answers}\par\smallskip
\begin{tabular*}{\linewidth}{@{}l@{\hspace{5pt}\extracolsep{\fill}}ccccc@{}}\toprule
Policy & \begin{tabular}[b]{@{}c@{}}Changed\\/ skipped\end{tabular} & \begin{tabular}[b]{@{}c@{}}Change rate (\%)\\{}[95\% CI]\end{tabular} & $p$ & Level & \begin{tabular}[b]{@{}c@{}}Always\\skip (\%)\end{tabular}\\\midrule
medium/OBQA/C2C, $q=0.55$ & 8/388 & 2.06 [0.89, 4.02] & $2.5\times10^{-3}$ & 0.025 & 17.47\\
medium/OBQA/C2C, $q=0.50$ & 6/346 & 1.73 [0.64, 3.74] & $1.4\times10^{-3}$ & 0.025 & 17.47\\
large/OBQA/Text, $q=0.80$ & 12/595 & 2.02 [1.05, 3.50] & $1.5\times10^{-4}$ & 0.001 & 8.20\\
large/OBQA/C2C, $q=0.80$ & 13/595 & 2.18 [1.17, 3.71] & $3.7\times10^{-4}$ & 0.001 & 8.33\\
large/OBQA/Text+fact, $q=0.75$ & 16/558 & 2.87 [1.65, 4.61] & $9.2\times10^{-3}$ & 0.025 & 11.02\\\bottomrule
\end{tabular*}
\par\vspace{17pt}
\textit{(b) Accuracy}\par\smallskip
\begin{tabular*}{\linewidth}{@{}l@{\hspace{5pt}\extracolsep{\fill}}cccc@{}}\toprule
& \multicolumn{2}{c}{Accuracy (\%)} & \multicolumn{2}{c}{$\Delta$acc, policy $-$ reference (pp)}\\\cmidrule(lr){2-3}\cmidrule(l){4-5}
Policy & Policy & Reference & & 95\% CI\\\midrule
medium/OBQA/C2C, $q=0.55$ & 61.16 & 60.89 & $+0.27$ & [$-0.54$, $+1.08$]\\
medium/OBQA/C2C, $q=0.50$ & 61.16 & 60.89 & $+0.27$ & [$-0.40$, $+0.94$]\\
large/OBQA/Text, $q=0.80$ & 86.02 & 86.96 & $-0.94$ & [$-1.88$, $-0.13$]\\
large/OBQA/C2C, $q=0.80$ & 81.45 & 80.24 & $+1.21$ & [$+0.40$, $+2.15$]\\
large/OBQA/Text+fact, $q=0.75$ & 89.65 & 91.13 & $-1.48$ & [$-2.55$, $-0.54$]\\\bottomrule
\end{tabular*}
\end{table}

\paragraph{Later out-of-sample tests.} The two Llama-8B policies (using the large pair's stored helper messages) and medium/ARC/C2C ran once on questions their respective pairs had never seen. They ran under a protocol fixed before any output, whose pass rule was a 95\% upper bound below 5\%. Before the runs, a gate confirmed that no output of the Llama-3.1-8B receiver or the medium pair's models existed for them. All three pass and reject at 0.001, with coverage within 1.2 points of development (Table~\ref{tab:oos_later}). This holds under the official extraction too (3 of 440, 15 of 792, and 16 of 730 changes on the A--D questions). One analysis run stopped after writing the seal receipts, so only the gold stage was recomputed from the sealed route files. The SQuAD test is in Appendix~\ref{app:squad}.

\begin{table}[!htbp]\centering\scriptsize
\caption{All three later out-of-sample tests pass their prespecified rule and reject at 0.001, with test coverage within 1.2 points of development coverage. Pass rule: a 95\% upper bound below 5\%. ARC test: held out for these pairs (Table~\ref{tab:e2e}). Change rate with its Clopper--Pearson 95\% CI; $p$: exact-test p-value; always skip: disagreement over the whole population; (b) accuracy in percent and $\Delta$acc, policy minus reference, with its paired-bootstrap 95\% CI; pp: percentage points.}
\label{tab:oos_later}
\renewcommand{\arraystretch}{1.1}\setlength{\tabcolsep}{2.5pt}
\vspace{5pt}\textit{(a) Changed answers}\par\smallskip
\begin{tabular*}{\linewidth}{@{}l@{\hspace{5pt}\extracolsep{\fill}}ccccccc@{}}\toprule
& & \multicolumn{2}{c}{Coverage (\%)} & & & & \\\cmidrule(lr){3-4}
Policy & \begin{tabular}[b]{@{}c@{}}Test\\questions\end{tabular} & Test & \begin{tabular}[b]{@{}c@{}}Develop-\\ment\end{tabular} & \begin{tabular}[b]{@{}c@{}}Changed\\/ skipped\end{tabular} & \begin{tabular}[b]{@{}c@{}}Change rate (\%)\\{}[95\% CI]\end{tabular} & $p$ & \begin{tabular}[b]{@{}c@{}}Always\\skip (\%)\end{tabular}\\\midrule
Llama-8B/OBQA/Text, $q=0.60$ & held-out OBQA (744) & 59.1 & 59.0 & 3/440 & 0.68 [0.14, 1.98] & $3.7\times10^{-7}$ & 11.96\\
Llama-8B/ARC/Text, $q=0.70$ & ARC test (1,172) & 67.7 & 68.9 & 15/794 & 1.89 [1.06, 3.10] & $4.4\times10^{-6}$ & 10.75\\
medium/ARC/C2C, $q=0.60$ & ARC test (1,172) & 62.5 & 63.2 & 15/733 & 2.05 [1.15, 3.35] & $3.1\times10^{-5}$ & 14.85\\\bottomrule
\end{tabular*}
\par\vspace{17pt}
\textit{(b) Accuracy}\par\smallskip
\begin{tabular*}{\linewidth}{@{}l@{\hspace{5pt}\extracolsep{\fill}}ccccc@{}}\toprule
& \multicolumn{3}{c}{Accuracy (\%)} & \multicolumn{2}{c}{$\Delta$acc, policy $-$ reference (pp)}\\\cmidrule(lr){2-4}\cmidrule(l){5-6}
Policy & $R$ & Reference & Policy & & 95\% CI\\\midrule
Llama-8B/OBQA/Text, $q=0.60$ & 80.38 & 83.06 & 82.66 & $-0.40$ & [$-0.94$, 0.00]\\
Llama-8B/ARC/Text, $q=0.70$ & 81.31 & 84.73 & 84.56 & $-0.17$ & [$-0.77$, $+0.43$]\\
medium/ARC/C2C, $q=0.60$ & 75.94 & 72.18 & 72.53 & $+0.34$ & [$-0.26$, $+0.94$]\\\bottomrule
\end{tabular*}
\end{table}

\paragraph{Option rotation.} With every option's text moved one position down (labels in place, as in the null control of Appendix~\ref{app:complementarity}), the five frozen large-pair policies ran once on the sealed ARC and held-out OBQA questions. They ran under a protocol fixed before any rotated output, with no recalibration and no gold label read; all outputs were written before the protocol's cutoff. With the rotation set to the identity, the run reproduced the stored scores bit for bit and the stored outputs byte for byte on 16 questions per population. The rotation check included 28 ARC questions with numeric labels or other than four options. The policies skip 73.3 to 93.9\% of questions and change 1.7 to 2.6\% of skipped answers, and every 95\% upper bound stays below 5\% (Table~\ref{tab:rotation}). Receiver-only answers before and after rotation agree on 98.7\% (ARC) and 98.0\% (OBQA) of questions with $u=0$ in the original order, against 85.9 and 75.9\% for the others. Of the $u=0$ questions, 90.0 and 77.7\% keep $u=0$. References agree with their original-order answers on 79.2 to 95.7\% of questions, with 0 to 2 unparseable answers per arm. All 27 rotated ARC/Text changes fall on questions with $u>0$, as in the original order.

\begin{table}[!htbp]\centering\scriptsize
\caption{With the options rotated by one position, every frozen large-pair policy keeps its 95\% upper bound below 5\% (one run, no recalibration; 1,172 ARC and 744 OBQA questions). Option rotation: each option's text moved one position down, labels kept. $u=0$: share of questions after rotation. Changed at $u=0$, $u>0$: changes among skipped questions by their $u$ on the rotated input. Intervals: Clopper--Pearson 95\% CI. $p$: exact-test p-value, descriptive. Original order: changed and skipped answers without rotation (Tables~\ref{tab:e2e} and~\ref{tab:holdout}).}
\label{tab:rotation}
\renewcommand{\arraystretch}{1.1}\setlength{\tabcolsep}{2.5pt}
\begin{tabular*}{\linewidth}{@{}l@{\hspace{5pt}\extracolsep{\fill}}cccccccccc@{}}\toprule
& & & \multicolumn{2}{c}{Coverage (\%)} & & & & \multicolumn{2}{c}{Changed at} & \\\cmidrule(lr){4-5}\cmidrule(lr){9-10}
Policy & \begin{tabular}[b]{@{}c@{}}$u=0$\\(\%)\end{tabular} & Skipped & rotated & original & Changed & \begin{tabular}[b]{@{}c@{}}Change rate (\%)\\{}[95\% CI]\end{tabular} & $p$ & $u=0$ & $u>0$ & \begin{tabular}[b]{@{}c@{}}Original order:\\changed / skipped\end{tabular}\\\midrule
large/ARC/Text & 72.1 & 1,100 & 93.9 & 93.9 & 27 & 2.45 [1.62, 3.55] & $1.5\times10^{-5}$ & 0 & 27 & 20/1,100\\
large/ARC/C2C & 72.1 & 1,049 & 89.5 & 89.3 & 18 & 1.72 [1.02, 2.70] & $1.9\times10^{-8}$ & 3 & 15 & 11/1,047\\
large/OBQA/Text & 46.1 & 582 & 78.2 & 80.0 & 13 & 2.23 [1.19, 3.79] & $5.4\times10^{-4}$ & 1 & 12 & 12/595\\
large/OBQA/C2C & 46.1 & 582 & 78.2 & 80.0 & 12 & 2.06 [1.07, 3.57] & $2.2\times10^{-4}$ & 1 & 11 & 13/595\\
large/OBQA/Text+fact & 46.1 & 545 & 73.3 & 75.0 & 14 & 2.57 [1.41, 4.27] & $3.4\times10^{-3}$ & 2 & 12 & 16/558\\\bottomrule
\end{tabular*}
\end{table}

%% file: new_app_I.tex
\section{What the certificates rest on}\label{app:restson}

The Qwen certificates rest on questions with exactly zero uncertainty, yet certification does not require them: the Llama-8B receiver agent has none, yet is certified and holds out of sample. Unless marked post hoc, each analysis below ran under rules fixed before computing. The unsaturated-AUROC and receiver-rotation analyses and both unparseable-answer variants are post hoc (Table~\ref{tab:provenance}).

\begin{figure}[!t]\centering
\includegraphics[width=\linewidth]{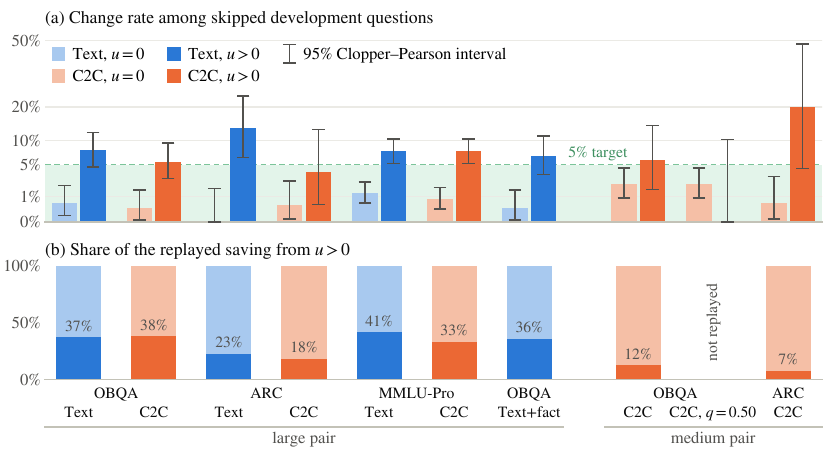}
\caption{The Qwen certificates rest on zero-uncertainty questions: skipped questions with $u=0$ rarely change, whereas those with $u>0$ change more often and supply 7 to 41\% of the replayed saving. One column per setting in both panels, named below (b). (a)~Change rate among skipped development questions with $u=0$ (light bars) and $u>0$ (dark bars), in the setting's colour (blue: Text and Text+fact; orange: C2C), with two-sided Clopper--Pearson 95\% intervals as whiskers (a rate of 0 has no bar, only its interval), at the nine certified Qwen-receiver thresholds and the medium/OBQA/C2C sensitivity threshold $q=0.50$ (Table~\ref{tab:zero_u}); square-root scale; green band: at or below the 5\% target. (b)~Share of the replayed saving on skipped questions that comes from $u>0$ questions (dark lower part, value printed; light upper part: the $u=0$ share; Table~\ref{tab:u0saving}; not replayed: no replay at $q=0.50$); the Llama-8B receiver (not plotted), which has no $u=0$ question, draws all of its saving from them.}
\label{fig:zero_u}
\end{figure}

\paragraph{Zero-uncertainty questions among skipped ones.} Over the nine certified Qwen-receiver thresholds, 60.3 to 92.1\% of skipped development questions have $u=0$. These change 0 to 2.2\% of the time, against 3.8 to 20.0\% for the others (Table~\ref{tab:zero_u}; Figure~\ref{fig:zero_u}a). For large/ARC/Text and both large MMLU-Pro policies, the $u>0$ part has a Clopper--Pearson lower bound above 5\%. Thus, their certificates rest on the zeros, and on calibration the $u>0$ part alone would pass at no frozen threshold ($p=0.175$ to $0.999$). For medium/ARC/C2C and medium/OBQA/C2C at $q=0.50$, over 90\% of skipped questions have $u=0$, so contamination \citep{deng2024contamination} could inflate such coverage without voiding a certificate. Out of sample, $u>0$ questions also change more often. On the held-out OBQA questions, 11 of 249 (large Text and C2C), 3 of 82 and 1 of 40 (medium C2C at $q=0.55$ and $0.50$), and 15 of 212 (Text+fact) of the skipped $u>0$ answers change, against 1 to 5 of 306 to 346 zero-uncertainty ones. On the sealed ARC test, every change falls on $u>0$ questions (Text 20 of 263, C2C 11 of 210); none of the 837 zero-uncertainty questions each policy skipped changed. Over all development questions, $u=0$ covers 43.3, 58.2, and 12.3\% for the medium receiver and 47.7, 72.6, and 24.4\% for the large receiver on OBQA, ARC, and MMLU-Pro.

\begin{table}[!htbp]\centering\scriptsize
\caption{At the nine deployed thresholds, skipped questions with $u=0$ change 0 to 2.2\% of the time and the others 3.8 to 20.0\% (development). $n$: skipped questions; share: $n_0/n$; $n_0$, $k_0$: skipped questions with $u=0$ and their changes; $n_+$, $k_+$: the others [Clopper--Pearson 95\% CI]. $p_+$: calibration p-value of the $u>0$ part alone (descriptive). Medium/OBQA/C2C is deployed at $q=0.55$; its $q=0.50$ row is a sensitivity threshold, accepted without the parser-review questions (341 calibration questions whose outputs were read with gold labels before the parser was frozen).}
\label{tab:zero_u}
\renewcommand{\arraystretch}{1.1}\setlength{\tabcolsep}{2.5pt}
\begin{tabular*}{\linewidth}{@{}l@{\hspace{5pt}\extracolsep{\fill}}lrcclcll@{}}\toprule
& & & \multicolumn{3}{c}{Skipped questions with $u=0$} & \multicolumn{3}{c}{Skipped questions with $u>0$}\\\cmidrule(lr){4-6}\cmidrule(l){7-9}
Setting & \multicolumn{1}{c}{$q$} & \multicolumn{1}{c}{$n$} & share (\%) & $k_0/n_0$ & \multicolumn{1}{c}{\begin{tabular}[b]{@{}c@{}}change rate (\%)\\{}[95\% CI]\end{tabular}} & $k_+/n_+$ & \multicolumn{1}{c}{\begin{tabular}[b]{@{}c@{}}change rate (\%)\\{}[95\% CI]\end{tabular}} & \multicolumn{1}{c}{\begin{tabular}[b]{@{}c@{}}calibration\\$p_+$\end{tabular}}\\\midrule
large/OBQA/Text & 0.80 & 572 & 61.9 & 2/354 & 0.56 [0.07, 2.03] & 17/218 & 7.80 [4.61, 12.19] & 0.214\\
large/OBQA/C2C & 0.80 & 572 & 61.9 & 1/354 & 0.28 [0.01, 1.56] & 12/218 & 5.50 [2.88, 9.42] & 0.46\\
large/ARC/Text & 0.95 & 284 & 76.4 & 0/217 & 0.00 [0.00, 1.69] & 9/67 & 13.43 [6.33, 23.97] & 0.427\\
large/ARC/C2C & 0.90 & 270 & 80.4 & 1/217 & 0.46 [0.01, 2.54] & 2/53 & 3.77 [0.46, 12.98] & 0.765\\
medium/OBQA/C2C & 0.55 & 390 & 82.3 & 7/321 & 2.18 [0.88, 4.44] & 4/69 & 5.80 [1.60, 14.18] & 0.979\\
medium/OBQA/C2C & 0.50 & 355 & 90.4 & 7/321 & 2.18 [0.88, 4.44] & 0/34 & 0.00 [0.00, 10.28] & 0.601\\
medium/ARC/C2C & 0.60 & 189 & 92.1 & 1/174 & 0.57 [0.01, 3.16] & 3/15 & 20.00 [4.33, 48.09] & 0.175\\
large/MMLU-Pro/Text & 0.40 & 1069 & 60.3 & 8/645 & 1.24 [0.54, 2.43] & 32/424 & 7.55 [5.22, 10.49] & 0.877\\
large/MMLU-Pro/C2C & 0.40 & 1069 & 60.3 & 5/645 & 0.78 [0.25, 1.80] & 32/424 & 7.55 [5.22, 10.49] & 0.999\\
large/OBQA/Text+fact & 0.75 & 536 & 66.0 & 1/354 & 0.28 [0.01, 1.56] & 12/182 & 6.59 [3.45, 11.23] & 0.747\\\bottomrule
\end{tabular*}
\end{table}

\begin{table}[!htbp]\centering\scriptsize
\caption{Without the zero-uncertainty block, ProbeMax still ranks disagreements with development AUROC 0.739 to 0.827. $u>0$: questions with nonzero uncertainty only, with a question-bootstrap 95\% CI and their disagreements / questions; tie-broken: ties broken by the log non-top mass; non-top mass: total probe probability of all labels but the top one; zero block: the $u=0$ questions; within $u=0$: that score inside the zero block (--: no disagreement there); $u=0$: share of questions; disagreements at $u>0$: share of all disagreements.}
\label{tab:auroc_u0}
\renewcommand{\arraystretch}{1.1}\setlength{\tabcolsep}{2.5pt}
\begin{tabular*}{\linewidth}{@{}l@{\hspace{5pt}\extracolsep{\fill}}cccccccc@{}}\toprule
& \multicolumn{4}{c}{Development AUROC} & \multicolumn{2}{c}{Within $u=0$} & & \\\cmidrule(lr){2-5}\cmidrule(lr){6-7}
Policy & all & $u>0$ [95\% CI] & \begin{tabular}[b]{@{}c@{}}disagreements\\/ questions\end{tabular} & tie-broken & AUROC & \begin{tabular}[b]{@{}c@{}}disagreements\\/ questions\end{tabular} & \begin{tabular}[b]{@{}c@{}}$u=0$\\(\%)\end{tabular} & \begin{tabular}[b]{@{}c@{}}Disagreements\\at $u>0$ (\%)\end{tabular}\\\midrule
medium/OBQA/C2C & 0.840 & 0.739 [0.690, 0.789] & 124/421 & 0.848 & 0.777 & 7/321 & 43.3 & 94.7\\
medium/ARC/C2C & 0.910 & 0.748 [0.658, 0.837] & 47/125 & 0.913 & 0.682 & 1/174 & 58.2 & 97.9\\
large/OBQA/Text & 0.873 & 0.769 [0.707, 0.826] & 70/388 & 0.876 & 0.753 & 2/354 & 47.7 & 97.2\\
large/OBQA/C2C & 0.883 & 0.779 [0.724, 0.831] & 59/388 & 0.884 & 0.697 & 1/354 & 47.7 & 98.3\\
large/ARC/Text & 0.952 & 0.796 [0.686, 0.896] & 15/82 & 0.952 & -- & 0/217 & 72.6 & 100\\
large/ARC/C2C & 0.909 & 0.827 [0.673, 0.954] & 11/82 & 0.931 & 0.861 & 1/217 & 72.6 & 91.7\\
large/MMLU-Pro/Text & 0.821 & 0.757 [0.734, 0.778] & 553/1996 & 0.822 & 0.723 & 8/645 & 24.4 & 98.6\\
large/MMLU-Pro/C2C & 0.846 & 0.774 [0.753, 0.795] & 751/1996 & 0.847 & 0.938 & 5/645 & 24.4 & 99.3\\
large/OBQA/Text+fact & 0.889 & 0.781 [0.728, 0.832] & 75/388 & 0.892 & 0.983 & 1/354 & 47.7 & 98.7\\\bottomrule
\end{tabular*}
\end{table}

\paragraph{Ranking without the saturated block.} Post hoc, on $u>0$ questions alone the development AUROC is 0.739 to 0.827 (median 0.774), against a median of 0.883 on all questions, as in Figure~\ref{fig:boundary_map}. With ties broken by the log of the non-top mass, the all-question AUROC is 0.822 to 0.952 (median 0.884; Table~\ref{tab:auroc_u0}); the $u>0$ questions hold 91.7 to 100\% of all disagreements. On the held-out OBQA questions the $u>0$ AUROC is 0.770 to 0.839; the tie-breaking score was not stored there.

\begin{table}[!htbp]\centering\scriptsize
\caption{Skipping exactly the $u=0$ questions passes in every setting ($\tau=0$ is an accepted grid threshold), and the deployed thresholds add skipped questions with $u>0$. Calibration: changed/skipped at $u=0$ and exact binomial p-value; development: coverage and change rate [Clopper--Pearson 95\% CI]. The medium/OBQA/C2C $q=0.50$ row is a sensitivity threshold (deployed: $q=0.55$).}
\label{tab:u0rule}
\renewcommand{\arraystretch}{1.1}\setlength{\tabcolsep}{2pt}
\begin{tabular*}{\linewidth}{@{}l@{\hspace{4pt}\extracolsep{\fill}}lrcccc@{}}\toprule
& \multicolumn{2}{c}{Calibration, $u=0$} & \multicolumn{2}{c}{Development, $u=0$} & \multicolumn{2}{c@{}}{Development, deployed}\\\cmidrule(lr){2-3}\cmidrule(lr){4-5}\cmidrule(l){6-7}
Setting & \multicolumn{1}{c}{$k_0/n_0$} & \multicolumn{1}{c}{$p$} & Coverage (\%) & \begin{tabular}[b]{@{}c@{}}Change rate (\%)\\{}[95\% CI]\end{tabular} & Coverage (\%) & \multicolumn{1}{c@{}}{\begin{tabular}[b]{@{}c@{}}Change rate (\%)\\{}[95\% CI]\end{tabular}}\\\midrule
large/OBQA/Text & 1/642 & $1.7\times10^{-13}$ & 47.7 & 0.56 [0.07, 2.03] & 77.1 & 3.32 [2.01, 5.14]\\
large/OBQA/C2C & 2/642 & $3.0\times10^{-12}$ & 47.7 & 0.28 [0.01, 1.56] & 77.1 & 2.27 [1.22, 3.86]\\
large/ARC/Text & 0/321 & $7.1\times10^{-8}$ & 72.6 & 0.00 [0.00, 1.69] & 95.0 & 3.17 [1.46, 5.93]\\
large/ARC/C2C & 2/321 & $1.1\times10^{-5}$ & 72.6 & 0.46 [0.01, 2.54] & 90.3 & 1.11 [0.23, 3.21]\\
large/MMLU-Pro/Text & 14/1442 & $2.1\times10^{-17}$ & 24.4 & 1.24 [0.54, 2.43] & 40.5 & 3.74 [2.69, 5.06]\\
large/MMLU-Pro/C2C & 9/1442 & $1.9\times10^{-21}$ & 24.4 & 0.78 [0.25, 1.80] & 40.5 & 3.46 [2.45, 4.74]\\
medium/OBQA/C2C ($q=0.55$) & 7/601 & $3.2\times10^{-7}$ & 43.3 & 2.18 [0.88, 4.44] & 52.6 & 2.82 [1.42, 4.99]\\
medium/OBQA/C2C ($q=0.50$) & 7/601 & $3.2\times10^{-7}$ & 43.3 & 2.18 [0.88, 4.44] & 47.8 & 1.97 [0.80, 4.02]\\
medium/ARC/C2C & 1/256 & $2.9\times10^{-5}$ & 58.2 & 0.57 [0.01, 3.16] & 63.2 & 2.12 [0.58, 5.33]\\
large/OBQA/Text+fact & 2/642 & $3.0\times10^{-12}$ & 47.7 & 0.28 [0.01, 1.56] & 72.2 & 2.43 [1.30, 4.11]\\\bottomrule
\end{tabular*}
\end{table}

\paragraph{Skipping exactly the zero-uncertainty questions.} In all nine settings, several low grid quantiles coincide at $\tau=0$ and that threshold is accepted. Therefore, the rule that skips a question if and only if $u=0$ was among the tested candidates (Table~\ref{tab:u0rule}). On development, it skips 24.4 to 72.6\% of questions and changes 0 to 2.2\% of them, against 40.5 to 95.0\% and 1.1 to 3.7\% for the deployed thresholds. With fit, calibration, and development restricted to $u>0$ questions and fit quantiles taken over them, the test certifies 1 of 9 settings: large/OBQA/C2C ($q=0.35$; 3 of 257 calibration questions change, $p=9.7\times10^{-4}$; 16.4\% of development questions skipped, 1 of 122 changed). The others fall back (smallest $p$ from 0.002 to 0.60).

\begin{table}[!htbp]\centering\scriptsize
\caption{Skipped questions with $u>0$ supply 7.3 to 41.3\% of the saving on skipped questions for the eight Qwen-receiver policies whose interval excludes zero, and all of it for the Llama-8B receiver. Paired mean saving (ms) by source, and the number $n$ of questions from each source, over the 128-question panels (Llama-8B and Text+fact on the second configuration, all others on the original one) and, $^\ast$, over the full 2,641-question MMLU-Pro C2C replay (second configuration); ms: each source's contribution to Saving; Routed: questions sent to the reference; Share: $u>0$ part over the skipped parts; Recomposed: saving if only $u=0$ questions were skipped.}
\label{tab:u0saving}
\renewcommand{\arraystretch}{1.1}\setlength{\tabcolsep}{2pt}
\begin{tabular*}{\linewidth}{@{}l@{\hspace{5pt}\extracolsep{\fill}}rrrrrrcrc@{}}\toprule
& \multicolumn{2}{c}{Skipped, $u=0$} & \multicolumn{2}{c}{Skipped, $u>0$} & \multicolumn{2}{c}{Routed} & & & \\\cmidrule(lr){2-3}\cmidrule(lr){4-5}\cmidrule(lr){6-7}
Policy & ms & $n$ & ms & $n$ & ms & $n$ & \begin{tabular}[b]{@{}c@{}}Saving (ms)\\{}[95\% CI]\end{tabular} & \begin{tabular}[b]{@{}c@{}}Share\\from $u>0$\end{tabular} & \begin{tabular}[b]{@{}c@{}}Recomposed (ms)\\{}[95\% CI]\end{tabular}\\\midrule
large/OBQA/Text & 332.9 & 60 & 194.5 & 37 & $-10.3$ & 31 & 517.1 [447.8, 583.6] & 0.369 & 294.9 [211.5, 372.4]\\
large/OBQA/C2C & 29.1 & 60 & 17.7 & 37 & $-10.2$ & 31 & 36.6 [18.0, 56.8] & 0.378 & 6.7 [$-9.5$, $+24.8$]\\
large/ARC/Text & 557.3 & 93 & 163.1 & 27 & $-2.8$ & 8 & 717.6 [670.6, 763.2] & 0.226 & 545.6 [476.1, 614.0]\\
large/ARC/C2C & 94.7 & 93 & 20.5 & 22 & $-4.2$ & 13 & 110.9 [82.0, 142.8] & 0.178 & 83.2 [55.2, 111.0]\\
large/MMLU-Pro/Text & 204.5 & 30 & 144.0 & 18 & $-2.2$ & 80 & 346.3 [249.8, 443.0] & 0.413 & 196.0 [113.6, 287.0]\\
large/MMLU-Pro/C2C & 23.2 & 30 & 11.6 & 18 & $-26.6$ & 80 & 8.2 [$-20.5$, $+44.0$] & 0.334 & $-9.8$ [$-33.7$, $+17.9$]\\
medium/OBQA/C2C & 67.0 & 55 & 9.4 & 11 & 3.2 & 62 & 79.7 [48.1, 129.1] & 0.123 & 67.4 [35.6, 115.2]\\
medium/ARC/C2C & 105.6 & 70 & 8.3 & 7 & $-13.3$ & 51 & 100.7 [74.5, 133.1] & 0.073 & 90.5 [63.5, 123.1]\\
large/OBQA/Text+fact & 224.5 & 60 & 124.3 & 32 & $-5.3$ & 36 & 343.4 [297.7, 386.0] & 0.356 & 211.2 [162.0, 258.0]\\
Llama-8B/OBQA/Text & 0.0 & 0 & 337.6 & 76 & $-19.0$ & 52 & 318.6 [261.1, 372.1] & 1.000 & $-37.7$ [$-48.3$, $-31.7$]\\
Llama-8B/ARC/Text & 0.0 & 0 & 427.3 & 88 & $-8.4$ & 40 & 418.9 [361.7, 473.7] & 1.000 & $-30.8$ [$-33.4$, $-27.8$]\\
\midrule
large/MMLU-Pro/C2C$^\ast$ & 35.9 & 645 & 12.1 & 424 & $-24.6$ & 1572 & 23.3 [14.5, 32.2] & 0.252 & 4.5 [$-2.9$, $+12.1$]\\\bottomrule
\end{tabular*}
\end{table}

\paragraph{Where the replayed saving comes from.} Splitting each replayed policy's paired mean saving by source, using the scores and routes logged by each policy request, skipped questions with $u>0$ supply 7.3 to 41.3\% of the saving on skipped questions among the eight Qwen-receiver policies whose interval excludes zero (33.4\% for large/MMLU-Pro/C2C). The Llama-8B receiver, which has no $u=0$ question, draws its whole saving from them (Table~\ref{tab:u0saving}; Figure~\ref{fig:zero_u}b). A recomposed rule that skips only the $u=0$ questions, charging each other skipped question its fixed latency plus its logged probe time, would keep 57 to 76\% of the Text savings, 75\% of the large/ARC/C2C saving, 18\% of the large/OBQA/C2C saving, and 85 to 90\% of the medium-pair savings. It would not save for large/MMLU-Pro/C2C ($-9.8$~ms).

\begin{table}[!htbp]\centering\scriptsize
\caption{At $u=0$, option rotation changes the medium and large receivers' own answers 1.8 to 6.3\% of the time, against 19.5 to 44.6\% at $u>0$ (receiver alone, development) [Clopper--Pearson 95\% CI]. Option rotation: each option's text moved one position down, labels kept (answers mapped back). Change rate: Changed$/n$, not a policy's change rate. Both answers parse: changed answers out of the questions on which both the original and the rotated answer parse; --: no question.}
\label{tab:rot_dev}
\renewcommand{\arraystretch}{1.1}\setlength{\tabcolsep}{2pt}
\begin{tabular*}{\linewidth}{@{}l@{\hspace{6pt}}l@{\hspace{4pt}\extracolsep{\fill}}rclrclc@{}}\toprule
& & \multicolumn{3}{c}{Development questions with $u=0$} & \multicolumn{3}{c}{Development questions with $u>0$} & \\\cmidrule(lr){3-5}\cmidrule(lr){6-8}
Receiver & Task & \multicolumn{1}{c}{$n$} & Changed & \multicolumn{1}{c}{\begin{tabular}[b]{@{}c@{}}Change rate (\%)\\{}[95\% CI]\end{tabular}} & \multicolumn{1}{c}{$n$} & Changed & \multicolumn{1}{c}{\begin{tabular}[b]{@{}c@{}}Change rate (\%)\\{}[95\% CI]\end{tabular}} & \multicolumn{1}{c@{}}{\begin{tabular}[b]{@{}c@{}}Both answers parse:\\changed / questions\end{tabular}}\\\midrule
small & OBQA & 0 & -- & \multicolumn{1}{c}{--} & 742 & 558 & 75.2 [71.9, 78.3] & 521/696\\
small & ARC & 0 & -- & \multicolumn{1}{c}{--} & 299 & 209 & 69.9 [64.4, 75.0] & 180/258\\
medium & OBQA & 321 & 19 & 5.9 [3.6, 9.1] & 421 & 164 & 39.0 [34.3, 43.8] & 176/732\\
medium & ARC & 174 & 11 & 6.3 [3.2, 11.0] & 125 & 42 & 33.6 [25.4, 42.6] & 51/297\\
large & OBQA & 354 & 9 & 2.5 [1.2, 4.8] & 388 & 128 & 33.0 [28.3, 37.9] & 134/739\\
large & ARC & 217 & 4 & 1.8 [0.5, 4.7] & 82 & 16 & 19.5 [11.6, 29.7] & 20/299\\
large & MMLU-Pro & 645 & 23 & 3.6 [2.3, 5.3] & 1996 & 891 & 44.6 [42.4, 46.9] & 906/2631\\\bottomrule
\end{tabular*}
\end{table}

\paragraph{Option order.} Post hoc, rotating the options as in the null control (answers mapped back) changes the medium and large receivers' own $u=0$ development answers 1.8 to 6.3\% of the time, against 19.5 to 44.6\% at $u>0$. The upper bound at $u=0$ exceeds 5\% for medium OBQA (9.1\%), medium ARC (11.0\%), and large MMLU-Pro (5.3\%), so part of the receiver's confident answers follow option position (Table~\ref{tab:rot_dev}). These are changes of the receiver's own answer, not a policy's change rate. On each policy's skipped development questions, the receiver's answer changes under rotation 3.3 to 10.0\% of the time. In contrast, the five rotated large-pair policies change 1.7 to 2.6\% of skipped answers relative to their rotated references (Table~\ref{tab:rotation}). The rotated receiver-only runs stored no probe score.

\paragraph{Precision and ties.} At an exact zero, every non-top label has probability below $2^{-24}$ and all of them together below $2^{-23}$ (at most $1.08\times10^{-7}$). Recomputing the probe in FP32 (where 48 and 72\% of the large receiver's OBQA and ARC questions, and 45 and 57\% of the medium receiver's, stay exactly zero, and development coverage moves by at most 0.94 points), or breaking ties by the log of the non-top mass, changes no certification decision in 23 settings. By contrast, the top-two logit gap, itself tied on 98 to 100\% of questions, moves medium/OBQA/C2C to $q=0.50$. No setting switches between deployment and fallback under either score.

\paragraph{Unparseable answers.} A question counts as a disagreement when exactly one of the two answers fails to parse. These make up 0.000 to 0.707 of disagreement across the 18 main settings (pooled 0.274 on calibration and 0.264 on development) and 0.187 to 0.495 across the seven OLMo and Llama settings. Post hoc, in separate 20-test families with unchanged fit thresholds, a non-deployable filter keeping only questions where both answers parse leaves all 11 certified multiple-choice policies at their quantile. It also certifies the three OLMo settings and medium/MMLU-Pro/C2C ($q=0.15$). A deployable rule that routes to the reference whenever the receiver's own answer fails to parse changes no outcome. Every development change of the large-pair OBQA and ARC, Text+fact, and Llama-8B policies is between parsed answers, as is every change on the sealed test, for Text+fact held out, and in the two Llama-8B tests. Exactly one \textsc{invalid} answer is involved in 2 of 40 and 3 of 37 development changes for the two MMLU-Pro policies and 2 of 11 for medium/OBQA/C2C, and out of sample in 1 of 12 and 2 of 13 held-out changes for large/OBQA/Text and C2C and 2 of 8 for medium/OBQA/C2C. For medium/ARC/C2C, 2 of 4 development and 7 of 15 out-of-sample changes involve a C2C answer that does not parse.

%% file: new_app_J.tex
\section{Predicting the boundary}\label{app:predict}

Given its calibration size, each setting's disagreement rate and AUROC predict its certification outcome, allowing a small shadow sample to anticipate it. However, the bound still comes from the calibration test. The rules of every analysis below were fixed before computing.

\paragraph{A small-sample check.} Before the four small- and medium-pair MMLU-Pro settings and GSM8K had calibration results, a manuscript draft already stated the pattern among the 22 settings then run: deployed settings had development AUROC of at least 0.82 and fallbacks at most 0.79. All five later settings have AUROC below 0.79 and fall back (the timestamps of their result files are in the code repository). Because the cut near 0.80 was read off the resulting 26 multiple-choice settings, a protocol fixed before computing drew 200 samples of $n$ fit questions per setting. It predicted deployment when the sample AUROC is at least 0.80. This prediction matches the outcome in 0.880, 0.911, and 0.924 of draws on average for $n=100$, 200, and 400, and at $n=200$ in at least 0.80 of draws for 21 of 26 settings (0.950 over deployed and 0.890 over fallback settings). Misses concentrate near the cut and where the fit and development splits differ: medium/ARC/Text matches in 14.5\% of draws (fit-sample AUROC median 0.829, development 0.742), while medium/OBQA/Text, large/MMLU-Pro/Text, and OLMo/OBQA match in 53.0, 66.5, and 78.5\%. This cut, fixed in that protocol, sorts four of the five settings that ran afterwards (the two stronger-helper settings, the two Llama-8B settings, and SQuAD). The exception is the stronger helper's ARC setting (AUROC 0.800), a fallback that the plug-in rule and the binormal model below, both fixed before it ran, predicted.

\paragraph{Predicting certification from the fit split.} A label-free plug-in rule applies the frozen test to fit-split counts rescaled to the calibration size. For candidate $j$, with $m_j$ fit questions at or below its threshold and $d_j$ of them disagreeing, it tests $\hat n_j=\mathrm{round}(N_{\rm cal}m_j/N_{\rm fit})$ and $\hat k_j=\mathrm{round}(\hat n_jd_j/m_j)$ and predicts the largest accepted $q$, with coverage $m_j/N_{\rm fit}$. The rule matches all 8 deployments and 17 fallbacks of the 25 settings with fit-split reference outputs ($q$ exact in 5 of 8, coverage off by a median of 2.7 points; Table~\ref{tab:fitpred}). This is a plug-in estimate of the test's outcome, not an independent check. The predicted saving (the predicted coverage times the gap between the mean development latencies of the reference and of the receiver alone, minus the mean probe cost) is off by a median of 15.4~ms (12.6\%), and by 15.3~ms with each panel's actual coverage. Thus, the error comes mainly from the latency model. The largest relative miss is large/MMLU-Pro/C2C, whose measured saving has an interval that includes zero. Run later, byte-identical, the rule predicted both stronger-helper fallbacks prospectively (2 of 2) and both Llama-8B deployments. The Llama-8B predictions are not prospective because those settings had begun producing outputs before the rule was fixed ($q=0.65$ against 0.60 and 0.70, coverage off by 6.0 and 3.8 points).

\begin{table}[!htbp]\centering\scriptsize
\caption{The fit split alone predicts every certification outcome of the 25 settings, with the deployed $q$ exact in five of the eight deployments. Predicted: the frozen test applied to fit-split counts rescaled to the calibration size. Coverage: development (certified) and fit (predicted); saving predicted with the predicted or the panel's coverage. Measured: saving of the end-to-end replay on the 128-question panel; predicted saving: coverage times the gap between the reference's and $R$'s mean development latencies, minus the mean probe cost. OLMo: large helper, OLMo-2-1124-7B-Instruct receiver; Llama: Llama-3.2-1B-Instruct helper, small receiver.}
\label{tab:fitpred}
\renewcommand{\arraystretch}{1.1}\setlength{\tabcolsep}{3pt}
\begin{tabular*}{\linewidth}{@{}l@{\hspace{5pt}\extracolsep{\fill}}ccccccc@{}}\toprule
& \multicolumn{2}{c}{Deployed $q$} & \multicolumn{2}{c}{Coverage (\%)} & \multicolumn{3}{c}{Saving (ms)}\\\cmidrule(lr){2-3}\cmidrule(lr){4-5}\cmidrule(l){6-8}
Setting & certified & predicted & \begin{tabular}[b]{@{}c@{}}development\\(certified)\end{tabular} & \begin{tabular}[b]{@{}c@{}}fit\\(predicted)\end{tabular} & measured & \begin{tabular}[b]{@{}c@{}}predicted, with\\predicted coverage\end{tabular} & \begin{tabular}[b]{@{}c@{}}predicted, with\\panel coverage\end{tabular}\\\midrule
medium/OBQA/C2C & 0.55 & 0.55 & 52.6 & 55.1 & 79.7 & 64.7 & 58.4\\
medium/ARC/C2C & 0.60 & 0.60 & 63.2 & 61.0 & 100.7 & 98.9 & 97.0\\
large/OBQA/Text & 0.80 & 0.80 & 77.1 & 80.0 & 517.1 & 536.6 & 506.1\\
large/OBQA/C2C & 0.80 & 0.85 & 77.1 & 85.0 & 36.6 & 26.6 & 19.0\\
large/ARC/Text & 0.95 & 0.90 & 95.0 & 90.0 & 717.6 & 670.6 & 700.4\\
large/ARC/C2C & 0.90 & 0.90 & 90.3 & 90.0 & 110.9 & 88.9 & 88.6\\
large/MMLU-Pro/Text & 0.40 & 0.40 & 40.5 & 40.0 & 346.3 & 360.9 & 335.4\\
large/MMLU-Pro/C2C & 0.40 & 0.35 & 40.5 & 35.2 & 8.2 & $-7.7$ & $-5.1$\\\midrule
\multicolumn{8}{@{}p{\linewidth}@{}}{\raggedright Certified fallback and predicted fallback (17 settings): small/OBQA/Text, small/OBQA/C2C, small/ARC/Text, small/ARC/C2C, medium/OBQA/Text, medium/ARC/Text, small/MMLU-Pro/Text, small/MMLU-Pro/C2C, medium/MMLU-Pro/Text, medium/MMLU-Pro/C2C, OLMo/OBQA/Text, OLMo/ARC/Text, OLMo/MMLU-Pro/Text, Llama/OBQA/Text, Llama/OBQA/C2C, Llama/ARC/Text, Llama/ARC/C2C.}\\\bottomrule
\end{tabular*}
\end{table}

\begin{figure}[!t]\centering
\includegraphics[width=\linewidth]{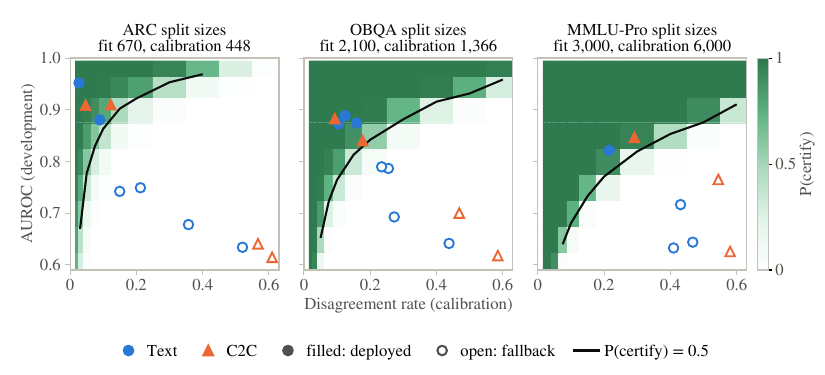}
\caption{The frozen test certifies a binormal score only above an AUROC that rises with the disagreement rate and falls with the calibration size, which reproduces all 28 certification outcomes. Binormal score: agreeing questions score $\mathcal{N}(0,1)$, disagreeing ones $\mathcal{N}(\mu,1)$, $\mu=\sqrt2\,\Phi^{-1}(\mathrm{AUROC})$. Probability that the frozen test certifies (1,000 simulations per cell) at the fit and calibration sizes of ARC, OBQA, and MMLU-Pro; black line: probability 0.5, not drawn where it falls below AUROC 0.60 or above 0.98. Points: the 28 settings (18 main, 9 cross-family, and Text+fact) at their calibration disagreement rate and development AUROC; filled if deployed; Text+fact is drawn as Text. Llama: Llama-3.2-1B-Instruct helper, small receiver. The Llama pair's ARC/C2C setting (disagreement 0.61) sits on the right edge of the first panel.}
\label{fig:binormal}
\end{figure}

\paragraph{A binormal model of the boundary.} We simulated the frozen procedure with binormal scores: agreeing questions draw from $\mathcal{N}(0,1)$ and disagreeing ones from $\mathcal{N}(\mu,1)$, $\mu=\sqrt2\,\Phi^{-1}(\mathrm{AUROC})$, at each setting's calibration disagreement rate and split sizes. Each of 2,000 simulations draws a fit split, sets the 20 thresholds, draws a calibration split, and applies the frozen test and selection rule. Predicting deployment when the simulated probability is at least 0.5 reproduces all 28 outcomes, using either the development or the calibration AUROC. Certified settings have probabilities from 0.61 (medium/OBQA/C2C) to 1, and no fallback exceeds 0.05. Run later, the same code predicted both stronger-helper fallbacks (probability 0.040 and 0.049; AUROC needed 0.853 and 0.885). Therefore, from each setting's disagreement rate and AUROC, the model accounts for all 30 outcomes that preceded SQuAD: the boundary is what the test implies given these two numbers, and the empirical content is where each path falls. From the pilot, it predicted the SQuAD certification (0.975, inflated by selecting the best of eight; conservative 0.081), a prospective test, and 0.749 from SQuAD's development statistics. The model has no point mass at zero uncertainty yet reproduces every outcome. The AUROC needed for probability 0.5 ranges from 0.63 (large/ARC/Text) to 0.996 (the Llama pair's ARC/C2C; Figure~\ref{fig:binormal}). The population condition of \citet{uddin2026proactive}, a candidate whose true-to-false-positive ratio on calibration reaches $(1-\pi)(1-\alpha)/(\pi\alpha)$, holds in all 11 certified multiple-choice settings, in SQuAD (18.0 against 3.38), and in eight fallback settings, including both stronger-helper ones, whose calibration splits are too small for the finite-sample test.

\paragraph{A helper-aware score.} Under a hypothesis fixed before any output, we added a helper-agreement term to ProbeMax: $s_G(x)=u_R(x)+\mathbb{I}[\arg\max_H(x)\neq\arg\max_R(x)]$. Here, $\arg\max_H$ comes from one helper prefill of the question and options with the same answer phrase and label-token rule. The extracted prompt construction reproduced the frozen receiver probes bit for bit, and setting $\arg\max_H:=\arg\max_R$ reproduced the deployed $q$ of all 14 OBQA, ARC, and large-pair MMLU-Pro settings. The term raises development AUROC by 0.021 on average for Text and lowers it by 0.049 for C2C, favoring Text in all seven populations (at least five were required). The four later small- and medium-pair MMLU-Pro settings point the same way (Text larger in all nine populations, $+0.019$ against $-0.043$), which extends the prespecified rule rather than replacing it. No setting crosses the boundary in either direction. Where the helper and receiver often differ, the term is counterproductive: on MMLU-Pro, it pushes a third of the small pair's and about 60\% of the medium pair's questions above every threshold. The helper prefill costs 17.4~ms (small pair), 20.2 to 20.8~ms (medium), and 25.9 to 35.5~ms (large), and it runs the helper that skipping avoids. This analysis covers the 18 main settings and Text+fact, is post hoc, and needed one key-naming fix to the script, which then reproduced every earlier number and the verdict.

%% file: new_app_F.tex
\section{Accuracy under skipping}\label{app:accuracy}

The change-rate certificate bounds accuracy only through $\kappa\alpha$, which can exceed a helpful path's gain. However, gold labels on a few skipped disagreements bound the loss directly: skipping can lower accuracy only on skipped questions where the reference is correct and the receiver alone is wrong. At $\alpha=0.05$, the three large-pair Text policies retain 33 to 105\% of the reference's gain. Stricter tolerances ($\alpha\le0.03$), which require labels to set, retain more gain on OBQA, ARC, and Text+fact, though not on MMLU-Pro.

\begin{figure}[!t]\centering
\includegraphics[width=\linewidth]{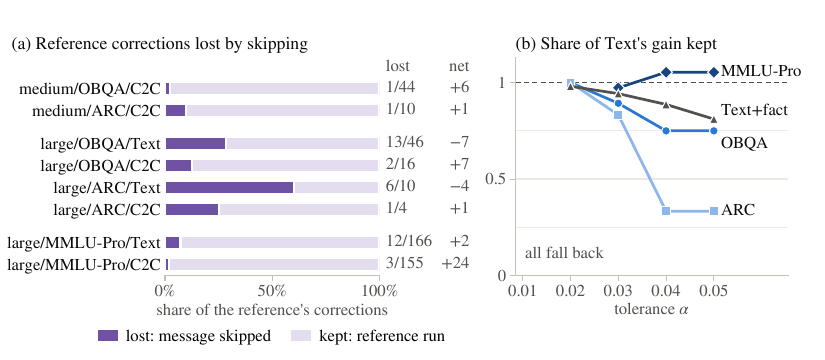}
\caption{Skipping loses few reference corrections under C2C but more under Text, and a stricter tolerance keeps more of Text's gain except on MMLU-Pro. (a)~Reference corrections (development questions the reference answers correctly and the receiver alone does not) that each deployed Qwen-pair policy loses by skipping or keeps by running the reference, with the lost count out of all the reference's corrections and the net change in correct answers (policy minus reference, gained minus lost). (b)~Share of the reference's gain over the receiver that the large-pair Text policies keep at each tolerance $\alpha$ (Table~\ref{tab:retention}); missing points fall back, and the dashed line marks the whole gain. Development populations; descriptive and gold-dependent; large/OBQA/Text, large/OBQA/C2C, and Text+fact are nominal (Appendix~\ref{app:history}).}
\label{fig:corrections}
\end{figure}

\paragraph{Gain retention.} Under rules fixed before computing, we measured the reference's gain over the receiver ($G=\mathrm{acc}(b)-\mathrm{acc}(R)$), the policy's gain ($\mathrm{acc}(\pi_\tau)-\mathrm{acc}(R)$), and the share of $G$ retained, using paired bootstrap intervals. We repeated the certification with $\alpha$ from 0.01 to 0.05 at the same per-candidate cutoff (Table~\ref{tab:retention}; Figure~\ref{fig:corrections}b). At $\alpha=0.05$, policies retain 0.750 [0.467, 1.067] of Text's gain on OBQA, 1.054 [0.717, 1.858] on MMLU-Pro, and 0.333 [$-1.0$, 1.4] on ARC, where the gain itself is uncertain ($G=2.01$ points [$-0.33$, $+4.68$]; $G\le0$ in 6.8\% of resamples), and Text+fact retains 0.811 [0.686, 0.927]. On the 744 held-out OBQA questions, where $R$ ran on every question, large Text retains 0.759 [0.500, 0.958] and Text+fact retains 0.817 [0.706, 0.933]; the sealed ARC run did not execute $R$ on routed questions. The largest $\alpha$ with $\kappa\alpha<G$ is 0.04 for OBQA/Text (threshold deployed at 0.05), 0.02 for ARC/Text, 0.03 for MMLU-Pro/Text, and 0.05 for Text+fact; any such $\alpha$ must be chosen with labels. For these thresholds, we recomposed the panel saving from the per-request timings of each existing replay, excluding cold requests: a question the stricter policy still skips retains its recorded policy latency, and a routed question is charged its recorded probe latency plus the fixed arm's latency on that question; at the original thresholds this reproduces the measured saving within 1~ms. The recomposed savings are 532.9, 587.3, and 247.0~ms on the original configuration and 420.3 to 429.5, 469.3 to 479.5, and 197.1 to 199.8~ms in the three repeats on the second one (OBQA, ARC, and MMLU-Pro), and 342.7~ms for Text+fact on the second configuration; these are not new replays.

\paragraph{Lost reference corrections.} On skipped questions a policy replaces the reference's answer by the receiver's own, so for the skip set $O_\tau=\{x:s(x)\le\tau\}$ and gold label $y$, $\mathrm{acc}(\pi_\tau)-\mathrm{acc}(b)=\Pr(x\in O_\tau,\,o_R=y,\,o_b\neq y)-\Pr(x\in O_\tau,\,o_R\neq y,\,o_b=y)$: the receiver's successes that skipping recovers minus the reference corrections (questions the reference answers correctly and $R$ does not) that it loses. The identity holds in all fourteen OBQA, ARC, and large-pair MMLU-Pro settings. No C2C policy loses more than 3 reference corrections, whereas large/OBQA/Text loses 13 of 46 and large/ARC/Text 6 of 10 (Figure~\ref{fig:corrections}a); net of the successes they recover, the C2C policies answer 1 to 24 more questions correctly than their references, large/MMLU-Pro/Text answers 2 more, and large/OBQA/Text and large/ARC/Text answer 7 and 4 fewer. These are descriptive development outcomes that the risk test does not target. Questions where some action is correct but the policy is not (unused three-action opportunity) are more common on the kept side in seven of the eight settings; this does not show that the probe identifies them or that a three-action router could realize the oracle.

\begin{table}[!htbp]\centering\scriptsize
\caption{Lowering $\alpha$ never raises coverage; at $\alpha\le0.03$ it keeps more of the reference's gain over the receiver on OBQA, ARC, and Text+fact, but not on MMLU-Pro (development sets). $q$: deployed quantile (fallback: no threshold passes; --: not applicable). Changed: changed/skipped development questions. $\kappa\alpha$: worst-case accuracy loss (points). Retained: share of $G$ kept; $G$ is 3.77 (OBQA), 2.01 (ARC), 1.40 (MMLU-Pro), and 7.14 points (Text+fact).}
\label{tab:retention}
\renewcommand{\arraystretch}{1.1}\setlength{\tabcolsep}{3pt}
\begin{tabular}{lcccccc@{\hspace{10pt}}cccccc}\toprule
& \multicolumn{5}{c}{large/OBQA/Text} & & \multicolumn{5}{c}{large/ARC/Text}\\\cmidrule(lr){2-6}\cmidrule(lr){8-12}
\multicolumn{1}{c}{$\alpha$} & $q$ & Coverage (\%) & Changed & $\kappa\alpha$ & Retained & & $q$ & Coverage (\%) & Changed & $\kappa\alpha$ & Retained\\\midrule
0.01 & fallback & 0 & -- & -- & -- & & fallback & 0 & -- & -- & --\\
0.02 & 0.65 & 62.0 & 6/460 & 1.24 & 1.000 & & 0.80 & 77.9 & 0/233 & 1.56 & 1.000\\
0.03 & 0.75 & 72.2 & 15/536 & 2.17 & 0.893 & & 0.85 & 84.9 & 1/254 & 2.55 & 0.833\\
0.04 & 0.80 & 77.1 & 19/572 & 3.08 & 0.750 & & 0.95 & 95.0 & 9/284 & 3.80 & 0.333\\
0.05 & 0.80 & 77.1 & 19/572 & 3.85 & 0.750 & & 0.95 & 95.0 & 9/284 & 4.75 & 0.333\\\midrule
& \multicolumn{5}{c}{large/MMLU-Pro/Text} & & \multicolumn{5}{c}{large/OBQA/Text+fact}\\\cmidrule(lr){2-6}\cmidrule(lr){8-12}
0.01 & fallback & 0 & -- & -- & -- & & fallback & 0 & -- & -- & --\\
0.02 & fallback & 0 & -- & -- & -- & & 0.55 & 53.6 & 1/398 & 1.07 & 0.981\\
0.03 & 0.30 & 30.7 & 16/812 & 0.92 & 0.973 & & 0.65 & 62.0 & 3/460 & 1.86 & 0.943\\
0.04 & 0.35 & 36.0 & 31/952 & 1.44 & 1.054 & & 0.70 & 66.4 & 7/493 & 2.66 & 0.887\\
0.05 & 0.40 & 40.5 & 40/1069 & 2.02 & 1.054 & & 0.75 & 72.2 & 13/536 & 3.61 & 0.811\\\bottomrule
\end{tabular}
\end{table}

\begin{table}[!htbp]\centering\scriptsize
\caption{The lost-correction test needs gold labels only on the calibration disagreements up to its first rejected candidate or the deployed threshold, whichever is lower ($D_{\rm need}$), and at $\varepsilon=0.01$ it is infeasible on the two ARC settings. Lost-correction test: exact binomial test of $H_0$: rate of lost corrections (skipped questions the reference gets right and $R$ wrong) $\ge\varepsilon$, the certified accuracy-loss bound. Accepted: candidates (of 20) accepted by the lost-correction test, the change-rate test, and both. $q_L$: largest candidate the lost-correction test accepts; $q_{\rm joint}$: largest candidate both accept (deployed by the joint certificate); $N$: calibration questions; infeasible: $N$ below the 688 that $\varepsilon=0.01$ needs. Disagreements: all ($D_{\rm all}$), up to the first rejected candidate ($D_{\rm seq}$), at $q_{\rm joint}$ ($D_{\rm dep}$), and up to that candidate or the deployed threshold, whichever is lower ($D_{\rm need}$). Without reviewed: $q_{\rm joint}$ without the parser-review questions (1,158 OBQA and 315 ARC remain; infeasible: 315 is below the 342 that $\varepsilon=0.02$ needs); parser-review questions: 341 calibration questions whose outputs were read with gold labels before the parser was frozen. --: not applicable. $^\ddagger$Nominal certificate.}
\label{tab:lostcal}
\renewcommand{\arraystretch}{1.1}\setlength{\tabcolsep}{2.5pt}
\vspace{5pt}\textit{(a) Accepted candidates and certified thresholds}\par\smallskip
\begin{tabular*}{\linewidth}{@{}l@{\hspace{5pt}\extracolsep{\fill}}ccccccccc@{}}\toprule
& & & \multicolumn{3}{c}{Candidates accepted (of 20)} & \multicolumn{3}{c}{Largest accepted $q$} & \\\cmidrule(lr){4-6}\cmidrule(lr){7-9}
Setting & $\varepsilon$ & $N$ & \begin{tabular}[b]{@{}c@{}}lost-\\correction\end{tabular} & \begin{tabular}[b]{@{}c@{}}change-\\rate\end{tabular} & both & \begin{tabular}[b]{@{}c@{}}lost-correction\\($q_L$)\end{tabular} & \begin{tabular}[b]{@{}c@{}}both\\($q_{\rm joint}$)\end{tabular} & \begin{tabular}[b]{@{}c@{}}change-rate\\(deployed)\end{tabular} & \begin{tabular}[b]{@{}c@{}}$q_{\rm joint}$ without\\reviewed\end{tabular}\\\midrule
large/OBQA/Text$^\ddagger$ & 0.01 & 1,366 & 13 & 16 & 13 & 0.65 & 0.65 & 0.80 & 0.60\\
large/OBQA/Text$^\ddagger$ & 0.02 & 1,366 & 15 & 16 & 15 & 0.75 & 0.75 & 0.80 & 0.75\\
\addlinespace[2pt]
large/ARC/Text & 0.01 & 448 & 0 & 19 & 0 & infeasible & infeasible & 0.95 & infeasible\\
large/ARC/Text & 0.02 & 448 & 16 & 19 & 16 & 0.80 & 0.80 & 0.95 & infeasible\\
\addlinespace[2pt]
large/MMLU-Pro/Text & 0.01 & 6,000 & 9 & 8 & 8 & 0.45 & 0.40 & 0.40 & 0.40\\
large/MMLU-Pro/Text & 0.02 & 6,000 & 12 & 8 & 8 & 0.60 & 0.40 & 0.40 & 0.40\\
\addlinespace[2pt]
large/OBQA/Text+fact$^\ddagger$ & 0.01 & 1,366 & 11 & 15 & 11 & 0.55 & 0.55 & 0.75 & 0.55\\
large/OBQA/Text+fact$^\ddagger$ & 0.02 & 1,366 & 14 & 15 & 14 & 0.70 & 0.70 & 0.75 & 0.65\\
\addlinespace[2pt]
Llama-8B/OBQA/Text & 0.01 & 1,366 & 9 & 10 & 7 & 0.45 & 0.45 & 0.60 & 0.35\\
Llama-8B/OBQA/Text & 0.02 & 1,366 & 12 & 10 & 10 & 0.60 & 0.60 & 0.60 & 0.50\\
\addlinespace[2pt]
Llama-8B/ARC/Text & 0.01 & 448 & 0 & 8 & 0 & infeasible & infeasible & 0.70 & infeasible\\
Llama-8B/ARC/Text & 0.02 & 448 & 9 & 8 & 3 & 0.45 & 0.45 & 0.70 & infeasible\\\bottomrule
\end{tabular*}
\par\vspace{17pt}
\textit{(b) Calibration disagreements that need gold labels}\par\smallskip
\begin{tabular*}{\linewidth}{@{}l@{\hspace{5pt}\extracolsep{\fill}}ccccc@{}}\toprule
& & \multicolumn{4}{c}{Calibration disagreements}\\\cmidrule(l){3-6}
Setting & $\varepsilon$ & all ($D_{\rm all}$) & up to the first rejection ($D_{\rm seq}$) & at $q_{\rm joint}$ ($D_{\rm dep}$) & labels needed ($D_{\rm need}$)\\\midrule
large/OBQA/Text$^\ddagger$ & 0.01 & 143 & 8 & 4 & 8\\
large/OBQA/Text$^\ddagger$ & 0.02 & 143 & 18 & 12 & 18\\
\addlinespace[2pt]
large/ARC/Text & 0.01 & 12 & 0 & -- & 0\\
large/ARC/Text & 0.02 & 12 & 1 & 0 & 1\\
\addlinespace[2pt]
large/MMLU-Pro/Text & 0.01 & 1,295 & 153 & 67 & 67\\
large/MMLU-Pro/Text & 0.02 & 1,295 & 361 & 67 & 67\\
\addlinespace[2pt]
large/OBQA/Text+fact$^\ddagger$ & 0.01 & 170 & 7 & 3 & 7\\
large/OBQA/Text+fact$^\ddagger$ & 0.02 & 170 & 22 & 16 & 22\\
\addlinespace[2pt]
Llama-8B/OBQA/Text & 0.01 & 217 & 12 & 7 & 12\\
Llama-8B/OBQA/Text & 0.02 & 217 & 32 & 21 & 21\\
\addlinespace[2pt]
Llama-8B/ARC/Text & 0.01 & 40 & 0 & -- & 0\\
Llama-8B/ARC/Text & 0.02 & 40 & 1 & 0 & 1\\\bottomrule
\end{tabular*}
\end{table}

\begin{table}[!htbp]\centering\scriptsize
\caption{At $\varepsilon=0.01$ the three counted settings keep 100, 98.1, and 106.1\% of the gain on development and lose at most two corrections on the held-out questions. $\varepsilon$: certified accuracy-loss bound; counted: coverage at least 20\% and $\varepsilon$ at most half of $G$. Values at $q_{\rm joint}$, the largest candidate that the lost-correction and change-rate tests both accept (Table~\ref{tab:lostcal}); infeasible: too few calibration questions (448; 688 needed); --: no joint certificate. Lost corrections: skipped questions that the reference answers correctly and the receiver alone does not; gained answers: the reverse. $\Delta$acc: policy minus reference accuracy (points) with its paired-bootstrap 95\% CI. $G$: reference minus receiver (points); Kept: share of $G$. (a) Saving: panel saving over all requests with its paired-bootstrap 95\% CI, recomposed from the per-request timings of the existing replays with the same configurations (a newly routed question costs its recorded probe latency plus the reference's); in row order, the deployed policies save 517.1, 717.6, and 346.3~ms (original configuration) and 343.4, 318.6, and 418.9~ms (second configuration). (b) Lost corrections over all held-out or sealed questions with a Clopper--Pearson 95\% CI for their rate, and coverage, gained answers, and $\Delta$acc on the same questions; MMLU-Pro has no unused questions. $^\ddagger$Nominal certificate.}
\label{tab:lostcert}
\renewcommand{\arraystretch}{1.15}\setlength{\tabcolsep}{2pt}
\vspace{5pt}\textit{(a) Development}\par\smallskip
\begin{tabular*}{\linewidth}{@{}l@{\hspace{4pt}\extracolsep{\fill}}ccccccccccc@{}}\toprule
& & & Coverage & Lost & Gained & \multicolumn{2}{c}{$\Delta$acc (points)} & $G$ & Kept & \multicolumn{2}{c}{Saving (ms)}\\\cmidrule(lr){7-8}\cmidrule(l){11-12}
Setting & $\varepsilon$ & $q_{\rm joint}$ & (\%) & corrections & answers & & 95\% CI & (points) & (of $G$) & & 95\% CI\\\midrule
large/OBQA/Text$^\ddagger$ & 0.01 & 0.65 & 62.0 & 3 & 3 & $+0.00$ & [$-0.67$, $+0.67$] & 3.77 & 1.000 & 393.4 & [310.3, 468.8]\\
large/OBQA/Text$^\ddagger$ & 0.02 & 0.75 & 72.2 & 9 & 6 & $-0.40$ & [$-1.48$, $+0.54$] & 3.77 & 0.893 & 490.0 & [411.9, 560.5]\\
\addlinespace[2pt]
large/ARC/Text & 0.01 & infeasible & -- & -- & -- & -- & -- & 2.01 & -- & -- & --\\
large/ARC/Text & 0.02 & 0.80 & 77.9 & 0 & 0 & $+0.00$ & [0.00, 0.00] & 2.01 & 1.000 & 588.6 & [522.4, 656.1]\\
\addlinespace[2pt]
large/MMLU-Pro/Text & 0.01 & 0.40 & 40.5 & 12 & 14 & $+0.08$ & [$-0.30$, $+0.45$] & 1.40 & 1.054 & 346.3 & [249.8, 443.0]\\
large/MMLU-Pro/Text & 0.02 & 0.40 & 40.5 & 12 & 14 & $+0.08$ & [$-0.30$, $+0.45$] & 1.40 & 1.054 & 346.3 & [249.8, 443.0]\\
\addlinespace[2pt]
large/OBQA/Text+fact$^\ddagger$ & 0.01 & 0.55 & 53.6 & 1 & 0 & $-0.13$ & [$-0.40$, 0.00] & 7.14 & 0.981 & 245.1 & [195.7, 293.1]\\
large/OBQA/Text+fact$^\ddagger$ & 0.02 & 0.70 & 66.4 & 6 & 0 & $-0.81$ & [$-1.48$, $-0.27$] & 7.14 & 0.887 & 306.2 & [259.2, 352.0]\\
\addlinespace[2pt]
Llama-8B/OBQA/Text & 0.01 & 0.45 & 45.0 & 1 & 3 & $+0.27$ & [$-0.27$, $+0.81$] & 4.45 & 1.061 & 218.1 & [164.1, 273.4]\\
Llama-8B/OBQA/Text & 0.02 & 0.60 & 59.0 & 2 & 4 & $+0.27$ & [$-0.40$, $+0.94$] & 4.45 & 1.061 & 318.6 & [261.1, 372.1]\\
\addlinespace[2pt]
Llama-8B/ARC/Text & 0.01 & infeasible & -- & -- & -- & -- & -- & 7.02 & -- & -- & --\\
Llama-8B/ARC/Text & 0.02 & 0.45 & 40.8 & 0 & 0 & $+0.00$ & [0.00, 0.00] & 7.02 & 1.000 & 199.0 & [144.6, 254.9]\\\bottomrule
\end{tabular*}
\par\vspace{17pt}
\textit{(b) Out of sample}\par\smallskip
\begin{tabular*}{\linewidth}{@{}l@{\hspace{4pt}\extracolsep{\fill}}cccccccc@{}}\toprule
& & & \multicolumn{2}{c}{Lost corrections} & Coverage & Gained & \multicolumn{2}{c}{$\Delta$acc (points)}\\\cmidrule(lr){4-5}\cmidrule(l){8-9}
Setting & $\varepsilon$ & Test questions & count & 95\% CI (\%) & (\%) & answers & & 95\% CI\\\midrule
large/OBQA/Text$^\ddagger$ & 0.01 & held out (744) & 2 & [0.03, 0.97] & 65.2 & 1 & $-0.13$ & [$-0.54$, $+0.27$]\\
large/OBQA/Text$^\ddagger$ & 0.02 & held out (744) & 4 & [0.15, 1.37] & 75.0 & 2 & $-0.27$ & [$-0.94$, $+0.40$]\\
\addlinespace[2pt]
large/ARC/Text & 0.02 & sealed (1,172) & 0 & [0.00, 0.31] & 79.0 & 0 & $+0.00$ & [0.00, 0.00]\\
\addlinespace[2pt]
large/OBQA/Text+fact$^\ddagger$ & 0.01 & held out (744) & 2 & [0.03, 0.97] & 54.3 & 0 & $-0.27$ & [$-0.67$, 0.00]\\
large/OBQA/Text+fact$^\ddagger$ & 0.02 & held out (744) & 8 & [0.47, 2.11] & 70.2 & 2 & $-0.81$ & [$-1.61$, 0.00]\\
\addlinespace[2pt]
Llama-8B/OBQA/Text & 0.01 & held out (744) & 1 & [0.00, 0.75] & 44.0 & 0 & $-0.13$ & [$-0.40$, 0.00]\\
Llama-8B/OBQA/Text & 0.02 & held out (744) & 3 & [0.08, 1.17] & 59.1 & 0 & $-0.40$ & [$-0.94$, 0.00]\\
\addlinespace[2pt]
Llama-8B/ARC/Text & 0.02 & held out (1,172) & 2 & [0.02, 0.62] & 44.3 & 2 & $+0.00$ & [$-0.34$, $+0.34$]\\\bottomrule
\end{tabular*}
\end{table}

\paragraph{Label-light accuracy loss certificate.} Let $k_L$ count the lost corrections (skipped questions where the reference is right and $R$ is wrong) among the $N$ calibration questions. Because $\mathrm{acc}(b)-\mathrm{acc}(\pi_\tau)\le\Pr[L]$, the exact binomial test of $H_0:\Pr[L]\ge\varepsilon$ certifies an accuracy loss of at most $\varepsilon$, and it needs gold labels only on calibration disagreements, taken in order of increasing $u$ up to the first rejected candidate or the deployed threshold. After the main results, under rules fixed before computing, we ran this test for the six helpful Text policies at $\varepsilon=0.01$ and $0.02$ over the same 20 frozen candidates ($\delta=0.001$ each) and intersected it with the change-rate test: the joint certificate is the largest candidate that both accept, with family-wise error of at most 0.04 per setting and $\varepsilon$ (0.06 per setting for both $\varepsilon$). A setting counts if the joint certificate skips at least 20\% of development questions and $\varepsilon$ is at most half the development gain $G$. At $\varepsilon=0.01$, large/OBQA/Text, Text+fact, and Llama-8B/OBQA/Text count (Tables~\ref{tab:lostcal} and~\ref{tab:lostcert}): with labels on 8, 7, and 12 of their 143, 170, and 217 calibration disagreements they skip 62.0, 53.6, and 45.0\% of development questions and keep 100, 98.1, and 106.1\% of the gain. On the 744 held-out questions they lose 2, 2, and 1 corrections (upper bounds 0.97, 0.97, and 0.75\%). Large/MMLU-Pro/Text certifies a loss of at most one point at its deployed threshold, but its gain of 1.40 points is less than twice $\varepsilon$; the two ARC settings have 448 calibration questions, fewer than the 688 that $\varepsilon=0.01$ needs. At $\varepsilon=0.02$, Text+fact's held-out upper bound (2.11\%) exceeds $\varepsilon$, and its development accuracy falls by 0.81 points [0.27, 1.48]. The analysis is post hoc and reuses calibration labels read earlier; two of the three counted settings are nominal (Appendix~\ref{app:history}). Our protocol stated that no calibration lost-correction count was known to us, but an earlier analysis had reported two for Text+fact (18 among the 999 questions skipped at $q=0.75$; 122 over the whole split). Neither was used to fix $\varepsilon$, the 20\% coverage bar, or the gain condition, which were fixed with development lost-correction counts known for some settings. Without the 341 parser-review questions (Appendix~\ref{app:parser}), the joint threshold falls or disappears in 6 of the 10 joint certificates; for the three counted settings it is 0.60, 0.55, and 0.35.

\paragraph{\rev{The medium pair's Text path.}} \rev{On OBQA, the medium pair's Text path helps ($G=4.45$ points on development; 523 against 490 of 742 correct) but fails the change-rate test at every threshold (AUROC 0.787), so no joint certificate can include it. Run alone after the main analysis, with the same 20 frozen candidates, $\delta=0.001$, and the whole calibration split (1,366 questions) as denominator, the lost-correction test accepts no candidate at $\varepsilon=0.01$ or $0.02$. Its first candidate ($q=0.05$) already skips 601 calibration questions, because ties at zero uncertainty merge the lowest quantiles, and loses 21 corrections where at most 3 and 12 would pass ($p=0.978$ and $0.128$). Reaching this decision takes labels on 36 of the 348 calibration disagreements, and the system keeps communicating. A loss-only certificate would also give up the label-free bound on answer changes.}

\paragraph{\rev{Single models.}}\label{app:singlemodel} \rev{Table~\ref{tab:singlemodel} compares the deployed paths with each model answering alone on development questions. The large pair with Text is the most accurate system on both benchmarks: 0.7 (OBQA) and 4.0 points (ARC) above the large helper agent alone, and 3.8 and 2.0 points above the large receiver agent alone. Large-pair C2C falls below its receiver agent alone. The helper agent's accuracy comes from the argmax over option-label probabilities under a fixed answer prefix, not from free generation and parsing, so that row is not scored under the same protocol as the others. The comparison holds the models fixed and does not equalize compute.}

\begin{table}[!htbp]\centering\scriptsize\revcolor
\caption{\rev{Development accuracy of the deployed paths and of each model alone. Correct answers out of 742 (OBQA) and 299 (ARC), with accuracy in \%. $^{a}$Option-label argmax under a fixed answer prefix; the other rows use free generation and our parser.}}
\label{tab:singlemodel}
\begin{tabular*}{\linewidth}{@{}l@{\extracolsep{\fill}}cc@{}}\toprule
System & OBQA & ARC\\\midrule
Qwen3-1.7B alone (medium receiver) & 490 (66.0) & 217 (72.6)\\
medium pair, Text & 523 (70.5) & 221 (73.9)\\
medium pair, C2C & 486 (65.5) & 202 (67.6)\\\midrule
Qwen3-8B alone (large receiver) & 617 (83.2) & 268 (89.6)\\
Qwen2.5-7B-Instruct alone (large helper)$^{a}$ & 640 (86.3) & 262 (87.6)\\
large pair, C2C & 598 (80.6) & 266 (89.0)\\
large pair, Text & 645 (86.9) & 274 (91.6)\\\bottomrule
\end{tabular*}
\end{table}

%% file: app_prospective.tex
\section{Prospective tests}\label{app:prospective}

Two tests apply the label-assisted certificate with every rule fixed before any calibration, development, or unseen-question output of their models existed: a three-agent system on QASC (Appendix~\ref{app:e1}) and a 32B helper and receiver pair on OBQA (Appendix~\ref{app:e2b}).

\subsection{A three-agent system (QASC)}\label{app:e1}
\begingroup\revcolor

This subsection tests the label-assisted certificate of Section~\ref{sec:rq2} prospectively, on a system and a dataset used nowhere else in this work. The protocol (models, prompts, score, candidate thresholds, tests, label rule, policies, and pass rules) was committed and time-stamped before any calibration, development, or sealed-test output existed, and the sealed test was evaluated once. The primary policy meets all five pass rules fixed in advance. On 2,000 sealed questions it skips the verifier agent on 39.9\% of questions, loses 5 corrections (95\% upper bound 0.58\%), and keeps 12.55 of the verifier's 12.80-point gain; on a live development panel it lowers mean latency by 483~ms.

\paragraph{System.} Three agents answer each question (Table~\ref{tab:e1_system}). The reference system $b$ runs all three; the cheap path $s$ omits the verifier agent B. One decision is made per question: after A's message, the probe on C's cheap-path input decides whether C waits for B. C sees only the agents' turns, in the receiver template of Appendix~\ref{app:prompts}: A's turn (helper instruction, then A's message) and, on the reference path, B's turn, whose user content is B's instruction without the two facts, A's note, and the clauses that introduce them, followed by B's message. This is the construction of the Text+fact receiver (Appendix~\ref{app:informed}); the facts reach C only through B's message. Each question's A and B messages are generated once and shared by both paths and the probe. All models run in bfloat16 with SDPA attention, greedy decoding, and batch size one; the parser (version 2) is extended to labels A to H.

\begin{table}[!htbp]\centering\scriptsize\revcolor
\caption{\rev{The three-agent system. A and B: Qwen2.5-7B-Instruct; C: Qwen3-8B with thinking disabled.}}\label{tab:e1_system}
\begin{tabular}{llp{0.42\linewidth}p{0.24\linewidth}}\toprule
Agent & Role & Input & Output (max.\ new tokens)\\\midrule
A & background & question and choices; the helper instruction of Appendix~\ref{app:prompts} & one background message (256)\\
B & verifier & question and choices, the two QASC facts, A's message & checked and extended background (256)\\
C & receiver & A's turn (cheap path), or A's and B's turns (reference) & answer (64)\\\bottomrule
\end{tabular}
\end{table}

\paragraph{Verifier instruction.} B's user message is the first line below, the question block of Appendix~\ref{app:prompts}, the two facts, A's note, and the closing rule. In C's view of B's turn, the two bracketed clauses, the facts, and the note are absent.
{\fontsize{6.3}{7.4}\selectfont
\begin{verbatim}
In at most three clear sentences, state the background knowledge needed to answer the question below.
[ Another assistant has already written a background note for this question: check that note, correct
anything in it that is wrong, and add the knowledge it is missing.] [ Use the two reference facts given
below as your source.]

{question}

Reference facts:
1. {fact1}
2. {fact2}

Background note from another assistant:
{A's message}

Do NOT directly solve or give answer to the question: do not say which choice is correct, do not discuss
or rule out any of the choices, and do NOT mention any option letter.
\end{verbatim}}

\paragraph{Data.} QASC's training and development splits contain 9,060 eight-option questions, each with two supporting facts; the official test split has neither answers nor facts and is not used. Deduplicating by question id and then by normalized question and choices leaves 9,051 questions. A fixed-seed shuffle assigns 2,000 to fit, 3,000 to calibration, 1,000 to development, and 2,000 to the sealed test; the remaining 1,051 were never processed by any model. Prompts use only the question, the choices, and the two facts. QASC is public, so the split is independent of our design process, not of the models' pretraining data.

\paragraph{Score.} Each label's probability sums the next-token probabilities over its token set and is renormalized over A to H. The score is the log of the non-top mass, $u=\log\sum_{l\ne l^\ast}p_l$, computed as a log-sum-exp in FP32; it orders questions as $1-\max_l p_l$ does wherever that is nonzero. On fit, $1-\max_l p_l$ is exactly zero in FP32 for 1,143 of 2,000 questions (57.15\%), which would merge the first 11 candidate thresholds, whereas $u$ takes 1,928 distinct values. This is the tie-breaking score of Appendix~\ref{app:restson}, chosen from the fit-split score distribution before any calibration output existed. A non-finite score routes to the reference (none occurred).

\paragraph{Candidates, tests, and labels.} Candidate $j=1,\ldots,19$ is the $\lceil jN_{\rm fit}/20\rceil$-th smallest fit score; $j=20$ always skips; a question is skipped when $u\le\tau_j$. On calibration ($N=3{,}000$), let $n$ be the skipped questions, $m$ those on which $s$ and $b$ answer differently, and $l$ those on which $b$ is correct and $s$ is not (lost corrections). The change-rate test passes when $\Pr[\mathrm{Bin}(n,0.05)\le m]\le0.001$ ($n=0$ fails); the lost-correction test passes when $\Pr[\mathrm{Bin}(N,\varepsilon)\le l]\le0.001$. Gold labels are queried in increasing $\tau$, only for newly skipped disagreements; a candidate is decided after all its disagreements are labeled; for each $\varepsilon$ the scan stops at the first candidate that fails the lost-correction test; $\varepsilon=0.02$ runs first, and no label is queried beyond the largest candidate that passes the change-rate test. The policies are P0, the largest candidate passing the change-rate test (label-free); P1 (primary), the largest passing both tests at $\varepsilon=0.01$; and P2, the same at $\varepsilon=0.02$. With 20 change-rate and 40 lost-correction tests at level 0.001, all calibration conclusions hold jointly with probability at least 0.94 (Bonferroni).

\paragraph{Calibration.} Table~\ref{tab:e1_cal} lists the candidates that decide the three policies, and the code repository lists all 20. P0 is $j=11$ (calibration coverage 53.7\%), P1 is $j=8$ (39.0\%), and P2 is $j=10$ (48.2\%). The label rule queried 40 of the 3,000 calibration labels, all disagreements at $j\le11$: 39 were lost corrections and one was wrong on both paths. Before calibration, the label-free plug-in rule of Appendix~\ref{app:predict}, applied to fit counts, predicted P0 at $j=12$; $j=12$ narrowly fails ($p=0.0012$).

\begin{table}[!htbp]\centering\scriptsize\revcolor
\caption{\rev{Calibration tests ($N=3{,}000$) at the candidates that decide the three policies: each policy's candidate and the next one. $n$: skipped; $m$: answer changes among them; $p_{\rm change}=\Pr[\mathrm{Bin}(n,0.05)\le m]$; $l$: lost corrections; $p_{\varepsilon}=\Pr[\mathrm{Bin}(N,\varepsilon)\le l]$. A test passes at $p\le0.001$. --: beyond the largest candidate passing the change-rate test, so not labeled. Bold: P1 ($j=8$), P2 ($j=10$), P0 ($j=11$).}}\label{tab:e1_cal}
\begin{tabular}{rrrrrrrr}\toprule
$j$ & $q$ & $n$ & $m$ & $p_{\rm change}$ & $l$ & $p_{0.01}$ & $p_{0.02}$\\\midrule
\textbf{8} & 0.40 & 1,171 & 7 & $6.09\times10^{-18}$ & 7 & $4.78\times10^{-7}$ & $3.41\times10^{-18}$\\
9 & 0.45 & 1,330 & 18 & $7.17\times10^{-13}$ & 18 & 0.0126 & $1.45\times10^{-10}$\\
\textbf{10} & 0.50 & 1,446 & 24 & $1.70\times10^{-11}$ & 23 & 0.113 & $3.38\times10^{-8}$\\
\textbf{11} & 0.55 & 1,612 & 40 & $2.45\times10^{-7}$ & 39 & 0.955 & 0.00235\\
12 & 0.60 & 1,783 & 62 & 0.0012 & -- & -- & --\\\bottomrule
\end{tabular}
\end{table}

\paragraph{Sealed test.} The pass rules use the upper end of the two-sided 95\% Clopper--Pearson interval, as in the main text: R1, changes among skipped questions below 5\%; R2, lost corrections over all 2,000 questions below $\varepsilon$; the lower ends of the paired-bootstrap 95\% intervals (2,000 resamples) of $G_B=\mathrm{acc}(b)-\mathrm{acc}(s)$ and $G_\pi=\mathrm{acc}(\pi)-\mathrm{acc}(s)$ above zero; and, on the development timing panel, a saving whose lower end is above zero. ``Keeps at least half of the gain'' requires the lower end of $G_\pi-G_B/2$ to exceed zero. B raises accuracy from 63.80\% ($s$) to 76.60\% ($b$), $G_B=12.80$ points [11.30, 14.40]. All three policies pass R1, both joint policies pass R2 at their $\varepsilon$, and all three keep more than half of the gain (Table~\ref{tab:e1_sealed}). B changes 16.4\% of C's answers (328 of 2,000), yet only 5 of these changes fall among P1's 798 skipped questions.

\begin{table}[!htbp]\centering\scriptsize\revcolor
\caption{\rev{Sealed test (2,000 questions, evaluated once). Upper bounds: upper end of the two-sided 95\% Clopper--Pearson interval. Gains in points with paired-bootstrap 95\% CIs. $s$: 63.80\%; $b$: 76.60\%; $G_B=12.80$ [11.30, 14.40].}}\label{tab:e1_sealed}
\setlength{\tabcolsep}{3pt}
\begin{tabular}{lcccccc}\toprule
 & & Changed / skipped & Lost corrections & Accuracy & $G_\pi$ & $G_\pi-G_B/2$\\
Policy & Skipped & (bound) & (bound) & (\%) & [95\% CI] & [95\% CI]\\\midrule
P1 ($\varepsilon=0.01$) & 798 (39.9\%) & 5 / 798 (1.46\%) & 5 (0.58\%) & 76.35 & 12.55 [11.10, 14.15] & 6.15 [5.40, 6.95]\\
P2 ($\varepsilon=0.02$) & 991 (49.6\%) & 15 / 991 (2.48\%) & 14 (1.17\%) & 75.90 & 12.10 [10.65, 13.65] & 5.70 [4.97, 6.50]\\
P0 (label-free) & 1,094 (54.7\%) & 29 / 1,094 (3.78\%) & 26 (1.90\%) & 75.35 & 11.55 [10.15, 13.05] & 5.15 [4.40, 5.93]\\\bottomrule
\end{tabular}
\end{table}

\paragraph{Latency.} The panel is the first 128 development questions in a fixed pseudorandom order. A and B generate live; the policy and the reference alternate per request in counterbalanced order; batch size is one; there is no warm-up, and cold requests are kept. A and B share one GPU and C and the probe use another, on an exclusively allocated node with four A100-SXM4-40GB GPUs. Each policy ran once. Before the panel, four fit questions regenerated on this layout matched the single-GPU records exactly, and the replay reproduced the offline records (A's messages and both answers on all 256 requests, B's messages on all 205 requests that ran B, and $u$ on all 128 questions). Every policy saves time (Table~\ref{tab:e1_latency}). Skipped queries avoid B (about 1.3~s each), whereas queries that wait for B pay the probe (about 40~ms), so 56\% of P1's panel queries run slightly slower and its paired median saving is $-22$~ms.

\begin{table}[!htbp]\centering\scriptsize\revcolor
\caption{\rev{End-to-end latency on the 128-question development panel (ms; savings: reference minus policy, paired-bootstrap 95\% CIs). Components are means over all 128 requests of each arm, with steps not run counted as zero.}}\label{tab:e1_latency}
\begin{tabular}{lcccccc}\toprule
 & Skipped & \multicolumn{2}{c}{Mean latency} & Mean saving & Relative & Components $T_A$ / $T_{\rm probe}$ / $T_B$ / $T_C$\\\cmidrule(lr){3-4}
Policy & on panel & Policy & Reference & [95\% CI] & saving (\%) & policy; reference\\\midrule
P1 & 39.8\% & 1,573 & 2,056 & 483 [365, 604] & 23.5 [17.6, 29.5] & 538 / 39 / 797 / 199; 534 / -- / 1,283 / 239\\
P2 & 50.8\% & 1,400 & 2,017 & 618 [501, 732] & 30.6 [24.7, 36.6] & 526 / 39 / 644 / 190; 524 / -- / 1,255 / 238\\
P0 & 56.2\% & 1,355 & 2,051 & 697 [581, 818] & 34.0 [28.1, 40.0] & 536 / 40 / 588 / 190; 532 / -- / 1,278 / 241\\\bottomrule
\end{tabular}
\end{table}

\paragraph{Controls fixed in the protocol.} On development (descriptive; Table~\ref{tab:e1_controls}), $G_B=14.4$ points [12.2, 16.7], and P0, P1, and P2 skip 54.9, 39.1, and 48.4\% of questions. C reading the two facts directly is more accurate than the pipeline (+15.7 points [13.3, 18.2]) and faster, so when the facts can be handed to C, the pipeline can be simplified; the experiment tests skipping within a fixed information flow, not the choice of architecture. B without A is also more accurate than A+B+C, and A with a fact-free B matches A+C (63.6 vs.\ 63.1\%). On fit, C alone answers 60.1\% and C reading the facts 92.7\%. A's message names an option in 9.55, 11.1, and 9.85\% of fit, development, and sealed questions, and B's message contains the correct option's text in 72.0, 72.4, and 72.2\%.

\begin{table}[!htbp]\centering\scriptsize\revcolor
\caption{\rev{Development controls and ablations (1,000 questions; descriptive). Accuracy with question-bootstrap 95\% CIs; generation time is the mean per question on one A100-SXM4-40GB from the offline runs, not the timing panel.}}\label{tab:e1_controls}
\begin{tabular}{lcc}\toprule
Path & Accuracy (\%) [95\% CI] & Mean generation time (ms)\\\midrule
C alone & 61.7 [58.7, 64.7] & 197\\
A+C (cheap path $s$) & 63.1 [60.2, 66.2] & 715\\
A + B without facts + C & 63.6 [60.7, 66.6] & 2,188\\
A+B+C (reference $b$) & 77.5 [74.9, 80.1] & 2,092\\
B+C (no A) & 85.2 [83.0, 87.3] & 1,205\\
C reading the two facts & 93.2 [91.7, 94.8] & 195\\\bottomrule
\end{tabular}
\end{table}

\paragraph{Deviations.} Thirteen entries record every difference between the protocol and its execution, each with its timing relative to the first reading of calibration, development, sealed-test, and timing outputs; the code repository lists them. None changes a model input, the score, a primary endpoint, or a sample rule.

\endgroup

\FloatBarrier
\subsection{A 32B helper and receiver pair (OBQA)}\label{app:e2b}
\begingroup\revcolor

This subsection repeats the label-assisted certificate of Appendix~\ref{app:e1} with both agents scaled to 32B parameters. OBQA is used throughout this paper, so the experiment extends model scale, not data. The protocol (models, prompts, score, candidate thresholds, tests, label rule, policies, and pass rules) was committed before any calibration, development, or held-out output of these models existed, and the held-out questions were evaluated once. The primary policy meets every pass rule. On the 744 held-out questions it skips the helper agent on 68.4\% of questions, changes 2 answers, loses 1 correction (95\% upper bound 0.75\%), and matches the accuracy of always sending Text (92.88\%); on a live development panel it lowers mean latency by 1,377~ms (56\%).

\paragraph{System and data.} The helper agent is Qwen2.5-32B-Instruct and the receiver agent is Qwen3-32B with thinking disabled. The reference is the Text path and the cheap path is the receiver agent alone, with the prompts, decoding, and parser of the main experiments (Appendices~\ref{app:protocol} and~\ref{app:prompts}). Both models run in bfloat16 in Hugging Face Transformers with greedy decoding and batch size one, each across two A100-SXM4-40GB GPUs of one node, and the same backend is used from fit to evaluation. The splits are those of the main OBQA experiments: 2,100 fit, 1,366 calibration, 742 development, and 744 held-out questions. The score is that of Appendix~\ref{app:e1}, the log of the non-top mass over the labels A to D in FP32; no score was non-finite.

\paragraph{Procedure.} Candidates, tests, label rule, and policies are those of Appendix~\ref{app:e1} with $N=1{,}366$: the label-free policy is the largest candidate passing the change-rate test, and the joint policies are the largest passing both tests at $\varepsilon=0.01$ (primary) and $\varepsilon=0.02$. The pass rules on the held-out questions are R1 (95\% upper bound of changes among skipped questions below 5\%), R2 (95\% upper bound of lost corrections over all 744 questions below $\varepsilon$, which allows at most 2 lost corrections at $\varepsilon=0.01$ and 7 at $\varepsilon=0.02$), lower ends of the paired-bootstrap intervals of $G=\mathrm{acc}(\text{Text})-\mathrm{acc}(R)$ and $G_\pi=\mathrm{acc}(\pi)-\mathrm{acc}(R)$ above zero, and a latency saving on the development panel whose lower end is above zero. A pilot on the first 300 fit questions was to decide only whether gain retention would be emphasized (at least 6 more correct answers with Text): Text answered 276 and the receiver agent alone 271. The authors had committed, before the pilot result, to run the full protocol in either case.

\paragraph{Calibration.} Table~\ref{tab:e2b_cal} lists the candidates that decide the three policies, and the code repository lists all 20. The label-free policy is $j=17$ (calibration coverage 83.7\%), the primary policy is $j=14$ (67.6\%), and the $\varepsilon=0.02$ policy is $j=16$ (79.0\%). The label rule queried 21 of the 1,366 calibration labels, all disagreements at $j\le17$: 13 were lost corrections, 7 were questions the receiver agent alone answers correctly and Text does not, and one was wrong on both paths. On fit, Text answers 92.95\% and the receiver agent alone 90.29\%.

\begin{table}[!htbp]\centering\scriptsize\revcolor
\caption{\rev{Calibration tests for the 32B pair ($N=1{,}366$) at the candidates that decide the three policies: each policy's candidate and the next one; notation as in Table~\ref{tab:e1_cal}. --: beyond the largest candidate passing the change-rate test, so not labeled. Bold: the primary policy ($j=14$), the $\varepsilon=0.02$ policy ($j=16$), and the label-free policy ($j=17$).}}\label{tab:e2b_cal}
\begin{tabular}{rrrrrrrr}\toprule
$j$ & $q$ & $n$ & $m$ & $p_{\rm change}$ & $l$ & $p_{0.01}$ & $p_{0.02}$\\\midrule
\textbf{14} & 0.70 & 924 & 5 & $6.51\times10^{-15}$ & 3 & $5.96\times10^{-4}$ & $4.16\times10^{-9}$\\
15 & 0.75 & 996 & 7 & $1.58\times10^{-14}$ & 5 & 0.00675 & $1.74\times10^{-7}$\\
\textbf{16} & 0.80 & 1,079 & 12 & $2.57\times10^{-12}$ & 8 & 0.0722 & $1.27\times10^{-5}$\\
\textbf{17} & 0.85 & 1,144 & 21 & $1.93\times10^{-8}$ & 13 & 0.501 & 0.00173\\
18 & 0.90 & 1,229 & 39 & 0.00117 & -- & -- & --\\\bottomrule
\end{tabular}
\end{table}

\paragraph{Held-out questions.} Text raises accuracy from 90.32\% (receiver agent alone) to 92.88\%, $G=2.55$ points [0.94, 4.30], and changes the receiver agent's answer on 45 of the 744 questions (6.0\%). All three policies pass R1, both joint policies pass R2 at their $\varepsilon$, and all three keep a gain whose interval excludes zero (Table~\ref{tab:e2b_heldout}). The primary policy's two changed answers are one lost correction and one question that only the receiver agent alone answers correctly, so its accuracy equals that of the reference. The joint policies keep at least half of the gain; for the label-free policy the lower end of $G_\pi-G/2$ is zero, so we do not make that statement, and it does not pass R2 at $\varepsilon=0.01$.

\begin{table}[!htbp]\centering\scriptsize\revcolor
\caption{\rev{Held-out questions for the 32B pair (744 questions, evaluated once). Upper bounds: upper end of the two-sided 95\% Clopper--Pearson interval. Gains in points with paired-bootstrap 95\% CIs (2,000 resamples). Receiver agent alone: 90.32\%; Text: 92.88\%; $G=2.55$ [0.94, 4.30].}}\label{tab:e2b_heldout}
\setlength{\tabcolsep}{3pt}
\begin{tabular}{lcccccc}\toprule
 & & Changed / skipped & Lost corrections & Accuracy & $G_\pi$ & $G_\pi-G/2$\\
Policy & Skipped & (bound) & (bound) & (\%) & [95\% CI] & [95\% CI]\\\midrule
Joint, $\varepsilon=0.01$ & 509 (68.4\%) & 2 / 509 (1.41\%) & 1 (0.75\%) & 92.88 & 2.55 [0.94, 4.17] & 1.28 [0.40, 2.15]\\
Joint, $\varepsilon=0.02$ & 594 (79.8\%) & 6 / 594 (2.19\%) & 4 (1.37\%) & 92.61 & 2.28 [0.67, 3.90] & 1.01 [0.13, 1.88]\\
Label-free & 632 (84.9\%) & 9 / 632 (2.69\%) & 6 (1.75\%) & 92.47 & 2.15 [0.67, 3.76] & 0.87 [0.00, 1.75]\\\bottomrule
\end{tabular}
\end{table}

\paragraph{Development questions and single models.} The development questions are reported descriptively and decide nothing (Table~\ref{tab:e2b_dev}). The policies behave as on the held-out questions (coverage 69.3 to 85.4\%, 1 to 10 changed answers), but Text's gain is smaller, $G=1.08$ points [$-0.54$, 2.70], and neither it nor the policies' gains exclude zero there. The helper agent answering alone, with the same free-generation protocol as the other rows, reaches 91.26\% on the held-out questions and 91.91\% on development; the pair with Text is 1.61 points [$-0.27$, 3.63] above it on the held-out questions.

\begin{table}[!htbp]\centering\scriptsize\revcolor
\caption{\rev{Development questions for the 32B pair (742 questions; descriptive). Notation as in Table~\ref{tab:e2b_heldout}. Receiver agent alone: 91.91\%; Text: 92.99\%; helper agent alone: 91.91\%; $G=1.08$ [$-0.54$, 2.70].}}\label{tab:e2b_dev}
\setlength{\tabcolsep}{3pt}
\begin{tabular}{lccccc}\toprule
 & & Changed / skipped & Lost corrections & Accuracy & $G_\pi$\\
Policy & Skipped & (bound) & (bound) & (\%) & [95\% CI]\\\midrule
Joint, $\varepsilon=0.01$ & 514 (69.3\%) & 1 / 514 (1.08\%) & 1 (0.75\%) & 92.86 & 0.94 [$-0.67$, 2.43]\\
Joint, $\varepsilon=0.02$ & 598 (80.6\%) & 5 / 598 (1.94\%) & 2 (0.97\%) & 93.13 & 1.21 [$-0.27$, 2.70]\\
Label-free & 634 (85.4\%) & 10 / 634 (2.88\%) & 6 (1.75\%) & 92.72 & 0.81 [$-0.54$, 2.16]\\\bottomrule
\end{tabular}
\end{table}

\paragraph{Latency.} The panel is the 128-question OBQA development panel of the main experiments. For each question and policy, the policy, the Text reference, and the receiver agent alone run live in rotating order at batch size one, on exclusively allocated nodes. The panel ran in four segments of 32 questions; the last three start with untimed warm-up requests on questions outside the panel, and the first has no additional warm-up beyond a two-question consistency check that every job runs. The replayed policy answers equal the offline answers on every request. Every policy saves time (Table~\ref{tab:e2b_latency}), and dropping the first question of each segment changes no conclusion (primary policy: 1,377~ms [1,213, 1,538] on 124 questions). The replay records whole-request latency, not component times, so the table splits the paired difference by route: a skipped question saves about 1.86~s, and a question that waits pays about 90~ms for the probe. The receiver agent alone takes 486~ms.

\begin{table}[!htbp]\centering\scriptsize\revcolor
\caption{\rev{End-to-end latency of the 32B pair on the 128-question development panel (ms; savings: reference minus policy, paired-bootstrap 95\% CIs). Route split: mean paired saving on skipped questions, and mean extra latency on questions that wait.}}\label{tab:e2b_latency}
\begin{tabular}{lcccccc}\toprule
 & Skipped & \multicolumn{2}{c}{Mean latency} & Mean saving & Relative & Saved if skipped;\\\cmidrule(lr){3-4}
Policy & on panel & Policy & Reference & [95\% CI] & saving (\%) & extra if waiting\\\midrule
Joint, $\varepsilon=0.01$ & 75.0\% & 1,064 & 2,441 & 1,377 [1,219, 1,533] & 56.4 & 1,866; 93\\
Joint, $\varepsilon=0.02$ & 82.8\% & 917 & 2,442 & 1,525 [1,377, 1,667] & 62.5 & 1,860; 88\\
Label-free & 87.5\% & 817 & 2,435 & 1,618 [1,485, 1,750] & 66.4 & 1,862; 91\\\bottomrule
\end{tabular}
\end{table}

\paragraph{Deviations.} Every difference between the protocol and its execution was recorded as it occurred; the code repository lists them. None changes a model input, the score, a test, a policy, or a pass rule. The label log-probabilities behind the score and the component times of the replay were not saved; the score code, which was fixed before the run, and the route split above stand in for them.

\endgroup

%% file: new_app_G.tex
\section{Latency}\label{app:e2e}

\subsection{Batch size one}\label{app:bs1}

\begin{figure}[!t]\centering
\includegraphics[width=\linewidth]{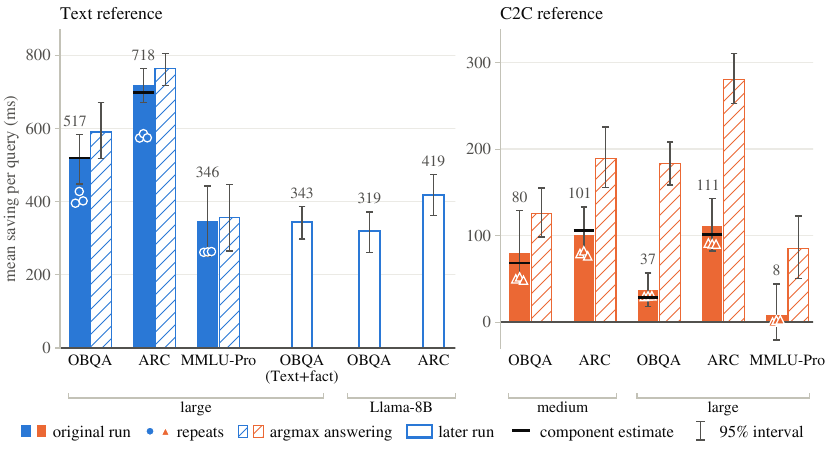}
\caption{Every saving class holds across configurations, repeats, and probe-cache reuse (under argmax answering, large/MMLU-Pro/C2C's interval moves above zero), and the component estimates fixed before the replays are close to the complete replays. Mean saving per query of each replayed policy, Text-reference policies left and C2C-reference policies right (separate scales), with paired bootstrap 95\% intervals as whiskers: solid bar, the original run (first configuration; Table~\ref{tab:e2e_full}), its mean printed above; markers on it, the means of the three repeats on the second configuration (Table~\ref{tab:repeats}); outlined bar, the single later run of Text+fact or a Llama-8B policy on the second configuration (Table~\ref{tab:u0saving}); hatched bar, answering skipped questions from the probe's argmax (second configuration; Table~\ref{tab:impl}); black tick, the component estimate (large-pair OBQA and ARC; from timed probe stages and historical action requests) or CPU recombination (medium pair; from development timings) fixed before the replays. Saving class: interval above, below, or including zero. Savings from the two configurations are not comparable in absolute terms; only large/MMLU-Pro/C2C has an interval containing zero, in every original and repeat run.}
\label{fig:savings}
\end{figure}

Every replayed OBQA and ARC policy saves time, all of it on skipped queries, and every saving class holds in three repeats on a second hardware configuration (Figure~\ref{fig:savings}). Replays reuse the frozen thresholds and 128-question panels without recalibration, and all replayed outputs matched their development or same-replay fixed-reference counterparts (including every route and output of the medium-pair C2C replays), so only latency is re-measured (Table~\ref{tab:e2e_full}).

\begin{table}[!htbp]\centering\scriptsize
\caption{Every deployed Qwen-pair policy saves time on its OBQA or ARC panel and on the sealed ARC test; on MMLU-Pro only the Text saving is established. (a), (b): development replays on 128-question panels; (c): sealed test (1,172 questions); all on the first configuration. Saving: paired mean of fixed reference minus policy, every probe and cold request included [paired bootstrap 95\% CI]; not comparable across pairs or benchmarks. Cold request: a path's first request after model loading. $\Delta$acc: policy minus reference [paired bootstrap 95\% CI].}
\label{tab:e2e_full}
\setlength{\tabcolsep}{2pt}
\begin{tabular*}{\linewidth}{@{}l@{\hspace{6pt}}l@{\hspace{4pt}\extracolsep{\fill}}rcclccl@{}}\toprule
& & & \multicolumn{2}{c}{Mean latency (ms)} & & \multicolumn{2}{c}{Correct answers} & \\\cmidrule(lr){4-5}\cmidrule(lr){7-8}
Population & Reference & \multicolumn{1}{c}{Skipped} & Policy & Reference & \multicolumn{1}{c}{Saving (ms) [95\% CI]} & Policy & Reference & \multicolumn{1}{c@{}}{$\Delta$acc (points) [95\% CI]}\\\midrule
\multicolumn{9}{@{}l}{\emph{(a) Development end-to-end replay on 128-question panels, OBQA and ARC}}\\
medium/OBQA & C2C & 66/128 & 335.6 & 415.2 & 79.7 [48.1, 129.1] & 94 & 93 & $+0.78$ [0.00, $+2.34$]\\
medium/ARC & C2C & 77/128 & 352.4 & 453.1 & 100.7 [74.5, 133.1] & 85 & 85 & $+0.00$ [0.00, 0.00]\\
large/OBQA & Text & 97/128 & 505.1 & 1022.2 & 517.1 [447.8, 583.6] & 112 & 113 & $-0.78$ [$-3.91$, $+1.56$]\\
large/OBQA & C2C & 97/128 & 325.9 & 362.4 & 36.6 [18.0, 56.8] & 102 & 101 & $+0.78$ [0.00, $+2.34$]\\
large/ARC & Text & 120/128 & 372.7 & 1090.3 & 717.6 [670.6, 763.2] & 117 & 117 & $+0.00$ [$-2.34$, $+2.34$]\\
large/ARC & C2C & 115/128 & 329.9 & 440.8 & 110.9 [82.0, 142.8] & 115 & 114 & $+0.78$ [$-1.56$, $+3.91$]\\
\midrule
\multicolumn{9}{@{}l}{\emph{(b) Development replay, MMLU-Pro (same threshold and route set for both references)}}\\
large/MMLU-Pro & Text & 48/128 & 984.3 & 1330.7 & 346.3 [249.8, 443.0] & 65 & 63 & $+1.56$ [0.00, $+3.91$]\\
large/MMLU-Pro & C2C & 48/128 & 332.6 & 340.8 & 8.2 [$-20.5$, $+44.0$] & 55 & 55 & $+0.00$ [0.00, 0.00]\\
\midrule
\multicolumn{9}{@{}l}{\emph{(c) Sealed ARC-Challenge test}}\\
large/ARC & Text & 1100/1172 & 362.0 & 1047.7 & 685.6 [669.6, 700.8] & 1080 & 1085 & $-0.43$ [$-1.19$, $+0.26$]\\
large/ARC & C2C & 1047/1172 & 323.6 & 414.5 & 90.9 [81.0, 101.0] & 1061 & 1057 & $+0.34$ [$-0.17$, $+0.85$]\\
\bottomrule\end{tabular*}
\end{table}

\paragraph{Per-request distribution.} The policy's 99th-percentile latency is never above the reference's in any replay; its 90th percentile is above it only for medium/OBQA/C2C and, on the second configuration, large/MMLU-Pro/Text. Routed queries are slower in 84.6 to 100\% of cases (median $-25$ to $-45$~ms, the probe cost), meaning the entire saving stems from skipped queries (also under argmax answering), and the paired median is negative in 16 of 58 arm-streams. The probe costs 32.88 and 43.37~ms on the small- and large-pair OBQA panels, of which the prefill with its last-position projection takes 31.25 and 41.79~ms (95 to 96\%), and 48.4 and 46.1~ms on the MMLU-Pro replay, whose paired median savings are $-41.3$ and $-43.6$~ms (median latencies 1170.6 vs.\ 1273.1~ms for Text, and 293.6 vs.\ 250.9~ms for C2C). The probe, vendor runner, and official fuser are single-row, so no batched comparison is reported.

\paragraph{Component estimates and complete replays.} Before any replay, each policy's cost was recomposed from timed components (the probe's stages plus the selected historical action request), without cache or prefill discounts. For the four large-pair policies, the complete replay saved 1.76~ms less than this estimate for OBQA/Text and 8.40, 20.50, and 9.74~ms more for OBQA/C2C, ARC/Text, and ARC/C2C (Figure~\ref{fig:savings}). A CPU recombination of development timings had projected 68.3 and 105.4~ms for the medium pair, whose replays saved 79.7 and 100.7~ms (median latencies 334.6 vs.\ 388.2~ms on OBQA, and 287.8 vs.\ 412.4~ms on ARC; paired median savings 79.3 and 81.6~ms).

\paragraph{Busy GPU time.} Under rules fixed before computing, busy GPU time (summed wall time of GPU-side stages on both GPUs) can be computed only for replays timing each stage separately; thus, it is available for four of the eight policies (large-pair OBQA and ARC replays), where the busy-time saving accounts for 88 to 99\% of the latency saving (Table~\ref{tab:busy}).

\begin{table}[!htbp]\centering\scriptsize
\caption{Busy GPU time saved is 88 to 99\% of the latency saved on the four large-pair panels. Original runs (first configuration). Busy GPU time: summed wall time of GPU-side stages on both GPUs. Intervals: paired bootstrap 95\% CI.}
\label{tab:busy}
\setlength{\tabcolsep}{2pt}
\begin{tabular*}{\linewidth}{@{}l@{\hspace{6pt}}l@{\hspace{6pt}}l@{\hspace{4pt}\extracolsep{\fill}}cccc@{}}\toprule
& & & \multicolumn{3}{c}{Saving (ms) [95\% CI]} & \\\cmidrule(lr){4-6}
Pair & Task & Reference & Busy GPU time & Latency & Busy minus latency & \multicolumn{1}{c@{}}{Busy / latency}\\\midrule
large & OBQA & Text & 513.5 [448.7, 574.2] & 517.1 [447.8, 583.6] & $-3.6$ [$-9.9$, $+8.3$] & 0.993\\
large & OBQA & C2C & 32.3 [14.4, 51.8] & 36.6 [18.0, 56.8] & $-4.3$ [$-6.3$, $-3.2$] & 0.882\\
large & ARC & Text & 705.2 [659.1, 750.0] & 717.6 [670.6, 763.2] & $-12.4$ [$-13.2$, $-11.6$] & 0.983\\
large & ARC & C2C & 106.6 [77.8, 138.2] & 110.9 [82.0, 142.8] & $-4.3$ [$-4.6$, $-3.9$] & 0.961\\\bottomrule
\end{tabular*}
\end{table}

\paragraph{Second configuration and port check.}\label{app:revision} Repeats and variants, added after main results under protocols fixed before outcomes, ran on exclusively allocated nodes with four A100-SXM4-40GB GPUs connected via NVLink (two used with the original placement, while original runs used two of eight A100-SXM4-40GB GPUs on a shared node). The repeats followed bit-for-bit reproduction of saved outputs: the receiver-only outputs of all seven development populations, large-pair ARC receiver-only and C2C outputs, and learned predictor features and predictions. In a post-hoc port check, our runtime and the official C2C evaluator \citep{fu2025c2c}, using byte-identical prompts, produce identical raw outputs on all 500 official OBQA test questions for every pair and action. Small-pair C2C answers 263/500 correctly, matching published results, whereas medium and large C2C answer fewer correctly than $R$ under our parser (317 vs.\ 327 and 405 vs.\ 425) and official scoring (321 vs.\ 326 and 395 vs.\ 425; small pair: 263 vs.\ 198). The published main evaluation uses Qwen3-0.6B as its only receiver, and the authors' scaling study uses several Qwen3 receivers with fusers trained on MMLU's auxiliary training split and reports smaller relative gains for larger receivers \citep{fu2025c2c}, so no published accuracy exists for released general-purpose medium and large fusers; for those pairs, byte-identical outputs serve as the relevant check.

\paragraph{Repeated replays.} Under a rule fixed before any repeat, each of the eight replays ran three additional times on the second configuration, maintaining frozen thresholds, panels, arm rotation, and the timing boundary. The rule defines a policy's class on savings excluding questions where either arm made its first request after model loading (Table~\ref{tab:repeats}b): \checkmark\ if the interval lies above zero, $\times$ if below, and ? otherwise. Original classes had to be reproduced by all eight policies across all three repeats, which they were; the classes are the same over all requests (Table~\ref{tab:repeats}a). Online scores, inputs, and routes matched frozen records in every repeat, and every output matched its development or same-run counterpart. Absolute savings differ across configurations and are not comparable; large/MMLU-Pro/Text saves 261.4 to 263.5~ms there, and MMLU-Pro paired medians are negative ($-21$ to $-41$~ms; $-21$ to $-32$~ms for Text), mirroring the original panel.

\begin{table}[!htbp]\centering\scriptsize
\caption{Every replayed policy keeps its prespecified saving class in all three repeats. Paired mean saving (ms) [95\% CI]. Saving class: all four intervals of the row lie above zero (\checkmark) or all include zero (?). Cold first request: a path's first request after model loading. Configurations are not comparable.}
\label{tab:repeats}
\setlength{\tabcolsep}{1pt}
\begin{tabular*}{\linewidth}{@{}l@{\hspace{3pt}\extracolsep{\fill}}llccccl@{}}\toprule
& & & First configuration & \multicolumn{3}{c}{Second configuration} & \\\cmidrule(lr){4-4}\cmidrule(lr){5-7}
Pair & Task & Reference & Original run & Repeat 1 & Repeat 2 & Repeat 3 & Saving class\\\midrule
\multicolumn{8}{l}{\emph{(a) All requests}}\\
large & OBQA & Text & 517.1 [447.8, 583.6] & 395.5 [325.3, 458.3] & 427.6 [376.4, 474.4] & 401.8 [335.6, 460.6] & \checkmark\ above 0\\
large & OBQA & C2C & 36.6 [18.0, 56.8] & 29.4 [14.9, 45.5] & 30.3 [15.7, 46.5] & 30.4 [15.5, 46.6] & \checkmark\ above 0\\
large & ARC & Text & 717.6 [670.6, 763.2] & 573.5 [536.0, 608.7] & 585.4 [546.7, 622.9] & 574.2 [536.3, 611.2] & \checkmark\ above 0\\
large & ARC & C2C & 110.9 [82.0, 142.8] & 91.4 [68.9, 116.2] & 90.5 [67.3, 115.4] & 90.2 [66.9, 115.8] & \checkmark\ above 0\\
medium & OBQA & C2C & 79.7 [48.1, 129.1] & 49.9 [35.8, 65.5] & 52.2 [37.7, 67.9] & 48.5 [34.3, 64.4] & \checkmark\ above 0\\
medium & ARC & C2C & 100.7 [74.5, 133.1] & 79.2 [58.5, 104.5] & 82.4 [61.4, 108.4] & 76.5 [56.5, 101.2] & \checkmark\ above 0\\
large & MMLU-Pro & Text & 346.3 [249.8, 443.0] & 261.4 [191.1, 331.5] & 262.4 [192.3, 331.1] & 263.5 [193.5, 333.7] & \checkmark\ above 0\\
large & MMLU-Pro & C2C & 8.2 [$-20.5$, $+44.0$] & $-0.1$ [$-23.4$, $+27.8$] & 2.4 [$-20.3$, $+30.6$] & 2.6 [$-20.1$, $+30.4$] & ? includes 0\\\midrule
\multicolumn{8}{l}{\emph{(b) Questions with a cold first request excluded ($N=127$); the class rule uses these}}\\
large & OBQA & Text & 532.8 [470.9, 594.7] & 419.6 [370.7, 468.7] & 430.2 [380.9, 479.9] & 423.1 [373.6, 471.5] & \checkmark\ above 0\\
large & OBQA & C2C & 33.5 [16.5, 51.7] & 27.4 [13.8, 42.3] & 29.3 [14.9, 44.2] & 27.8 [13.9, 42.5] & \checkmark\ above 0\\
large & ARC & Text & 717.2 [670.1, 761.8] & 573.2 [535.8, 609.6] & 585.0 [547.0, 621.8] & 573.8 [535.2, 610.6] & \checkmark\ above 0\\
large & ARC & C2C & 112.3 [83.7, 142.3] & 92.4 [70.2, 115.8] & 91.6 [68.5, 115.7] & 91.2 [68.9, 115.1] & \checkmark\ above 0\\
medium & OBQA & C2C & 61.0 [43.9, 79.9] & 47.3 [34.0, 61.9] & 50.0 [36.4, 65.2] & 45.6 [32.5, 60.1] & \checkmark\ above 0\\
medium & ARC & C2C & 100.8 [73.0, 132.8] & 79.3 [57.6, 104.3] & 82.5 [60.1, 108.4] & 76.6 [55.4, 101.1] & \checkmark\ above 0\\
large & MMLU-Pro & Text & 322.1 [234.8, 412.5] & 260.8 [189.9, 333.9] & 261.0 [189.7, 335.1] & 262.7 [191.1, 336.5] & \checkmark\ above 0\\
large & MMLU-Pro & C2C & 6.5 [$-23.3$, $+41.1$] & $-0.0$ [$-24.5$, $+28.3$] & 2.6 [$-21.1$, $+30.5$] & 2.8 [$-20.7$, $+30.8$] & ? includes 0\\\bottomrule
\end{tabular*}
\end{table}

\begin{table}[!htbp]\centering\scriptsize
\caption{Answering skipped questions from the probe's argmax raises every saving, whereas reusing the probe's cache changes none by more than 5~ms (one run on the second configuration; paired mean saving, ms [95\% CI]). Original: receiver rerun after the probe.}
\label{tab:impl}
\setlength{\tabcolsep}{2pt}
\begin{tabular*}{\linewidth}{@{}l@{\hspace{6pt}}l@{\hspace{6pt}}l@{\hspace{4pt}\extracolsep{\fill}}rrr@{}}\toprule
& & & \multicolumn{3}{c@{}}{Paired mean saving (ms) [95\% CI]}\\\cmidrule(l){4-6}
Pair & Task & Reference & \multicolumn{1}{c}{Original} & \multicolumn{1}{c}{Reuse} & \multicolumn{1}{c@{}}{Argmax}\\\midrule
large & OBQA & Text & 436.3 [378.6, 503.2] & 437.8 [378.1, 506.0] & 589.7 [518.0, 670.3]\\
large & OBQA & C2C & 31.2 [16.7, 47.3] & 31.0 [16.5, 47.0] & 182.9 [158.2, 208.1]\\
large & ARC & Text & 565.6 [528.6, 601.4] & 564.4 [527.5, 599.6] & 762.7 [717.4, 804.3]\\
large & ARC & C2C & 91.9 [69.5, 116.6] & 90.5 [68.3, 115.0] & 280.8 [252.7, 310.4]\\
medium & OBQA & C2C & 50.1 [36.0, 65.6] & 49.9 [35.7, 65.6] & 125.2 [98.3, 154.9]\\
medium & ARC & C2C & 77.9 [57.4, 103.0] & 78.2 [57.5, 103.5] & 189.5 [155.4, 225.4]\\
large & MMLU-Pro & Text & 275.2 [199.6, 348.5] & 280.0 [203.5, 354.2] & 356.5 [265.1, 446.7]\\
large & MMLU-Pro & C2C & 4.6 [$-18.0$, $+32.6$] & 6.9 [$-15.8$, $+35.4$] & 85.0 [50.3, 122.8]\\\bottomrule
\end{tabular*}
\end{table}

\paragraph{Cheaper implementations.} Our replays rerun the receiver after the probe. In an additional run (protocol fixed before any output), four arms were interleaved per policy: the fixed reference, the original policy, a \emph{reuse} policy continuing the receiver-only answer from the probe's KV cache (the receiver prompt is an exact token prefix of the probe input), and an \emph{argmax} policy answering skipped questions from the probe's argmax without sending a receiver request. Argmax labels match native receiver answers on 88 to 100\% of calibration questions, and recertifying with them retains every threshold except medium/OBQA/C2C ($q=0.50$). Argmax answering boosts every saving, for instance MMLU-Pro C2C from 4.6 to 85.0~ms [50.3, 122.8], and narrows the Text-to-C2C saving ratio from 14.0 and 6.2 to 3.2 and 2.7 on OBQA and ARC (4.2 on MMLU-Pro). Its paired medians are negative on MMLU-Pro and medium/OBQA/C2C because gains concentrate on the skipped minority. Reuse alters no saving by more than 5~ms (Table~\ref{tab:impl}): it skips only the receiver's prefill of its own prompt, which changes a skipped request's action time by between 0.4~ms more and 6.3~ms less, while the probe is still paid on every query. Reuse outputs differed from native outputs on 13 of 668 skipped requests (exclusively via trailing periods or extended probe answer phrases) and never in parsed labels.

\begin{table}[!htbp]\centering\scriptsize
\caption{Only the two large MMLU-Pro policies are flagged by the prespecified rule (coverage off by more than 10 points (0.10) or an absolute length SMD (standardized mean difference) above 0.2). Population: all development questions. Eq.~\eqref{eq:saving} saving (ms) with the panel's latencies (first configuration; Text+fact and Llama-8B: second) and the panel's or the population's coverage.}
\label{tab:panelrep}
\renewcommand{\arraystretch}{1.1}\setlength{\tabcolsep}{3pt}
\begin{tabular*}{\linewidth}{@{}l@{\hspace{5pt}\extracolsep{\fill}}cccccc@{}}\toprule
& \multicolumn{2}{c}{Coverage} & \multicolumn{2}{c}{Largest $|$SMD$|$} & \multicolumn{2}{c}{Eq.~\eqref{eq:saving} saving (ms) with}\\\cmidrule(lr){2-3}\cmidrule(lr){4-5}\cmidrule(l){6-7}
Policy & panel & population & value & feature & panel coverage & population coverage\\\midrule
large/OBQA/Text & 0.758 & 0.771 & 0.136 & helper message & 517.2 & 527.1\\
large/OBQA/C2C & 0.758 & 0.771 & 0.115 & reference output & 36.6 & 38.0\\
large/ARC/Text & 0.938 & 0.950 & 0.053 & question & 717.8 & 727.8\\
large/ARC/C2C & 0.898 & 0.903 & 0.057 & reference output & 110.9 & 111.7\\
large/MMLU-Pro/Text & 0.375 & 0.405 & 0.212 & receiver output & 317.0 & 346.0\\
large/MMLU-Pro/C2C & 0.375 & 0.405 & 0.212 & receiver output & 5.6 & 9.7\\
medium/OBQA/C2C ($q=0.55$) & 0.516 & 0.526 & 0.107 & receiver output & 56.8 & 58.6\\
medium/ARC/C2C & 0.602 & 0.632 & 0.053 & question & 100.6 & 107.4\\
large/OBQA/Text+fact & 0.719 & 0.722 & 0.123 & helper message & 339.5 & 341.4\\
Llama-8B/OBQA/Text & 0.594 & 0.590 & 0.136 & helper message & 322.4 & 320.3\\
Llama-8B/ARC/Text & 0.688 & 0.689 & 0.069 & receiver output & 417.2 & 418.2\\\bottomrule
\end{tabular*}
\end{table}

\paragraph{Full MMLU-Pro population and panel representativeness.} Across all 2,641 MMLU-Pro development questions, the C2C policy saves a mean of 23.3~ms [14.5, 32.2] on the second configuration (22.0~ms [14.0, 30.5] excluding the single cold question; paired median $-37.3$~ms), an order of magnitude below the Text panel saving on that configuration. Under rules fixed before computing, each panel was compared with its full development population across question, message, and output lengths, share of $u=0$ questions, and coverage, flagging policies whose coverage differs by over 10 points or whose length SMD exceeds 0.2. Only the two large MMLU-Pro policies are flagged due to slightly longer receiver-only outputs (5.70 vs.\ 5.54 tokens; SMD 0.212) and 3.0 points lower coverage (Table~\ref{tab:panelrep}). Applying population coverage to Eq.~\eqref{eq:saving} alters the saving by 0.8 to 29.0~ms; C2C saves only 5.6~ms (9.7~ms with population coverage). The 128 panel questions of the full MMLU-Pro C2C replay save 4.5~ms, versus 23.3~ms across all 2,641 questions (SMD $-0.08$).

\paragraph{Later replays and costs.} A Text-only replay driver copy (22 changed lines, including two-arm rotation; an approved deviation) reproducing 16 saved rows byte for byte measured Llama-8B savings of 318.6~ms [261.1, 372.1] and 418.9~ms [361.7, 473.7] on the second configuration (fixed Text 770.3 and 818.4~ms, policy 451.7 and 399.5~ms; 325.5 and 417.6~ms excluding the first request), though 39 and 27\% of queries run slower since queries keeping Text also pay the probe. There, the Text+fact policy skips 92 of 128 panel questions, changes 5, reduces mean latency from 726.5 to 383.1~ms, a saving of 343.4~ms [297.7, 386.0] (paired median 448.5~ms), and answers 116 panel questions correctly vs.\ 121 for reference. Standalone calibration requires $2N_{\rm cal}$ machine requests ($3N_{\rm cal}$ comparing both references with shared $R$). Using mean savings from the second configuration, if reference outputs come from served traffic, fit and calibration requests amortize after 230 (large/ARC/Text) to 76,205 queries (large/MMLU-Pro/C2C), within 6,415 queries for every Text reference, and after 905 to 152,171 queries if references must also run. A small-LLM controller branch (small helper producing communicate-or-skip probabilities) halted at a predetermined cost screen: the controller alone took 21.6~ms (54.5~ms with probe), and even the fastest controller plus probe (51.7~ms) exceeded an optimistic 43.0~ms saving bound, precluding signal evaluation.

\FloatBarrier
\subsection{Batched serving}\label{app:e2a}
\begingroup\revcolor

Section~\ref{sec:rq3} measures latency at batch size one. This subsection replays the four deployed Text policies in static batches of 1, 8, and 32. With skipped requests returned as soon as they are answered, skipping lowers mean latency in all twelve settings, by 310 to 1,728~ms per request (35 to 67\% of the reference latency). The development questions and the deployed thresholds were already used in the main experiments, so this is a retrospective measurement of latency; the design was fixed before any of its outputs existed, after an earlier run of the whole-batch variant on the 128-question panels.

\paragraph{Design.} The policies are the deployed large-pair and Llama-8B Text policies on OBQA and ARC, with their thresholds unchanged. Each development set is replayed in full (742 OBQA and 299 ARC questions) in its original order, grouped into consecutive batches of $B\in\{1,8,32\}$. Three arms process every batch. The \emph{reference} generates the helper agent's message for all requests and then answers them. The \emph{early-return} policy runs the probe for the whole batch, answers the skipped requests and returns them, and then generates messages and answers for the rest. The \emph{whole-batch} policy also skips messages but returns all requests of a batch together. The arm order rotates from batch to batch. Each request's latency is measured from the start of its batch to its own completion. Models run in Hugging Face Transformers with the probe batched and prefix caching off, the helper agent on one A100-SXM4-40GB GPU and the receiver agent on another, on exclusively allocated nodes. Each of the four policies ran as one job; the first has no warm-up and the others start with untimed warm-up batches on questions outside the measured sets. Intervals are paired bootstraps over batches (20,000 resamples). A saving is called established when its 95\% interval lies above zero.

\paragraph{Latency.} Table~\ref{tab:e2a_latency} gives every arm. The early-return policy establishes a saving against the reference in all twelve settings, and the absolute saving grows with the batch size. The whole-batch policy establishes a saving in ten settings; for the Llama-8B receiver agent on OBQA its saving is 36~ms [$-3$, 82] at $B=8$ and $-63$~ms [$-121$, 1] at $B=32$, where the probe and the split of the batch offset the shorter helper call. Against the whole-batch policy, early return saves 200 to 1,120~ms at $B=8$ and $32$ (all eight intervals above zero); at $B=1$ the two schedules coincide and differ by at most 5~ms. Early return answers the skipped requests first, so at $B=8$ the requests that wait finish 110 to 168~ms later than under the whole-batch policy, and throughput is 0.5 to 0.9 requests per second lower; at $B=32$ neither cost appears. Dropping the first measured question of each job changes no conclusion.

\begin{table}[!htbp]\centering\scriptsize\revcolor
\caption{\rev{Batched replay of the four deployed Text policies on the full development sets. Mean latency per request (ms) for the reference (Ref.), the whole-batch policy (Whole), and the early-return policy (Early); savings against the reference with paired-batch bootstrap 95\% CIs; throughput in requests per second.}}\label{tab:e2a_latency}
\setlength{\tabcolsep}{3pt}
\begin{tabular}{lrrrrccc}\toprule
 & & \multicolumn{3}{c}{Mean latency} & Saving, early return & Saving, whole batch & Throughput\\\cmidrule(lr){3-5}
Policy & $B$ & Ref. & Whole & Early & [95\% CI] & [95\% CI] & Ref.\ / Whole / Early\\\midrule
large / OBQA & 1 & 797 & 380 & 376 & 421 [402, 441] & 416 [395, 437] & 1.26 / 2.63 / 2.66\\
 & 8 & 1,219 & 1,044 & 613 & 606 [565, 647] & 176 [116, 239] & 6.55 / 7.64 / 6.75\\
 & 32 & 2,293 & 2,134 & 1,232 & 1,061 [1,000, 1,120] & 159 [92, 224] & 13.73 / 14.72 / 14.75\\
large / ARC & 1 & 866 & 285 & 285 & 581 [558, 604] & 581 [558, 604] & 1.15 / 3.51 / 3.51\\
 & 8 & 1,420 & 720 & 520 & 900 [852, 948] & 700 [568, 823] & 5.54 / 10.88 / 10.18\\
 & 32 & 3,093 & 2,142 & 1,365 & 1,728 [1,585, 1,873] & 950 [542, 1,402] & 9.93 / 14.34 / 14.71\\
Llama-8B / OBQA & 1 & 748 & 438 & 438 & 310 [289, 331] & 310 [289, 331] & 1.34 / 2.29 / 2.28\\
 & 8 & 1,145 & 1,109 & 724 & 421 [382, 461] & 36 [$-$3, 82] & 6.97 / 7.20 / 6.53\\
 & 32 & 2,258 & 2,321 & 1,460 & 797 [727, 866] & $-$63 [$-$121, 1] & 13.96 / 13.57 / 13.74\\
Llama-8B / ARC & 1 & 810 & 401 & 402 & 409 [372, 445] & 409 [373, 445] & 1.23 / 2.49 / 2.49\\
 & 8 & 1,359 & 1,223 & 709 & 650 [590, 717] & 136 [52, 235] & 5.79 / 6.42 / 5.92\\
 & 32 & 3,005 & 2,698 & 1,578 & 1,427 [1,275, 1,592] & 307 [140, 488] & 10.21 / 11.33 / 11.75\\\bottomrule
\end{tabular}
\end{table}

\paragraph{Answers and routes.} At $B=1$ the replay reproduces the development skip and change counts of the main experiments for three policies (9 of 284, 7 of 438, and 4 of 206) and differs by one skipped question for large/OBQA (19 of 571 against 19 of 572). Batching pads the inputs, which changes a few probe scores, routes, and answers (Table~\ref{tab:e2a_consistency}): relative to $B=1$, 1 to 6 routes and 0 to 9 policy answers change per setting, and the reference's own answers change on 1 to 7 questions. The share of skipped answers that differ from the reference at the same batch size stays between 1.6 and 3.3\%. Because batching changes outputs, we do not claim that a certificate obtained at batch size one transfers to another batch size; an operator should calibrate at the batch size it serves.

\begin{table}[!htbp]\centering\scriptsize\revcolor
\caption{\rev{Routes and answers under batching (early-return policy). Changed: skipped requests whose answer differs from the reference at the same $B$. Versus $B=1$: requests whose route, policy answer, or reference answer differs from the same arm at $B=1$.}}\label{tab:e2a_consistency}
\begin{tabular}{lrcccccc}\toprule
 & & & Changed among & \multicolumn{3}{c}{Versus $B=1$}\\\cmidrule(lr){5-7}
Policy & $B$ & Skipped & skipped & Routes & Policy answers & Reference answers\\\midrule
large / OBQA & 1 & 571 / 742 & 19 (3.3\%) & -- & -- & --\\
 & 8 & 573 / 742 & 18 (3.1\%) & 6 & 5 & 4\\
 & 32 & 572 / 742 & 19 (3.3\%) & 3 & 5 & 7\\
large / ARC & 1 & 284 / 299 & 9 (3.2\%) & -- & -- & --\\
 & 8 & 283 / 299 & 9 (3.2\%) & 1 & 0 & 2\\
 & 32 & 283 / 299 & 8 (2.8\%) & 3 & 2 & 1\\
Llama-8B / OBQA & 1 & 438 / 742 & 7 (1.6\%) & -- & -- & --\\
 & 8 & 442 / 742 & 7 (1.6\%) & 6 & 9 & 7\\
 & 32 & 440 / 742 & 7 (1.6\%) & 6 & 9 & 4\\
Llama-8B / ARC & 1 & 206 / 299 & 4 (1.9\%) & -- & -- & --\\
 & 8 & 206 / 299 & 5 (2.4\%) & 2 & 5 & 3\\
 & 32 & 206 / 299 & 5 (2.4\%) & 4 & 3 & 2\\\bottomrule
\end{tabular}
\end{table}

\endgroup

%% file: new_app_K.tex
\section{Related approaches}\label{app:related}

Table~\ref{tab:related} summarizes the comparison of Section~\ref{sec:related}: what each family of methods controls, whether it gives a finite-sample guarantee, and whether the cheap path is a prefix of the skipped one.

\begin{table}[!htbp]\centering\scriptsize
\caption{\rev{Skipping a message shares the consistency-risk framework of risk-controlled early exiting; its cheap path is not a prefix of the skipped one.} Skipped path: the costlier computation avoided; cheap path: what runs instead. Cheap path is a prefix: the skipped path begins with it \rev{(for early exit, within each decoding step)}; some: only some methods.}
\label{tab:related}
\setlength{\tabcolsep}{2pt}
\begin{tabular*}{\linewidth}{@{}l@{\hspace{4pt}\extracolsep{\fill}}lcc@{}}\toprule
Method family & Controlled quantity & \begin{tabular}[b]{@{}c@{}}Finite-sample\\guarantee\end{tabular} & \begin{tabular}[b]{@{}c@{}}Cheap path\\is a prefix\end{tabular}\\\midrule
Risk-controlled early exit & agreement with the full model & yes & yes\\
Routing and cascades & correctness or quality & some & no\\
Online cascade learning & agreement with the large model & no & no\\
Selective LLM judging & agreement with human labels & yes & no\\
Skipping a message (ours) & agreement with a deployed path & yes & no\\\bottomrule
\end{tabular*}
\end{table}

\paragraph{Cascades and selective judges.} Cascades share both structural features of message skipping: a gate that does not see the skipped computation, and a gate cost paid on every query. However, they mostly target correctness. Online cascade learning \citep{nie2024online} has no finite-sample guarantee, and selective LLM judges \citep{jung2025trust} need human labels and assume a better escalated judge.

\paragraph{Pruning communication and skipping retrieval.} Other work also removes communication or evidence to save cost, but optimizes task quality rather than agreement with a deployed path. Sparse communication topologies let debating agents exchange fewer messages with little loss in accuracy \citep{li2024sparse}, and GPTSwarm optimizes which edges of an agent graph to keep \citep{zhuge2024gptswarm}, like AgentPrune \citep{zhang2025agentprune} and AgentDropout, which also removes agents \citep{wang2025agentdropout}; Optima trains agents to communicate with fewer tokens \citep{chen2025optima}. Self-RAG learns when to retrieve \citep{asai2024selfrag}, and Adaptive-RAG routes queries to retrieval strategies by predicted complexity \citep{jeong2024adaptiverag}. \citet{yang2026protocol} find that predicting failure is easier than predicting which collaboration protocol pays off. None of these gives a finite-sample bound on how often skipping changes the system's answers, and none treats the skipped path as a fixed reference.

%% file: new_app_L.tex
\section{Prompts and answer parser}\label{app:prompts}

This appendix gives every prompt template and the answer parser verbatim. Templates are filled by the frozen prompt code and rendered by each model's own chat template with a generation prompt: the Qwen2.5 helper template inserts its default system message, the Qwen3 template with thinking disabled inserts an empty think block, the Llama-3.1 template inserts its default system header (knowledge-cutoff and date lines), and no system prompt is added for any receiver. Receiver inputs are tokenized without adding a second BOS token. Every rendering below was checked against saved run records (token counts, input ids, and probe ids). The released dataset contains fully rendered examples for each receiver and benchmark.

\paragraph{Helper instruction (Text).} \texttt{\{question\}} is the question stem, a blank line, \texttt{Choices:}, and one line \texttt{X.~<option>} per displayed option. The helper writes at most 256 tokens.
{\fontsize{6.3}{7.4}\selectfont
\begin{verbatim}
In one clear sentence, describe the most essential background knowledge needed to answer the question:

{question}

Do NOT directly solve or give answer to the question.
\end{verbatim}}

\paragraph{Receiver (receiver-only, C2C, and the last turn of Text).} The label list in the format line is the question's displayed labels joined by \texttt{/} (for example \texttt{A/B/C/D}, or \texttt{A/.../J} for ten MMLU-Pro options). ARC questions labeled 1--4 are displayed and parsed as A--D. The receiver generates at most 64 tokens.
{\fontsize{6.3}{7.4}\selectfont
\begin{verbatim}
Accurately answer the following question:

{question}

Choices:
{choices}

Instructions:
- Carefully read the question and all options.
- Select the single most correct answer.
- Respond ONLY in the following format: "The correct answer is A/B/C/D".
- Do not include any explanations, additional text, or punctuation besides the answer.

The correct answer is
\end{verbatim}}

\paragraph{Text path, C2C, and probe.} The Text receiver sees three turns: the helper instruction as a user turn, the helper's message as an assistant turn, and the receiver message above as the final user turn. The C2C receiver sees the receiver message alone, with the helper's cache fused in. The probe is one forward pass of the receiver on the rendered receiver-only input followed by the four tokens \texttt{The correct answer is} (so the phrase that ends the user turn also opens the receiver's answer); it sums the next-token probability over each displayed label's token set, renormalizes over the displayed labels, and returns $1-\max_l p_l$. The helper-aware score of Appendix~\ref{app:predict} runs the same prefill with the helper model.

\paragraph{Text+fact.} Only the helper's input changes: the line \texttt{Useful fact: <fact1>} is placed immediately before the question stem inside \texttt{\{question\}}. The receiver's first turn is the fact-free helper instruction.

\paragraph{GSM8K (Appendix~\ref{app:gsm}).} The helper instruction is the one above with \texttt{\{question\}} the bare question. The receiver message, whose second line keeps the trailing spaces of the frozen template:
{\fontsize{6.3}{7.4}\selectfont
\begin{verbatim}
Accurately answer the following question:

{question}

Instructions:
- Carefully read the question.
- Let's think step by step and explain your reasoning briefly.
- End your response with the exact string "The answer is <number>."
\end{verbatim}}

\paragraph{Short-answer tasks.} For the short-answer pilot and full run, the receiver template above was edited only where the task requires, each edit applied exactly once: the options block, ``and all options'', the answer-format line, and the trailing cue were removed, a task instruction was added, and for passage datasets the passage was placed before the question. The helper instruction is unchanged, with the passage (if any) and the question as \texttt{\{question\}}. The receiver generates at most 32 tokens. Passage datasets:
{\fontsize{6.3}{7.4}\selectfont
\begin{verbatim}
Accurately answer the following question:

Passage:
{passage}

{question}

Instructions:
- Carefully read the question.
- Select the single most correct answer.
- Answer with the shortest span from the passage that answers the question. Do not explain.
- Do not include any explanations, additional text, or punctuation besides the answer.
\end{verbatim}}
Closed-book datasets use the same template without the passage block and with this third bullet:
{\fontsize{6.3}{7.4}\selectfont
\begin{verbatim}
- Answer with only the answer: a name, number, date or short phrase, not a sentence. Do not explain.
\end{verbatim}}
\noindent The answer is extracted by stripping whitespace and leading Markdown or quote characters, removing a leading \texttt{Answer:}, \texttt{Final answer:}, or \texttt{The answer is}, keeping the first nonempty line and removing a trailing period, then applying Unicode NFKD with accents removed and the official SQuAD normalization (lowercase, no punctuation, no articles, collapsed whitespace). An empty result is \textsc{invalid}.

\paragraph{Multiple-choice answer parser (version 2).} The parser reads the receiver's raw output and the displayed labels. It never sees the gold answer, the question id, the method, or the option texts.
\begin{itemize}
\item \emph{Normalization.} Unicode NFKC; strip; remove one surrounding Markdown code fence; delete \texttt{**}, \texttt{\_\_}, and backticks; map curly quotes to straight quotes; remove one pair of quotes wrapping the whole output. An empty output is \textsc{invalid}.
\item \emph{Selections.} A selection comes from (1) a leading label at the very start, \texttt{(X)}, \texttt{[X]}, or a capital letter followed by punctuation or the end of the text; (2) a declaration such as \texttt{the final answer is} or \texttt{answer:}; (3) an embedded selection of the form \texttt{the correct \ldots\ is}; (4) a choice verb, \texttt{I choose/select/pick}; or (5) a reverse selection, \texttt{option X is correct}. After (2)--(4) the next label is read and upper-cased.
\item \emph{Not selections.} Patterns inside quotes or on quoted lines; patterns in a sentence with a hedge (\emph{if, whether, unless, assuming, suppose, hypothetically, example, quoted, says, said, claims, incorrect, not}); a label followed by \texttt{=}, \texttt{+}, \texttt{*}, \texttt{/}, \texttt{\^{}}, or a degree sign; a lower-case letter followed by a word; a label followed by a conditional verb; a declaration followed by an explanation (\emph{because, since, due to, that, based on}).
\item \emph{Joined labels.} Labels joined by \texttt{/}, \texttt{|}, commas, \emph{or}, \emph{and}, or \emph{versus} form a multiple selection.
\item \emph{Resolution.} A final declaration or one preceded by a correction word (\emph{actually, correction, on second thought, instead, sorry}) starts a new scope. If the remaining single selections agree on one displayed label, an earlier listing of exactly all labels is ignored. Otherwise the output is \textsc{invalid} when a selected letter is not displayed, when a multiple selection, negated declaration, or declaration without a label remains, when two different labels remain, when several option headings follow each other with no selection after the last, when the chosen label is later called incorrect, or when the output refuses without a final selection. An output with no selection is \textsc{invalid}.
\item \emph{Convention.} \textsc{invalid} is its own answer value: it never matches the gold label, two \textsc{invalid} answers agree, and \textsc{invalid} against a letter is an answer change.
\end{itemize}
The code repository lists the exact regular expressions and cue-word lists.

%% file: manuscript.bbl
\begin{thebibliography}{56}
\providecommand{\natexlab}[1]{#1}
\providecommand{\url}[1]{\texttt{#1}}
\expandafter\ifx\csname urlstyle\endcsname\relax
  \providecommand{\doi}[1]{doi: #1}\else
  \providecommand{\doi}{doi: \begingroup \urlstyle{rm}\Url}\fi

\bibitem[Angelopoulos et~al.(2024)Angelopoulos, Bates, Fisch, Lei, and
  Schuster]{angelopoulos2024crc}
Anastasios~N. Angelopoulos, Stephen Bates, Adam Fisch, Lihua Lei, and Tal
  Schuster.
\newblock Conformal risk control.
\newblock In \emph{International Conference on Learning Representations}, 2024.

\bibitem[Angelopoulos et~al.(2025)Angelopoulos, Bates, Cand\`es, Jordan, and
  Lei]{angelopoulos2022ltt}
Anastasios~N. Angelopoulos, Stephen Bates, Emmanuel~J. Cand\`es, Michael~I.
  Jordan, and Lihua Lei.
\newblock Learn then test: Calibrating predictive algorithms to achieve risk
  control.
\newblock \emph{The Annals of Applied Statistics}, 19\penalty0 (2), 2025.
\newblock \doi{10.1214/24-AOAS1998}.

\bibitem[Asai et~al.(2024)Asai, Wu, Wang, Sil, and Hajishirzi]{asai2024selfrag}
Akari Asai, Zeqiu Wu, Yizhong Wang, Avirup Sil, and Hannaneh Hajishirzi.
\newblock Self-{RAG}: Learning to retrieve, generate, and critique through
  self-reflection.
\newblock In \emph{International Conference on Learning Representations}, 2024.

\bibitem[Chen et~al.(2024)Chen, Zaharia, and Zou]{chen2023frugalgpt}
Lingjiao Chen, Matei Zaharia, and James Zou.
\newblock {FrugalGPT}: How to use large language models while reducing cost and
  improving performance.
\newblock \emph{Transactions on Machine Learning Research}, 2024.
\newblock URL \url{https://openreview.net/forum?id=cSimKw5p6R}.

\bibitem[Chen et~al.(2025)Chen, Yuan, Qian, Yang, Liu, and Sun]{chen2025optima}
Weize Chen, Jiarui Yuan, Chen Qian, Cheng Yang, Zhiyuan Liu, and Maosong Sun.
\newblock Optima: Optimizing effectiveness and efficiency for {LLM}-based
  multi-agent system.
\newblock In \emph{Findings of the Association for Computational Linguistics:
  ACL 2025}, pp.\  11534--11557. Association for Computational Linguistics,
  2025.
\newblock \doi{10.18653/v1/2025.findings-acl.601}.
\newblock URL \url{https://aclanthology.org/2025.findings-acl.601/}.

\bibitem[Clark et~al.(2018)Clark, Cowhey, Etzioni, Khot, Sabharwal, Schoenick,
  and Tafjord]{clark2018arc}
Peter Clark, Isaac Cowhey, Oren Etzioni, Tushar Khot, Ashish Sabharwal, Carissa
  Schoenick, and Oyvind Tafjord.
\newblock Think you have solved question answering? {T}ry {ARC}, the {AI2}
  reasoning challenge, 2018.
\newblock URL \url{https://arxiv.org/abs/1803.05457}.
\newblock arXiv:1803.05457.

\bibitem[Clopper \& Pearson(1934)Clopper and Pearson]{clopper1934}
C.~J. Clopper and E.~S. Pearson.
\newblock The use of confidence or fiducial limits illustrated in the case of
  the binomial.
\newblock \emph{Biometrika}, 26\penalty0 (4):\penalty0 404--413, 1934.
\newblock \doi{10.1093/biomet/26.4.404}.

\bibitem[Cobbe et~al.(2021)Cobbe, Kosaraju, Bavarian, Chen, Jun, Kaiser,
  Plappert, Tworek, Hilton, Nakano, Hesse, and Schulman]{cobbe2021gsm8k}
Karl Cobbe, Vineet Kosaraju, Mohammad Bavarian, Mark Chen, Heewoo Jun, Lukasz
  Kaiser, Matthias Plappert, Jerry Tworek, Jacob Hilton, Reiichiro Nakano,
  Christopher Hesse, and John Schulman.
\newblock Training verifiers to solve math word problems, 2021.
\newblock URL \url{https://arxiv.org/abs/2110.14168}.
\newblock arXiv:2110.14168.

\bibitem[Deng et~al.(2024)Deng, Zhao, Tang, Gerstein, and
  Cohan]{deng2024contamination}
Chunyuan Deng, Yilun Zhao, Xiangru Tang, Mark Gerstein, and Arman Cohan.
\newblock Investigating data contamination in modern benchmarks for large
  language models.
\newblock In \emph{Proceedings of the 2024 Conference of the North American
  Chapter of the Association for Computational Linguistics: Human Language
  Technologies (Volume 1: Long Papers)}, pp.\  8706--8719. Association for
  Computational Linguistics, 2024.
\newblock \doi{10.18653/v1/2024.naacl-long.482}.
\newblock URL \url{https://aclanthology.org/2024.naacl-long.482/}.

\bibitem[Ding et~al.(2024)Ding, Mallick, Wang, Sim, Mukherjee, R\"{u}hle,
  Lakshmanan, and Awadallah]{ding2024hybrid}
Dujian Ding, Ankur Mallick, Chi Wang, Robert Sim, Subhabrata Mukherjee, Victor
  R\"{u}hle, Laks Lakshmanan, and Ahmed~H Awadallah.
\newblock Hybrid {LLM}: Cost-efficient and quality-aware query routing.
\newblock In \emph{International Conference on Learning Representations}, 2024.
\newblock URL
  \url{https://proceedings.iclr.cc/paper_files/paper/2024/file/b47d93c99fa22ac0b377578af0a1f63a-Paper-Conference.pdf}.

\bibitem[Du et~al.(2024)Du, Li, Torralba, Tenenbaum, and
  Mordatch]{du2024debate}
Yilun Du, Shuang Li, Antonio Torralba, Joshua~B. Tenenbaum, and Igor Mordatch.
\newblock Improving factuality and reasoning in language models through
  multiagent debate.
\newblock In \emph{Proceedings of the 41st International Conference on Machine
  Learning}, volume 235 of \emph{Proceedings of Machine Learning Research},
  pp.\  11733--11763. PMLR, 2024.
\newblock URL \url{https://proceedings.mlr.press/v235/du24e.html}.

\bibitem[Dunn(1961)]{dunn1961multiple}
Olive~Jean Dunn.
\newblock Multiple comparisons among means.
\newblock \emph{Journal of the American Statistical Association}, 56\penalty0
  (293):\penalty0 52--64, 1961.
\newblock \doi{10.1080/01621459.1961.10482090}.

\bibitem[Dutta et~al.(2024)Dutta, Krishnan, Kwatra, and
  Ramjee]{dutta2024accuracy}
Abhinav Dutta, Sanjeev Krishnan, Nipun Kwatra, and Ramachandran Ramjee.
\newblock Accuracy is not all you need.
\newblock In \emph{Advances in Neural Information Processing Systems},
  volume~37, 2024.

\bibitem[Fisch et~al.(2019)Fisch, Talmor, Jia, Seo, Choi, and
  Chen]{fisch2019mrqa}
Adam Fisch, Alon Talmor, Robin Jia, Minjoon Seo, Eunsol Choi, and Danqi Chen.
\newblock {MRQA} 2019 shared task: Evaluating generalization in reading
  comprehension.
\newblock In \emph{Proceedings of the 2nd Workshop on Machine Reading for
  Question Answering}, pp.\  1--13, 2019.

\bibitem[Fu et~al.(2026)Fu, Min, Zhang, Yan, Dai, Ouyang, and Wang]{fu2025c2c}
Tianyu Fu, Zihan Min, Hanling Zhang, Jichao Yan, Guohao Dai, Wanli Ouyang, and
  Yu~Wang.
\newblock Cache-to-cache: Direct semantic communication between large language
  models.
\newblock In \emph{International Conference on Learning Representations}, 2026.
\newblock URL \url{https://openreview.net/forum?id=LeatkxrBCi}.

\bibitem[Geifman \& El-Yaniv(2017)Geifman and El-Yaniv]{geifman2017selective}
Yonatan Geifman and Ran El-Yaniv.
\newblock Selective classification for deep neural networks.
\newblock In \emph{Advances in Neural Information Processing Systems},
  volume~30. Curran Associates, Inc., 2017.
\newblock URL
  \url{https://proceedings.neurips.cc/paper_files/paper/2017/file/4a8423d5e91fda00bb7e46540e2b0cf1-Paper.pdf}.

\bibitem[Geng et~al.(2024)Geng, Cai, Wang, Koeppl, Nakov, and
  Gurevych]{geng2024survey}
Jiahui Geng, Fengyu Cai, Yuxia Wang, Heinz Koeppl, Preslav Nakov, and Iryna
  Gurevych.
\newblock A survey of confidence estimation and calibration in large language
  models.
\newblock In \emph{Proceedings of the 2024 Conference of the North American
  Chapter of the Association for Computational Linguistics: Human Language
  Technologies (Volume 1: Long Papers)}, pp.\  6577--6595. Association for
  Computational Linguistics, 2024.
\newblock \doi{10.18653/v1/2024.naacl-long.366}.
\newblock URL \url{https://aclanthology.org/2024.naacl-long.366/}.

\bibitem[Grattafiori et~al.(2024)Grattafiori, Dubey, Jauhri, Pandey, Kadian,
  et~al.]{grattafiori2024llama3}
Aaron Grattafiori, Abhimanyu Dubey, Abhinav Jauhri, Abhinav Pandey, Abhishek
  Kadian, et~al.
\newblock The {Llama} 3 herd of models, 2024.
\newblock URL \url{https://arxiv.org/abs/2407.21783}.
\newblock arXiv:2407.21783.

\bibitem[Gupta et~al.(2024)Gupta, Narasimhan, Jitkrittum, Rawat, Menon, and
  Kumar]{gupta2024cascades}
Neha Gupta, Harikrishna Narasimhan, Wittawat Jitkrittum, Ankit~Singh Rawat,
  Aditya~Krishna Menon, and Sanjiv Kumar.
\newblock Language model cascades: Token-level uncertainty and beyond.
\newblock In \emph{International Conference on Learning Representations}, 2024.

\bibitem[Hendrycks \& Gimpel(2017)Hendrycks and Gimpel]{hendrycks2017baseline}
Dan Hendrycks and Kevin Gimpel.
\newblock A baseline for detecting misclassified and out-of-distribution
  examples in neural networks.
\newblock In \emph{International Conference on Learning Representations}, 2017.
\newblock URL \url{https://arxiv.org/abs/1610.02136}.

\bibitem[Heo et~al.(2026)Heo, Shafipour, Zhao, Golub, Kamani, Borkar, Chandran,
  Zardoshti, and Rouhani]{heo2026crossmodelkv}
Taekyung Heo, Rasoul Shafipour, Ritchie Zhao, Maximilian Golub, Mohammad~Mahdi
  Kamani, Ritika Borkar, Makesh~Tarun Chandran, Pantea Zardoshti, and
  Bita~Darvish Rouhani.
\newblock Cross-model {KV} cache transfer in {LLM} families: A closed-form
  linear mapping for prefill reuse, 2026.
\newblock URL \url{https://arxiv.org/abs/2608.03893}.
\newblock arXiv:2608.03893.

\bibitem[Hu et~al.(2024)Hu, Bieker, Li, Jiang, Keigwin, Ranganath, Keutzer, and
  Upadhyay]{hu2024routerbench}
Qitian~Jason Hu, Jacob Bieker, Xiuyu Li, Nan Jiang, Benjamin Keigwin, Gaurav
  Ranganath, Kurt Keutzer, and Shriyash~Kaustubh Upadhyay.
\newblock {RouterBench}: A benchmark for multi-{LLM} routing system, 2024.
\newblock URL \url{https://arxiv.org/abs/2403.12031}.
\newblock arXiv:2403.12031.

\bibitem[Jazbec et~al.(2024)Jazbec, Timans, Veljkovi\'{c}, Sakmann, Zhang,
  Naesseth, and Nalisnick]{jazbec2024fastyetsafe}
Metod Jazbec, Alexander Timans, Tin~Had\v{z}i Veljkovi\'{c}, Kaspar Sakmann,
  Dan Zhang, Christian~A. Naesseth, and Eric Nalisnick.
\newblock Fast yet safe: Early-exiting with risk control.
\newblock In \emph{Advances in Neural Information Processing Systems},
  volume~37, pp.\  129825--129854. Curran Associates, Inc., 2024.
\newblock \doi{10.52202/079017-4124}.
\newblock URL
  \url{https://proceedings.neurips.cc/paper_files/paper/2024/file/ea5a63f7ddb82e58623693fd1f4933f7-Paper-Conference.pdf}.

\bibitem[Jeong et~al.(2024)Jeong, Baek, Cho, Hwang, and
  Park]{jeong2024adaptiverag}
Soyeong Jeong, Jinheon Baek, Sukmin Cho, Sung~Ju Hwang, and Jong~C. Park.
\newblock Adaptive-{RAG}: Learning to adapt retrieval-augmented large language
  models through question complexity.
\newblock In \emph{Proceedings of the 2024 Conference of the North American
  Chapter of the Association for Computational Linguistics: Human Language
  Technologies (Volume 1: Long Papers)}, pp.\  7036--7050, 2024.
\newblock \doi{10.18653/v1/2024.naacl-long.389}.

\bibitem[Jiang et~al.(2023{\natexlab{a}})Jiang, Ren, and Lin]{jiang2023blender}
Dongfu Jiang, Xiang Ren, and Bill~Yuchen Lin.
\newblock {LLM-Blender}: Ensembling large language models with pairwise ranking
  and generative fusion.
\newblock In \emph{Proceedings of the 61st Annual Meeting of the Association
  for Computational Linguistics (Volume 1: Long Papers)}, pp.\  14165--14178,
  2023{\natexlab{a}}.

\bibitem[Jiang et~al.(2023{\natexlab{b}})Jiang, Xu, Gao, Sun, Liu, Dwivedi-Yu,
  Yang, Callan, and Neubig]{jiang2023flare}
Zhengbao Jiang, Frank Xu, Luyu Gao, Zhiqing Sun, Qian Liu, Jane Dwivedi-Yu,
  Yiming Yang, Jamie Callan, and Graham Neubig.
\newblock Active retrieval augmented generation.
\newblock In \emph{Proceedings of the 2023 Conference on Empirical Methods in
  Natural Language Processing}, pp.\  7969--7992, 2023{\natexlab{b}}.
\newblock \doi{10.18653/v1/2023.emnlp-main.495}.

\bibitem[Joshi et~al.(2017)Joshi, Choi, Weld, and
  Zettlemoyer]{joshi2017triviaqa}
Mandar Joshi, Eunsol Choi, Daniel Weld, and Luke Zettlemoyer.
\newblock {TriviaQA}: A large scale distantly supervised challenge dataset for
  reading comprehension.
\newblock In \emph{Proceedings of the 55th Annual Meeting of the Association
  for Computational Linguistics}, pp.\  1601--1611, 2017.

\bibitem[Jung et~al.(2025)Jung, Brahman, and Choi]{jung2025trust}
Jaehun Jung, Faeze Brahman, and Yejin Choi.
\newblock Trust or escalate: {LLM} judges with provable guarantees for human
  agreement.
\newblock In \emph{International Conference on Learning Representations}, 2025.

\bibitem[Khot et~al.(2020)Khot, Clark, Guerquin, Jansen, and
  Sabharwal]{khot2020qasc}
Tushar Khot, Peter Clark, Michal Guerquin, Peter Jansen, and Ashish Sabharwal.
\newblock {QASC}: A dataset for question answering via sentence composition.
\newblock In \emph{Proceedings of the AAAI Conference on Artificial
  Intelligence}, volume~34, pp.\  8082--8090, 2020.
\newblock \doi{10.1609/aaai.v34i05.6319}.

\bibitem[Kwiatkowski et~al.(2019)Kwiatkowski, Palomaki, Redfield, Collins,
  Parikh, Alberti, Epstein, Polosukhin, Devlin, Lee, Toutanova, Jones, Kelcey,
  Chang, Dai, Uszkoreit, Le, and Petrov]{kwiatkowski2019nq}
Tom Kwiatkowski, Jennimaria Palomaki, Olivia Redfield, Michael Collins, Ankur
  Parikh, Chris Alberti, Danielle Epstein, Illia Polosukhin, Jacob Devlin,
  Kenton Lee, Kristina Toutanova, Llion Jones, Matthew Kelcey, Ming-Wei Chang,
  Andrew~M. Dai, Jakob Uszkoreit, Quoc Le, and Slav Petrov.
\newblock {Natural Questions}: A benchmark for question answering research.
\newblock \emph{Transactions of the Association for Computational Linguistics},
  7:\penalty0 452--466, 2019.

\bibitem[Lee et~al.(2019)Lee, Chang, and Toutanova]{lee2019latent}
Kenton Lee, Ming-Wei Chang, and Kristina Toutanova.
\newblock Latent retrieval for weakly supervised open domain question
  answering.
\newblock In \emph{Proceedings of the 57th Annual Meeting of the Association
  for Computational Linguistics}, pp.\  6086--6096, 2019.

\bibitem[Li et~al.(2023)Li, Hammoud, Itani, Khizbullin, and
  Ghanem]{li2023camel}
Guohao Li, Hasan Hammoud, Hani Itani, Dmitrii Khizbullin, and Bernard Ghanem.
\newblock {CAMEL}: Communicative agents for ``mind'' exploration of large
  language model society.
\newblock In \emph{Advances in Neural Information Processing Systems},
  volume~36, pp.\  51991--52008. Curran Associates, Inc., 2023.
\newblock \doi{10.52202/075280-2264}.
\newblock URL
  \url{https://proceedings.neurips.cc/paper_files/paper/2023/file/a3621ee907def47c1b952ade25c67698-Paper-Conference.pdf}.

\bibitem[Li et~al.(2024)Li, Du, Zhang, Hou, Grabowski, Li, and
  Ie]{li2024sparse}
Yunxuan Li, Yibing Du, Jiageng Zhang, Le~Hou, Peter Grabowski, Yeqing Li, and
  Eugene Ie.
\newblock Improving multi-agent debate with sparse communication topology.
\newblock In \emph{Findings of the Association for Computational Linguistics:
  EMNLP 2024}, pp.\  7281--7294, 2024.

\bibitem[Mallen et~al.(2023)Mallen, Asai, Zhong, Das, Khashabi, and
  Hajishirzi]{mallen2023trust}
Alex Mallen, Akari Asai, Victor Zhong, Rajarshi Das, Daniel Khashabi, and
  Hannaneh Hajishirzi.
\newblock When not to trust language models: Investigating effectiveness of
  parametric and non-parametric memories.
\newblock In \emph{Proceedings of the 61st Annual Meeting of the Association
  for Computational Linguistics (Volume 1: Long Papers)}, pp.\  9802--9822,
  2023.
\newblock \doi{10.18653/v1/2023.acl-long.546}.

\bibitem[Mihaylov et~al.(2018)Mihaylov, Clark, Khot, and
  Sabharwal]{mihaylov2018obqa}
Todor Mihaylov, Peter Clark, Tushar Khot, and Ashish Sabharwal.
\newblock Can a suit of armor conduct electricity? {A} new dataset for open
  book question answering.
\newblock In \emph{Proceedings of the 2018 Conference on Empirical Methods in
  Natural Language Processing}, pp.\  2381--2391. Association for Computational
  Linguistics, 2018.
\newblock \doi{10.18653/v1/D18-1260}.
\newblock URL \url{https://aclanthology.org/D18-1260/}.

\bibitem[Nie et~al.(2024)Nie, Ding, Hu, Jermaine, and Chaudhuri]{nie2024online}
Lunyiu Nie, Zhimin Ding, Erdong Hu, Christopher Jermaine, and Swarat Chaudhuri.
\newblock Online cascade learning for efficient inference over streams.
\newblock In \emph{Proceedings of the 41st International Conference on Machine
  Learning}, volume 235 of \emph{Proceedings of Machine Learning Research},
  pp.\  38071--38090, 2024.
\newblock URL \url{https://proceedings.mlr.press/v235/nie24a.html}.

\bibitem[Ong et~al.(2025)Ong, Almahairi, Wu, Chiang, Wu, Gonzalez, Kadous, and
  Stoica]{ong2024routellm}
Isaac Ong, Amjad Almahairi, Vincent Wu, Wei-Lin Chiang, Tianhao Wu, Joseph~E.
  Gonzalez, M.~Waleed Kadous, and Ion Stoica.
\newblock {RouteLLM}: Learning to route {LLM}s from preference data.
\newblock In \emph{International Conference on Learning Representations}, 2025.

\bibitem[Pezeshkpour \& Hruschka(2024)Pezeshkpour and
  Hruschka]{pezeshkpour2024order}
Pouya Pezeshkpour and Estevam Hruschka.
\newblock Large language models sensitivity to the order of options in
  multiple-choice questions.
\newblock In \emph{Findings of the Association for Computational Linguistics:
  NAACL 2024}, pp.\  2006--2017, 2024.

\bibitem[Pona et~al.(2026)Pona, Kazemi, Hosseini, Du, Watson, Simeone, and
  Paoletti]{pona2026ctd}
Edoardo Pona, Milad Kazemi, Mehran Hosseini, Yali Du, David Watson, Osvaldo
  Simeone, and Nicola Paoletti.
\newblock Calibrate-then-delegate: Safety monitoring with risk and budget
  guarantees via model cascades, 2026.
\newblock URL \url{https://arxiv.org/abs/2604.14251}.
\newblock arXiv:2604.14251.

\bibitem[Qwen et~al.(2025)Qwen, Yang, Yang, Zhang, Hui, Zheng, Yu, Li, Liu,
  Huang, Wei, Lin, Yang, Tu, Zhang, Yang, Yang, Zhou, Lin, Dang, Lu, Bao, Yang,
  Yu, Li, Xue, Zhang, Zhu, Men, Lin, Li, Tang, Xia, Ren, Ren, Fan, Su, Zhang,
  Wan, Liu, Cui, Zhang, Qiu, et~al.]{qwen2025qwen25}
Qwen, An~Yang, Baosong Yang, Beichen Zhang, Binyuan Hui, Bo~Zheng, Bowen Yu,
  Chengyuan Li, Dayiheng Liu, Fei Huang, Haoran Wei, Huan Lin, Jian Yang,
  Jianhong Tu, Jianwei Zhang, Jianxin Yang, Jiaxi Yang, Jingren Zhou, Junyang
  Lin, Kai Dang, Keming Lu, Keqin Bao, Kexin Yang, Le~Yu, Mei Li, Mingfeng Xue,
  Pei Zhang, Qin Zhu, Rui Men, Runji Lin, Tianhao Li, Tianyi Tang, Tingyu Xia,
  Xingzhang Ren, Xuancheng Ren, Yang Fan, Yang Su, Yichang Zhang, Yu~Wan,
  Yuqiong Liu, Zeyu Cui, Zhenru Zhang, Zihan Qiu, et~al.
\newblock {Qwen2.5} technical report, 2025.
\newblock URL \url{https://arxiv.org/abs/2412.15115}.
\newblock arXiv:2412.15115.

\bibitem[Rajpurkar et~al.(2016)Rajpurkar, Zhang, Lopyrev, and
  Liang]{rajpurkar2016squad}
Pranav Rajpurkar, Jian Zhang, Konstantin Lopyrev, and Percy Liang.
\newblock {SQuAD}: 100,000+ questions for machine comprehension of text.
\newblock In \emph{Proceedings of the 2016 Conference on Empirical Methods in
  Natural Language Processing}, pp.\  2383--2392, 2016.

\bibitem[Schuster et~al.(2022)Schuster, Fisch, Gupta, Dehghani, Bahri, Tran,
  Tay, and Metzler]{schuster2022calm}
Tal Schuster, Adam Fisch, Jai Gupta, Mostafa Dehghani, Dara Bahri, Vinh Tran,
  Yi~Tay, and Donald Metzler.
\newblock Confident adaptive language modeling.
\newblock In \emph{Advances in Neural Information Processing Systems},
  volume~35, pp.\  17456--17472. Curran Associates, Inc., 2022.
\newblock \doi{10.52202/068431-1269}.
\newblock URL
  \url{https://proceedings.neurips.cc/paper_files/paper/2022/file/6fac9e316a4ae75ea244ddcef1982c71-Paper-Conference.pdf}.

\bibitem[Singh et~al.(2019)Singh, Jain, and Sukhbaatar]{singh2019ic3net}
Amanpreet Singh, Tushar Jain, and Sainbayar Sukhbaatar.
\newblock Learning when to communicate at scale in multiagent cooperative and
  competitive tasks.
\newblock In \emph{International Conference on Learning Representations}, 2019.
\newblock URL \url{https://arxiv.org/abs/1812.09755}.

\bibitem[{Team OLMo} et~al.(2025){Team OLMo}, Walsh, Soldaini, Groeneveld,
  et~al.]{olmo2025olmo2}
{Team OLMo}, Pete Walsh, Luca Soldaini, Dirk Groeneveld, et~al.
\newblock 2 {OLMo} 2 furious, 2025.
\newblock URL \url{https://arxiv.org/abs/2501.00656}.
\newblock arXiv:2501.00656.

\bibitem[Uddin \& Bauer(2026)Uddin and Bauer]{uddin2026conformal}
Iqtedar Uddin and Andr{\'e} Bauer.
\newblock Conformal {LLM} routing with distribution-free safety guarantees.
\newblock In \emph{Proceedings of the 64th Annual Meeting of the Association
  for Computational Linguistics (Volume 4: Student Research Workshop)}, pp.\
  791--799. Association for Computational Linguistics, 2026.
\newblock \doi{10.18653/v1/2026.acl-srw.70}.
\newblock URL \url{https://aclanthology.org/2026.acl-srw.70/}.

\bibitem[Uddin et~al.(2026)Uddin, Khider, and Bauer]{uddin2026proactive}
Iqtedar Uddin, Mazin Khider, and Andr{\'e} Bauer.
\newblock Proactive routing to interpretable surrogates with distribution-free
  safety guarantees, 2026.
\newblock URL \url{https://arxiv.org/abs/2603.14623}.
\newblock arXiv:2603.14623.

\bibitem[Wang et~al.(2024)Wang, Ma, Zhang, Ni, Chandra, Guo, Ren, Arulraj, He,
  Jiang, Li, Ku, Wang, Zhuang, Fan, Yue, and Chen]{wang2024mmlupro}
Yubo Wang, Xueguang Ma, Ge~Zhang, Yuansheng Ni, Abhranil Chandra, Shiguang Guo,
  Weiming Ren, Aaran Arulraj, Xuan He, Ziyan Jiang, Tianle Li, Max Ku, Kai
  Wang, Alex Zhuang, Rongqi Fan, Xiang Yue, and Wenhu Chen.
\newblock {MMLU-Pro}: A more robust and challenging multi-task language
  understanding benchmark.
\newblock In \emph{Advances in Neural Information Processing Systems (Datasets
  and Benchmarks Track)}, volume~37, pp.\  95266--95290, 2024.

\bibitem[Wang et~al.(2025)Wang, Wang, Liu, Ding, Zhang, Liu, and
  Zhang]{wang2025agentdropout}
Zhexuan Wang, Yutong Wang, Xuebo Liu, Liang Ding, Miao Zhang, Jie Liu, and Min
  Zhang.
\newblock {A}gent{D}ropout: Dynamic agent elimination for token-efficient and
  high-performance {LLM}-based multi-agent collaboration.
\newblock In \emph{Proceedings of the 63rd Annual Meeting of the Association
  for Computational Linguistics (Volume 1: Long Papers)}, pp.\  24013--24035.
  Association for Computational Linguistics, 2025.
\newblock \doi{10.18653/v1/2025.acl-long.1170}.
\newblock URL \url{https://aclanthology.org/2025.acl-long.1170/}.

\bibitem[Yan et~al.(2021)Yan, Xiong, Kundu, Yang, Deng, Wang, Xia, and
  Soatto]{yan2021pct}
Sijie Yan, Yuanjun Xiong, Kaustav Kundu, Shuo Yang, Siqi Deng, Meng Wang, Wei
  Xia, and Stefano Soatto.
\newblock Positive-congruent training: Towards regression-free model updates.
\newblock In \emph{Proceedings of the IEEE/CVF Conference on Computer Vision
  and Pattern Recognition}, pp.\  14299--14308, 2021.

\bibitem[Yang et~al.(2025)Yang, Li, Yang, Zhang, Hui, Zheng, Yu, Gao, Huang,
  Lv, Zheng, Liu, Zhou, Huang, Hu, Ge, Wei, Lin, Tang, Yang, Tu, Zhang, Yang,
  Yang, Zhou, Zhou, Lin, Dang, Bao, Yang, Yu, Deng, Li, Xue, Li, Zhang, Wang,
  Zhu, Men, Gao, Liu, Luo, Li, Tang, Yin, Ren, Wang, Zhang, Ren, Fan, Su,
  Zhang, Zhang, Wan, Liu, Wang, Cui, Zhang, Zhou, and Qiu]{yang2025qwen3}
An~Yang, Anfeng Li, Baosong Yang, Beichen Zhang, Binyuan Hui, Bo~Zheng, Bowen
  Yu, Chang Gao, Chengen Huang, Chenxu Lv, Chujie Zheng, Dayiheng Liu, Fan
  Zhou, Fei Huang, Feng Hu, Hao Ge, Haoran Wei, Huan Lin, Jialong Tang, Jian
  Yang, Jianhong Tu, Jianwei Zhang, Jianxin Yang, Jiaxi Yang, Jing Zhou,
  Jingren Zhou, Junyang Lin, Kai Dang, Keqin Bao, Kexin Yang, Le~Yu, Lianghao
  Deng, Mei Li, Mingfeng Xue, Mingze Li, Pei Zhang, Peng Wang, Qin Zhu, Rui
  Men, Ruize Gao, Shixuan Liu, Shuang Luo, Tianhao Li, Tianyi Tang, Wenbiao
  Yin, Xingzhang Ren, Xinyu Wang, Xinyu Zhang, Xuancheng Ren, Yang Fan, Yang
  Su, Yichang Zhang, Yinger Zhang, Yu~Wan, Yuqiong Liu, Zekun Wang, Zeyu Cui,
  Zhenru Zhang, Zhipeng Zhou, and Zihan Qiu.
\newblock {Qwen3} technical report, 2025.
\newblock URL \url{https://arxiv.org/abs/2505.09388}.
\newblock arXiv:2505.09388.

\bibitem[Yang et~al.(2026)Yang, Jiang, Yang, Vasudevan, Zheng, Vishwanath, and
  Thakur]{yang2026protocol}
Chih-Hsuan Yang, Jingyan Jiang, Cheng-Hau Yang, Vikram Vasudevan, Huihuo Zheng,
  Venkatram Vishwanath, and Rajeev Thakur.
\newblock {LLM}s can predict failure risk, but struggle to predict which
  collaboration protocol pays off: Cost-aware protocol routing across reasoning
  tasks, 2026.
\newblock URL \url{https://arxiv.org/abs/2608.14927}.
\newblock arXiv:2608.14927.

\bibitem[Yue et~al.(2024)Yue, Zhao, Zhang, Du, and Yao]{yue2024mot}
Murong Yue, Jie Zhao, Min Zhang, Liang Du, and Ziyu Yao.
\newblock Large language model cascades with mixture of thought representations
  for cost-efficient reasoning.
\newblock In \emph{International Conference on Learning Representations}, 2024.

\bibitem[Zhang et~al.(2025)Zhang, Yue, Li, Yun, Wan, Wang, Cheng, Yu, and
  Chen]{zhang2025agentprune}
Guibin Zhang, Yanwei Yue, Zhixun Li, Sukwon Yun, Guancheng Wan, Kun Wang, Dawei
  Cheng, Jeffrey~Xu Yu, and Tianlong Chen.
\newblock Cut the crap: An economical communication pipeline for {LLM}-based
  multi-agent systems.
\newblock In \emph{International Conference on Learning Representations}, 2025.
\newblock URL \url{https://arxiv.org/abs/2410.02506}.

\bibitem[Zheng et~al.(2024)Zheng, Zhou, Meng, Zhou, and Huang]{zheng2024mcq}
Chujie Zheng, Hao Zhou, Fandong Meng, Jie Zhou, and Minlie Huang.
\newblock Large language models are not robust multiple choice selectors.
\newblock In \emph{International Conference on Learning Representations}, 2024.

\bibitem[Zhuge et~al.(2024)Zhuge, Wang, Kirsch, Faccio, Khizbullin, and
  Schmidhuber]{zhuge2024gptswarm}
Mingchen Zhuge, Wenyi Wang, Louis Kirsch, Francesco Faccio, Dmitrii Khizbullin,
  and J{\"u}rgen Schmidhuber.
\newblock {GPTSwarm}: Language agents as optimizable graphs.
\newblock In \emph{Proceedings of the 41st International Conference on Machine
  Learning}, volume 235 of \emph{Proceedings of Machine Learning Research},
  pp.\  62743--62767, 2024.

\bibitem[Zou et~al.(2026)Zou, Qiu, Li, Yang, Tieu, Lu, Shen, Tong, Choi, He,
  Zou, Wang, and Yang]{zou2025latentmas}
Jiaru Zou, Ruizhong Qiu, Gaotang Li, Xiyuan Yang, Katherine Tieu, Pan Lu,
  Ke~Shen, Hanghang Tong, Yejin Choi, Jingrui He, James Zou, Mengdi Wang, and
  Ling Yang.
\newblock Latent collaboration in multi-agent systems.
\newblock In \emph{International Conference on Machine Learning}, 2026.
\newblock URL \url{https://arxiv.org/abs/2511.20639}.
\newblock arXiv:2511.20639.

\end{thebibliography}
